\documentclass[runningheads]{llncs}

\usepackage{eccv}

\usepackage{eccvabbrv}

\usepackage{graphicx}
\usepackage{booktabs}

\usepackage[accsupp]{axessibility}  
\usepackage{minitoc}
\usepackage{multirow}
\usepackage[table]{xcolor}

\newcommand{\nazia}[1]{{\color{orange}\textbf{[NT:} #1]}}

\usepackage[normalem]{ulem}

\renewcommand{\nazia}{}

\definecolor{lightred}{rgb}{1,0.8,0.8}
\definecolor{lightgreen}{rgb}{0.8,1,0.8}
\definecolor{col1}{HTML}{2e59a7}
\definecolor{col2}{HTML}{db824a}
\definecolor{col3}{HTML}{ec5a6b}
\definecolor{col4}{HTML}{5ba17b}
\definecolor{rank1}{RGB}{198,239,206}   
\definecolor{rank2}{RGB}{220,245,214}
\definecolor{rank3}{RGB}{255,235,156}   
\definecolor{rank4}{RGB}{255,220,120}
\definecolor{rank5}{RGB}{255,199,100}
\definecolor{rank6}{RGB}{255,150,100}
\definecolor{rank7}{RGB}{255,105,97}    

\definecolor{iccvblue}{rgb}{0.21,0.49,0.74}
\definecolor{bluegreen}{RGB}{0, 150, 130}  

\usepackage{hyperref}

\usepackage{orcidlink}

\begin{document}
\title{Seeing Isn't Orienting: A Cognitively Informed Hierarchical Benchmark for Object Orientation in MLLMs} 

\titlerunning{Seeing Isn't Orienting}

\author{
Nazia Tasnim\thanks{Equal contribution} \and
Keanu Nichols$^\star$ \and
Yuting Yan\thanks{Work done while at Boston University; currently at Johns Hopkins University - yyan79@jh.edu} \and
Nicholas Ikechukwu \and
Elva Zou \and
Deepti Ghadiyaram \and
Bryan A. Plummer
}

\authorrunning{N. Tasnim \& K. Nichols et al.}

\institute{
Boston University\\
\email{\{nimzia, kmn5409, ncholas, elzou030, dghadiya, bplum\}@bu.edu}
}

\maketitle

\begin{figure}[h]
\centering
\includegraphics[width=0.95\columnwidth]{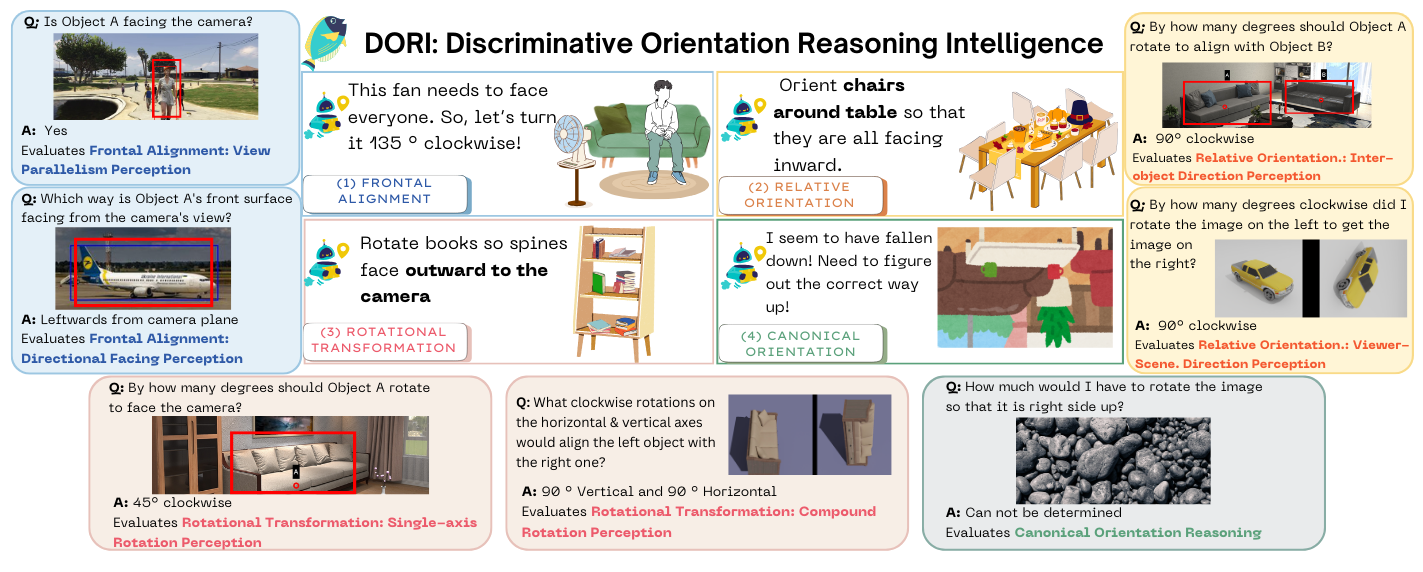}
\vspace{-2mm}
\caption{
DORI organizes object orientation evaluation into four diagnostic dimensions: (1) object's \textbf{directional alignment}, (2) its \textbf{orientation relative to} viewers, scenes, and other objects, (3) required \textbf{rotational transformation} for different objectives, and (4) its natural/\textbf{canonical orientation} in the world. Each dimension evaluates a targeted aspect of object orientation reasoning across diverse visual settings.}
\label{fig:intro_fig}
\vspace{-4mm}
\end{figure}

\begin{abstract}
Humans develop object orientation understanding progressively, from recognizing which way
an object faces, to mentally rotating it, to reasoning about how multiple objects are
oriented relative to each other. Yet current vision-language benchmarks treat orientation
as an afterthought, conflating it with positional relationships and general scene
understanding. We introduce \textbf{D}iscriminative \textbf{O}rientation \textbf{R}easoning
\textbf{I}ntelligence (\textbf{DORI}), a cognition-informed hierarchical benchmark that establishes object orientation as the primary evaluation target. Specifically, DORI decomposes orientation into four dimensions, each assessed at both coarse (categorical) and granular (metric) levels. This results in 33,656 multiple-choice questions over diverse open-vocabulary objects within 13,652 real-world and synthetic images taken from 14 sources. DORI's coarse-to-granular design isolates orientation from confounds such as object recognition difficulty, scene clutter, and linguistic ambiguity through bounding-box isolation, standardized spatial reference frames, and structured prompts. Our evaluation of {$26$} state-of-the-art vision-language models reveals a consistent pattern: \textit{models competent on general spatial benchmarks remain near-random on object-centric orientation tasks}. Even the best models achieve only {$64.2\%$ on coarse and $42.9\%$ on granular judgments, with the largest drops on compound rotations and inter-object reference frame shifts}. Large coarse-to-granular gaps further expose that models rely on categorical heuristics rather than geometric reasoning, a limitation invisible to existing benchmarks. These findings establish orientation understanding as an unsolved challenge in multimodal systems, with direct implications for robotic manipulation, 3D scene reconstruction, and human-AI interaction. Find the dataset here: \url{https://huggingface.co/datasets/appledora/DORI-Benchmark}
\end{abstract}

\section{Introduction}

Object orientation understanding demands complex, multi-stage processing of intrinsic object features, viewer perspective, angular relationships, and reference frame transformations~\cite{harris2024interpreting, kallmayer2023viewpoint, vigano2023mental, chavez2023independent}. 
Humans master this fundamental aspect of visual cognition~\cite{10.1007/3540634932_1,  10.1016/j.artint.2021.103637} through various inherent sensory-motor experiences, proprioceptive integration, and neural formation~\cite{tuthill2018proprioception}. In the human cognitive system, these mechanisms develop progressively from basic frontal orientation recognition to complex rotational transformations~\cite{10.1093/acprof:oso/9780195381634.003.0004,alma991032026461708041,vasilyeva2012development}. This equips us with a crucial aptitude for real-world visual tasks that require precise spatial interaction with the environment, such as tool manipulation and navigation. 
This capability also
underpins real-world tasks such as robotic grasping~\cite{cong2021comprehensive, 
chen2023vision}, autonomous navigation~\cite{ganesan2024comprehensive, 
janai2020computer}, and augmented reality object alignment~\cite{10.5555/3666122.3667212, Pei2024AutonomousWF,wang2023holoassist}.
Multimodal large language models (MLLMs)~\cite{bi2024deepseek, Qwen2VL, journals/corr/abs-2302-13971} are currently attractive for many of these applications, as they perform well on various spatial reasoning tasks~\cite{DBLP:journals/corr/abs-2405-05363,lin2023sphinxjointmixingweights, Pei2024AutonomousWF,10.5555/3666122.3667212,wei2024editable,10610090}. 
 Yet, existing benchmarks reveal consistent failures in orientation tasks~\cite{tong2024cambrian,li2024seed2plus, cheng2024spatialrgpt, Chen_2024_CVPR, liu2024mmbenchmultimodalmodelallaround, lei-etal-2025-scaffolding} while overlooking key limitations: narrow focus on simple directions~\cite{Weston2015TowardsAQ}, synthetic-only data~\cite{wang2025pulsecheck457diagnosticbenchmark6d}, ambiguous prompts~\cite{mirzaee-etal-2021-spartqa}, egocentric bias~\cite{jung2024rightrightenhancingobject}, or small scales~\cite{blink}. 
 These limitations lead to an incomplete assessment of a model's orientation reasoning abilities, potentially failing to identify critical weaknesses in real-world scenarios and 
  mask true geometric reasoning deficits, favoring memorized shortcuts over human-like cognition.
 
As summarized in \cref{fig:intro_fig}, we address these shortcomings by introducing \textbf{D}iscriminative \textbf{O}rientation \textbf{R}easoning \textbf{I}ntelligence (\textbf{DORI}), a multifaceted benchmark to evaluate object orientation understanding in multimodal language models. 
\cref{tab:dimensions} shows how we draw on developmental literature to motivate a principled decomposition 
of orientation understanding into four dimensions of increasing complexity:   (1) frontal alignment perception, (2) rotational transformations, (3) ego and allocentric relative orientation understanding, and (4) natural or canonical orientation of objects.
Our benchmark further employs a two-tiered assessment framework to provide a more holistic understanding of MLLM performance: \textbf{coarse-grained} questions to evaluate basic categorical understanding (\eg, ``is the car facing toward or away from the camera?") and \textbf{fine-grained} questions to probe precise metric relationships (\eg, ``at what angle is the car oriented relative to the camera?'').
DORI leverages \textbf{13652} images from \textbf{14} diverse sources to generate \textbf{33,656} multiple-choice questions, combining real-world images (37\%) with simulated renders (63\%) to ensure we have a large dataset with varying levels of visual complexity. 

\begin{table*}[t]
\centering
\caption{Four diagnostic dimensions in DORI, motivated by prior work on human
visuospatial cognition. The cited literature provides task-level motivation for
evaluating orientation across perception, transformation, reference-frame reasoning,
and canonical pose understanding.}
\label{tab:dimensions}
\renewcommand{\arraystretch}{1.5}
\setlength{\tabcolsep}{4pt}
\resizebox{\textwidth}{!}{%
\begin{tabular}{@{}p{2.2cm} p{3cm} p{3.5cm} p{7.cm}@{}}
\toprule
\textbf{Dimension} & \textbf{Motivating Process} & \textbf{Description} & \textbf{Relevant Human Cog. Dev. Studies} \\
\midrule

\textbf{\textcolor{col1}{(1) Frontal Alignment}}
& Basic Perception
& Identifying a front-facing surface's orientation relative to the viewer.
& Early visual cortex~\cite{BRADDICK20053169, 10.1016/j.visres.2011.01.003} supports viewpoint-invariant frontal recognition via pathways distinct from angular estimation~\cite{harris2024interpreting, Kourtzi01022002}. \\

\midrule

\textbf{\textcolor{col3}{(2) Rotational Transformation}}
& Mental rotation and spatial transformation.
& Simulating orientation changes through single- or multi-axis rotation.
& Overlapping premotor and parietal circuits activate during physical and imagined rotation~\cite{doganci2023embodied, 10.1027/1618-3169/a000505, xue2017uncovering}, with load scaling with transformation complexity~\cite{frick2014development, 10.1080/10400419.2022.2049532, ter2010mental, li2019simon}. \\

\midrule

\textbf{\textcolor{col2}{(3) Relative Orientation}}
& Multi-object frame-of-reference integration.
& Inter-object and viewer-relative orientation reasoning.
& Hippocampal-parietal circuits~\cite{10.3390/neurosci5040050, doi:10.1073/pnas.1504242112, gramann2010human, zaehle2007neural, 10.5334/joc.321} support cross-viewpoint orientation tracking essential for scene comprehension. \\

\midrule

\textbf{\textcolor{col4}{(4) Canonical Orientation}}
& Semantic, physics-grounded orientation knowledge.
& Detecting deviations from expected poses and inferring corrective transformations.
& Ventral stream and prefrontal regions~\cite{10.1038/nn1865, 10.1002/hbm.20328} encode intuitive physics~\cite{ballaz2005visual, mezuman2012learning, harris2024interpreting}, constrained by gravity, feature positioning, and ecological validity~\cite{10.1145/1399504.1360641, bramley2018intuitive, hamrick2011internal}. \\

\bottomrule
\end{tabular}%
\vspace{-8mm}
}
\end{table*}

We generate clear, unambiguous prompt-answer pairs through a rigorous three-step process: (1) isolating objects with bounding boxes to tackle cluttered scenes, (2) employing standardized orientation terminology (\eg, ``frontal alignment'') with explicit spatial frames of reference (\eg, egocentric, allocentric, and object-centric), examples, and task descriptions, and (3) ensuring difficulty progression from simple categorical judgments to precise angular measurements across all orientation dimensions.  This systematic approach isolates orientation from scene perception skills, minimizes confounding factors such as object recognition difficulty, scene clutter, linguistic ambiguity, and contextual distractions that plague existing benchmarks~\cite{cheng2025spatialrgpt, Chen_2024_CVPR,10.24963/ijcai.2024/701}.
Our extensive experiments with 26 state-of-the-art MLLMs on DORI reveal several key findings:
\begin{itemize}
    \item Models struggle most with tasks requiring spatial transformation 
    and perspective shifts: performance drops by $30\%$ on dynamic 
    rotational tasks relative to static pose identification, and by 
    $25\%$ on allocentric tasks requiring non-egocentric viewpoint 
    adaptation (\eg, determining if two objects face each other from 
    their own frames of reference).
    \item Architectural design and prompt tuning matter more than 
    scale: token-based integration (\eg, Mantis-Idfs-8B\cite{Jiang2024MANTISIM}) 
    consistently outperforms linear projection, and smaller 
    instruction-tuned models (\eg, DeepSeek-1.3B-Chat\cite{bi2024deepseek}) often surpass 
    larger base counterparts (\eg, DeepSeek-7B-Base\cite{bi2024deepseek}).
    \item LoRA finetuning on DORI generalizes broadly, boosting 
    performance by up to $27\%$ on standard spatial reasoning benchmarks including 
    BLINK~\cite{fu2024blink}, SAT~\cite{ray2024satspatialaptitudetraining}, 
    and 3DSRBench~\cite{stogiannidis2025mind}.
\end{itemize}
\section{Related Work}
 \begin{table*}[t]
\centering
\scriptsize
\setlength{\tabcolsep}{1.5pt}
\renewcommand{\arraystretch}{0.85}
\caption{
\textbf{Comparison with standard benchmarks for orientation-reasoning}. Asterisks (*) indicate counts refer only to orientation-related tasks, not the full dataset.
DORI is larger and/or more diverse than similar datasets.
\emph{C = Coarse, G = Granular;
View: A = Allocentric, E = Egocentric;
$\Omega$ = Open-vocabulary
}
}
    \resizebox{\textwidth}{!}{%
    \begin{tabular}{@{}lccccccc@{}}
        \toprule
        \textbf{Dataset} & \textbf{Obj.} & \textbf{Granul.} & \textbf{View} & \textbf{\#Sources} & \textbf{N-Imgs} & \textbf{S-Imgs} & \textbf{VQA} \\
        \midrule

        Spatial-MM~\cite{shiri2024empirical}              & 342  & C  & AE & 1  & 537        & --             & 0.6K   \\
        ScanQA~\cite{azuma_2022_CVPR}                     & 20   & C  & A  & 1  & 800        & --             & 7.4*K  \\
        BLINK~\cite{fu2024blink}                          & 71*  & C  & AE & 2  & 552*       & --             & 0.5*K  \\
        KiVA~\cite{yiu2025kiva}                           & 62*  & G  & A  & 2* & --         & 400*           & 0.6*K  \\
        EgoOrientBench~\cite{jung2024rightrightenhancingobject} & 197 & C & E & 5 & 6K    & 2.3K           & 33.4K  \\
        EmbSpatial-Bench~\cite{du-etal-2024-embspatial}   & 294  & C  & E  & 3  & 1.5K*      & 683*           & 2.4*K  \\
        CLEVR-3D~\cite{yan2023comprehensive}              & 160  & C  & A  & 2  & 1.0K       & --             & 2.3*K  \\
        3DSRBench~\cite{ma20243dsrbench}                  & $\Omega$ & C & A & 1* & 2.0K*   & --             & 2.7*K  \\
        PO3D-VQA~\cite{wang20233d}                        & 5    & C  & A  & 1  & --         & $\sim$30K      & 300*K  \\
        Mind the Gap~\cite{stogiannidis2025mind}           & 197  & C  & E  & 2* & $\sim$280* & $\sim$520*     & 1.2*K  \\
        Spatial457~\cite{wang2025spatial457}              & 5    & C  & A  & 1  & --         & 992*           & 4.5*K  \\
        {VSI-Bench~\cite{yang2024think}}                  & 47   & C  & A  & 3  & 288         & --         & 5K     \\
        {OmniSpatial~\cite{jia2026omnispatial}}           & $\Omega$ & C & AE & 9 & 8.4K       & --            & 8.4K   \\
        {VLM4D~\cite{zhou2025vlm4d}}                      & $\Omega$ & C & AE & 4 & 600       & 400             & 1.8K   \\
        {MMSI-Bench~\cite{yang2025mmsi}}                  & $\Omega$ & C & AE & 8 & 1.9K       & --             & 1K     \\
        \midrule
        \rowcolor{teal!25}\textbf{DORI (Ours)} & $\Omega$ & \textbf{CG} & \textbf{AE} & \textbf{14} & \textbf{5K} & \textbf{9K} & \textbf{33.6K} \\
        \bottomrule
    \end{tabular}}
    \vspace{-6mm}
    \label{tab:data_comparison}
\end{table*}

Understanding where an object is pointing, how it has rotated, or whether it sits in its expected pose are deceptively simple questions, yet MLLMs consistently struggle with them~\cite{tong2024cambrian, cheng2024spatialrgpt, 
Chen_2024_CVPR}. Tab.~\ref{tab:data_comparison} contextualizes DORI against prior spatial and orientation-related benchmarks. Existing datasets have made important progress in broader visual reasoning, but orientation is often treated as a secondary subset rather than the primary evaluation target. For example, BLINK~\cite{fu2024blink}, Spatial-MM~\cite{shiri2024empirical}, and EmbSpatial-Bench~\cite{du-etal-2024-embspatial} include orientation-related questions, but primarily emphasize coarse directional judgments and do not systematically probe rotational transformation, allocentric reasoning, or canonical pose understanding.


Tab.~\ref{tab:data_comparison} also highlights limitations in scale, granularity, and visual diversity. Most prior benchmarks support only coarse or categorical questions, making it difficult to distinguish failures in basic facing-direction perception from failures in precise angular reasoning. Synthetic datasets such as PO3D-VQA~\cite{wang20233d} and KiVA~\cite{yiu2025kiva} provide precise ground truth but contain limited or no natural images, while EgoOrientBench~\cite{jung2024rightrightenhancingobject} is restricted to egocentric frames and discrete orientation classes. Other benchmarks such as EmbSpatial-Bench~\cite{du-etal-2024-embspatial}, 3DSRBench~\cite{ma20243dsrbench}, OmniSpatial~\cite{jia2026omnispatial}, MMSI-Bench~\cite{yang2025mmsi}, and RoboSpatial~\cite{song2025robospatial} evaluate spatial reasoning more broadly but do not treat object orientation as a primary, multi-dimensional target. Pose estimation 
approaches~\cite{deng2025multimodalanimalposeestimation, 
10.1007/s10462-020-09888-5, 10.1145/3603618} offer precise 6-DoF outputs 
but require costly pipelines~\cite{Corsetti_2024_CVPR, Feng_2024_CVPR} 
and resist natural-language interpretation.


DORI addresses these limitations by evaluating object orientation as a primary task across complementary dimensions and difficulty levels. It combines open-vocabulary objects, both egocentric and allocentric coverage, and coarse-plus-granular assessment, a combination not jointly provided by prior benchmarks in Tab.~\ref{tab:data_comparison}. Drawing from 14 sources across natural and simulated environments, DORI yields 13,652 images and 33,656 questions, providing the diagnostic resolution needed to identify where and why orientation reasoning fails in current MLLMs.

 \begin{figure}[t]
 \centering
        \includegraphics[width=\linewidth]{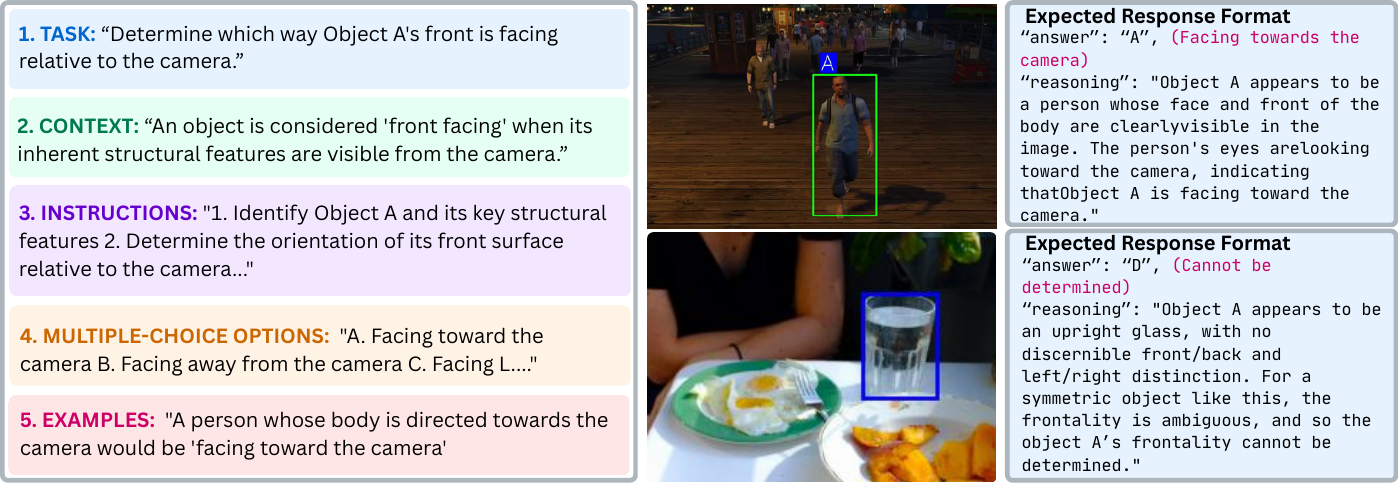}
\caption{Structured prompt design and example question–answer pairs from DORI. Each query follows a consistent format comprising a task description, contextual definition, step-by-step instructions, multiple-choice options, and illustrative examples, enabling systematic evaluation of orientation perception. All questions include the option “Cannot be determined,” maintaining a uniform answer space while explicitly modeling cases of frontality ambiguity or insufficient visual evidence.}
    \label{fig: prompt_strructure}
  \vspace{-4mm}
\end{figure}
\section{\textit{DORI}: Discriminative Orientation Reasoning Intelligence}
\nazia{
Object orientation understanding spans a wide spectrum of perceptual 
and cognitive demands - from static pose recognition and mental rotation 
to egocentric and allocentric spatial reasoning and physics-grounded 
semantic inference. Evaluating MLLMs across this full range requires 
a principled framework that can expose not just \emph{whether} models 
fail, but \emph{where} and \emph{why}.  For example, how closely do they match 
human-level orientation performance, do their reasoning patterns 
reflect those observed in human visuospatial cognition~\cite{blades1994development, 
vasilyeva2012development, spinelli1999hierarchical}, and what 
architectural or training factors drive their shortcomings? 
These questions jointly motivate DORI's task structure, data 
selection, and evaluation design which we will discuss further in the next section. 
 \begin{figure}[t]
 \centering
        \includegraphics[width=0.8\linewidth]{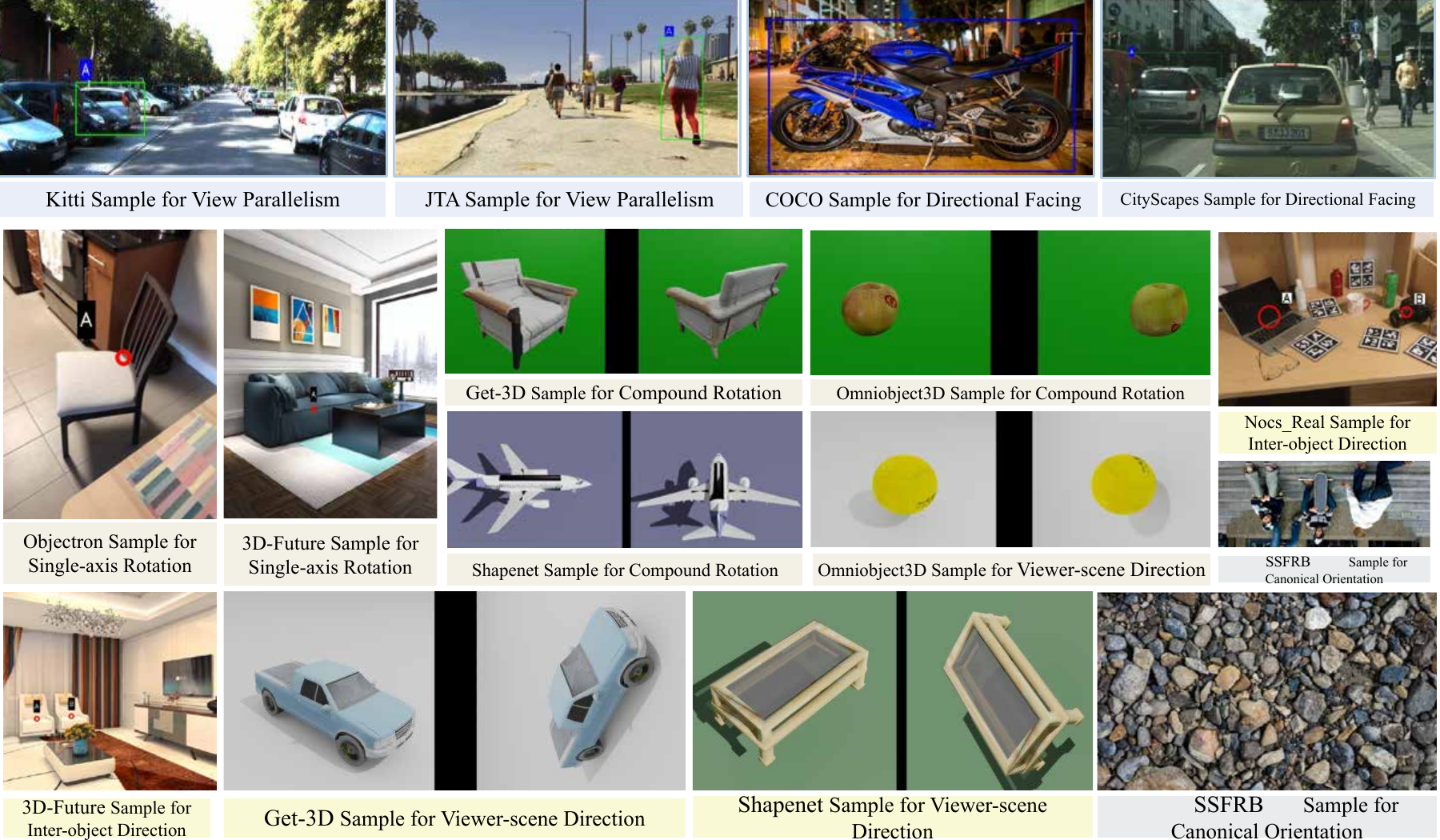}
        \vspace{-2mm}
    \caption{Representative samples from each dataset, including natural (Kitti, Cityscapes, Coco, SSFRB, etc.) and simulated (JTA, 3D-Future, Get-3D, etc.) sources.}
    \label{fig:dataset_samples}
  \vspace{-4mm}
\end{figure}

\subsection{Core Capabilities}
Drawing on established frameworks in cognitive neuroscience, DORI 
decomposes object orientation understanding into the four dimensions. \cref{tab:dimensions} shows how each axis corresponds to a 
separable perceptual or reasoning mechanism identified in the visuospatial 
literature as diagnostically meaningful~\cite{harris2024interpreting, 
frick2014development, 10.1002/hbm.20328, 10.3389/fpsyg.2015.00490}. 
We do not claim a strict developmental mapping.  Instead, this decomposition 
motivates \emph{why} orientation understanding must be evaluated along 
multiple targeted axes, and \emph{what} visual and cognitive 
processing types each axis demands. Below we describe how each dimension manifests as concrete evaluation tasks in DORI.
\smallskip

\noindent \textbf{\textcolor{col1}{1. Frontal Alignment}} evaluates the fundamental ability to perceive how an object's front-facing surface is oriented relative to the viewer - a prerequisite for any orientation-based reasoning. Humans rapidly identify which way an object is facing (\eg, deciding if a vehicle is approaching or departing) by recognizing structural and functional features such as faces, headlights, or entrances. However, we require additional reasoning steps for orientation interpretation~\cite{Kourtzi01022002}, such as assessing objects'  angular relationship with the viewing plane {defined as the imaginary plane onto which a scene is projected onto}. 

For MLLMs, we assess this frontal alignment capability through two complementary tasks: \textbf{view parallelism analysis}, which quantifies the degree to which an object's frontal surface deviates from being parallel to the image plane, providing angular measurements (\eg, is a chair directly facing the camera or at a 45-degree angle). \textbf{Directional facing perception} asks the cardinal direction an object's front is oriented relative to the camera position (\eg, whether a desk is facing toward, away, leftward, or rightward from the viewer's perspective). This dual assessment is supported by research that reports viewpoint-invariant recognition of frontal features operates through different neural mechanisms than precise angular estimation~\cite{harris2024interpreting}. Prior work also demonstrates MLLMs' inability to perceive object frontality (\eg, confusing left/right perspectives) even when provided with bounding boxes~\cite{tong2024eyes, shiri2024empirical}, often being on par with random predictions~\cite{yin2025multimodal}. {For objects whose frontality is difficult to define (\eg, symmetric objects like a ball), models are expected to respond with the "Cannot be determined" option, which is provided as a choice in every question (see example in \cref{fig: prompt_strructure}).} 
\smallskip


\noindent\textbf{\textcolor{col3}{2. Rotational Transformation}} examines the ability of an MLLM to comprehend orientation changes through rotation, reflecting requirements such as 
embodied object manipulation (\eg, fitting a key into a lock), and viewpoint-dependent navigation (\eg, reorienting a map for wayfinding). Mental rotation capabilities allow humans to predict how objects align when rotated~\cite{10.1027/1618-3169/a000505},  with neuroimaging evidence showing premotor and parietal activation during both physical \& imagined rotations~\cite{doganci2023embodied, 10.1027/1618-3169/a000505}. Both processes trigger similar neural activations~\cite{xue2017uncovering}, highlighting the inherent complexity of spatial transformation tasks that MLLMs must simulate.
Inspired by this, we design the rotational tasks in our benchmark to progress from simple to complex levels mirroring human cognitive processing demands~\cite{li2019simon,ter2010mental, xue2017uncovering}. Thus, the first subtask examines \textbf{single-axis rotation} \& evaluates basic angular transformations rotated along one spatial dimension (\eg, determining the shortest rotation to face a chair toward the camera). This establishes baseline capabilities before progressing to more cognitively demanding compound rotation~\cite{ter2010mental}, involving \textbf{multiple-axes rotation} (\eg, aligning objects through a sequence of horizontal/vertical rotations). These subtasks represent common scenarios in assembling products through item manipulation and real-world scene planning by embodied agents, where object orientation understanding directly impacts final task performance. 
\smallskip

 \noindent\textbf{\textcolor{col2}{3. Relative Orientation}}  examines the understanding of how objects are oriented in relation to each other and with respect to the viewer. Humans navigate a complex visual world by seamlessly tracking orientation changes across scenes and viewpoints, which is crucial for scene comprehension and geometric coordination. The human brain contains a specific interconnected region facilitating ``mental orientation'' -the ability to effectively spatially orient an object to different viewpoints and perspectives~\cite{doi:10.1073/pnas.1504242112, 10.5334/joc.321,gramann2010human, zaehle2007neural}. In contrast, studies have shown that MLLMs struggle significantly with questions posed from non-egocentric perspectives~\cite{shiri2024empirical,wang2023holoassist, yeh2025seeing} and with maintaining consistent dimensional relationships between multiple objects across time and viewpoints~\cite{yang2024think} We systematically probe this aspect of object's relative orientation via the following sub-tasks: (1) \textbf{inter-object directional relationships}, to assess the relative facing directions of objects (\eg, determine if two cars are facing the same or opposite directions), and (2) \textbf{image-pair rotational relationships}, to measure the ability to track orientation changes between two images (\eg, identify the degree of rotation between two views of the same object).

\subsection{Benchmark Creation Process}
\label{sec:benchmark}
DORI comprises seven orientation-reasoning tasks spanning the four 
dimensions above (Fig.~\ref{fig:intro_fig}), each evaluated at two tiers: 
\textbf{coarse-grained} questions for basic categorical judgments (\eg, 
``Are objects A and B facing the same direction?'' for relative orientation; 
``Has the object rotated between the two images?'' for rotational 
transformation) and \textbf{fine-grained} questions for precise 
quantitative estimation (\eg, ``How many degrees clockwise did the object 
rotate?''; ``What specific transformations would restore this image to its 
canonical orientation?''). This hierarchical design enables systematic 
evaluation from fundamental perception to advanced metric reasoning.
We generated the questions via two pipelines: converting existing 3D 
annotations into orientation questions, drawing from JTA~\cite{jta}, 
3D-Future~\cite{3d_future}, Get3D~\cite{get3d}, ShapeNet~\cite{shapenet}, 
OmniObject3D~\cite{wu2023omniobject3d}, NOCS REAL~\cite{wang2019normalized}, 
Objectron~\cite{ahmadyan2021objectron}, SSFRB~\cite{unsplash, delaunay, 
fabric_data, galaxy_zoo, rocks_dataset}, and the OmniNOCS~\cite{omninocs} 
subsets of KITTI~\cite{kitti} and Cityscapes~\cite{cityscapes}; and 
manually annotating real-world samples from COCO~\cite{mscoco}. 
Source selection is also task-specific: for instance, inter-object directional relationship 
questions require multi-object scenes, so single-object datasets like 
ShapeNet~\cite{shapenet} are excluded in favor of 3D-Future~\cite{3d_future} 
and NOCS REAL~\cite{wang2019normalized} (further details in 
Sec.~\ref{sec:statistics}). To handle orientation ambiguity in symmetric 
objects, every question includes a ``Cannot be determined'' option 
regardless of whether it is correct for that instance 
(Fig.~\ref{fig: prompt_strructure}). We describe the full curation details in the supplementary;
representative image samples are shown in Fig.~\ref{fig:dataset_samples}.

We designed the prompts to isolate orientation perception from 
confounding factors such as object recognition difficulty, scene clutter, 
linguistic ambiguity, and contextual distractions~\cite{castle2024using} 
(see supplementary for detailed discussion). 
The prompts follow a structured five-component format 
(Fig.~\ref{fig: prompt_strructure}): a \textbf{task description} specifying the 
orientation dimension being tested; contextual grounding of key concepts 
(\eg, ``An object is considered `front facing' when its inherent structural 
features are visible from the camera''); \textbf{step-by-step analysis instructions }
(\eg, ``1. Identify Object A and its key structural features. 2. Determine 
the orientation...''); \textbf{multiple-choice options}; and concrete reasoning 
\textbf{examples} (\eg, ``A person whose body is directed towards the camera would 
be `facing toward the camera.'\hspace{0pt}''). This design draws on 
instruction-tuning principles showing that explicit task framing and 
example-driven guidance improve model comprehension~\cite{hewing2024prompt}.
We have iteratively refined the prompts through multiple cycles of human feedback from non-expert annotators to address ambiguities, clarify terminology, and improve the task specificity~\cite{lin2025prompt}. For instance, early rotational 
transformation prompts yielded inconsistent axis interpretations, 
addressed by introducing concrete analogies (\eg, ``like a ballerina 
spinning clockwise'' to distinguish vertical-axis rotation from abstract 
directional descriptions). Response formats were standardized to require 
both an explicit answer and a chain-of-thought explanation, ensuring that 
performance differences reflect genuine orientation understanding rather 
than prompt interpretation variance. See supplementary for 
full prompt templates and examples across all question types.  
\smallskip

\noindent\textbf{Label reliability.} DORI's labels are primarily derived from structured 3D annotations, rotation matrices, and controlled rendering pipelines, which provide direct access to object orientation parameters. The main exception is the directional-facing subset from COCO, where orientation labels require manual judgment from real-world images. To assess ambiguity in this manually annotated subset, we conducted an agreement analysis with three
annotators and observed complete agreement in 96.7\% of cases. This suggests that the
manual labels are largely unambiguous, while the majority of DORI's labels are obtained
from deterministic geometric annotations rather than subjective judgment.

\subsection{DORI Statistics}
\label{sec:statistics}


\begin{figure}[t]
\centering
    \includegraphics[width=0.8\linewidth]{sec/assets/data_analysis/dual_donut_questions_category.png}
    \caption{\textbf{Dataset composition in DORI.} The left chart shows the distribution of examples across the seven orientation-reasoning tasks and natural/simulated image sources. The right chart shows the distribution of examples across broad object groups. Together, the plots summarize DORI's task, source, and object-group diversity.}
    \label{fig:question_sample_distribution}
\vspace{-6mm}
\end{figure}

Collectively, DORI encompasses $\mathbf{13,652}$ images spanning both natural (37\%) and simulated (63\%) environments, and includes $\mathbf{33,656}$ carefully constructed multiple-choice questions over diverse open-vocabulary objects across \textbf{14} computer vision datasets, including KITTI~\cite{kitti}, Cityscapes~\cite{cityscapes}, COCO~\cite{mscoco}, JTA~\cite{jta}, 3D-FUTURE~\cite{3d_future}, Objectron~\cite{ahmadyan2021objectron}, ShapeNet~\cite{shapenet}, and OmniObject3D~\cite{wu2023omniobject3d}, among others (full list in Sec.~\ref{sec:benchmark}). Fig.~\ref{fig:question_sample_distribution} illustrates the broad object-group and dataset distribution.  This multi-source approach allows us to balance complex, real-world environments (37\% of images) with controlled synthetic environments (63\%) where orientation parameters are precisely known. Knowing precise orientation parameters in synthetic data is crucial as it provides ground truth angular measurements with known accuracy, eliminating visual ambiguity or occlusion that might confound assessment. Meanwhile, real-world images introduce natural complexity and diversity while maintaining clear ground truth through expert annotation. Additional statistics on fine-grained and broad categories across tasks are available in the supplementary.

\section{Experiments} \label{sec:expts}

\begin{table*}[t]
\centering
\tiny
\setlength{\tabcolsep}{1.5pt}
\renewcommand{\arraystretch}{0.80}
\caption{{Performance of several open-source MLLMs on DORI}. Most models perform poorly, particularly on granular questions. These experiments reveal a systemic gap in object orientation understanding across all four dimensions studied in DORI. See supplementary for results with model variance. C: coarse, G: granular.}
\vspace{-2mm}
\begin{tabular}{l
  >{\columncolor{col1!12}}c
  >{\columncolor{col1!25}}c
  >{\columncolor{col1!12}}c
  >{\columncolor{col1!25}}c
  >{\columncolor{col3!12}}c
  >{\columncolor{col3!25}}c
  >{\columncolor{col3!12}}c
  >{\columncolor{col3!25}}c
  >{\columncolor{col2!12}}c
  >{\columncolor{col2!25}}c
  >{\columncolor{col2!12}}c
  >{\columncolor{col2!25}}c
  >{\columncolor{col4!12}}c
  >{\columncolor{col4!25}}c
  c c}
\toprule
 & \multicolumn{4}{c}{\cellcolor{col1!30}\begin{tabular}[c]{@{}c@{}}Frontal\\Alignment\end{tabular}}
 & \multicolumn{4}{c}{\cellcolor{col3!30}\begin{tabular}[c]{@{}c@{}}Rotational\\Transformation\end{tabular}}
 & \multicolumn{4}{c}{\cellcolor{col2!30}\begin{tabular}[c]{@{}c@{}}Relative\\Orient.\end{tabular}}
 & \multicolumn{2}{c}{\cellcolor{col4!30}\begin{tabular}[c]{@{}c@{}}Canonical\end{tabular}}
 & \multicolumn{2}{c}{} \\

 & \multicolumn{2}{c}{\cellcolor{col1!20}\begin{tabular}[c]{@{}c@{}}View\\Parallel.\end{tabular}}
 & \multicolumn{2}{c}{\cellcolor{col1!20}\begin{tabular}[c]{@{}c@{}}Dir.\\Facing\end{tabular}}
 & \multicolumn{2}{c}{\cellcolor{col3!20}\begin{tabular}[c]{@{}c@{}}Single-\\axis Rot.\end{tabular}}
 & \multicolumn{2}{c}{\cellcolor{col3!20}\begin{tabular}[c]{@{}c@{}}Compo-\\und Rot.\end{tabular}}
 & \multicolumn{2}{c}{\cellcolor{col2!20}\begin{tabular}[c]{@{}c@{}}Inter-\\Obj. Dir.\end{tabular}}
 & \multicolumn{2}{c}{\cellcolor{col2!20}\begin{tabular}[c]{@{}c@{}}Viewer-\\scene Dir.\end{tabular}}
 & \multicolumn{2}{c}{\cellcolor{col4!20}\begin{tabular}[c]{@{}c@{}}Orient.\end{tabular}}
 & \multicolumn{2}{c}{\begin{tabular}[c]{@{}c@{}}Avg.\end{tabular}} \\

\cmidrule(lr){2-5}\cmidrule(lr){6-9}\cmidrule(lr){10-13}\cmidrule(lr){14-15}\cmidrule(lr){16-17}
 & C & G & C & G & C & G & C & G & C & G & C & G & C & G & C & G \\
\midrule

Random                                                      & 35.7& 25.5& 20.2& 19.8& 20.7& 17.4& 20.4& 6.6& 15.0& 10.1& 33.1&20.3& 34.8& 16.0& 25.7& 16.5 \\
\hline
LLaVA-v1.6-\cite{liu2023llava} &55.9&26.8&22.6&19.8&23.4&16.5&32.8&\textbf{10.6}&7.2&12.4&46.6&20.1&34.7&11.3&31.9&16.7\\
LLaVA-v1.6-34B\cite{liu2023llava} &52.8&35.3&{32.6}&{26.4}&22.1&26.2&13.0&4.3&16.2&14.8&34.8&25.4&40.1&11.6&30.2&{20.5}\\
LLava-Next-8B\cite{liu2024llavanext}  & 33.2& 25.2& 21.3& 21.0& 20.5& 17.9& {40.6}& 6.3& 10.3& 12.0& 61.5& 25.2& 39.5& 9.2& 32.4& 14.4\\
Yi-VL-6B\cite{ai2024yi}       &38.9&31.0&25.0&25.2&{28.5}&14.7&29.6&3.0&15.2&12.3&41.4&10.8&32.8&14.4&30.2&15.9 \\
Yi-VL-34B\cite{ai2024yi}      &53.1&{35.1}&23.3&24.0&28.1&21.2&32.5&4.4&13.4&14.5&61.1&19.7&36.7&12.5&35.4&18.7\\

Mantis-CLIP\cite{Jiang2024MANTISIM}    &{60.1}&24.3&26.1&16.3&15.4&20.9&9.1&5.8&13.2&10.1&37.7&17.9&30.1&{34.5}&27.3&15.0 \\
Mantis-Idfs-8B\cite{Jiang2024MANTISIM} &57.8&33.0&22.5&12.7&25.7&{23.4}&25.9&6.6&17.6&9.0&55.4&24.5&\textbf{51.2}&\textbf{41.0}&{36.5}&17.5\\

DS-1.3B-Base\cite{bi2024deepseek}   & 58.1&24.8 & 15.0& 19.4& 21.1& 15.2& 17.0& 5.3& 17.0& 11.9& 61.3& 26.0& 1.8& 2.3& 27.3& 14.7\\
DS-1.3B-Chat\cite{bi2024deepseek}   & 58.1& 24.8& 22.6& 22.0& 21.2& 15.5& 33.9& 5.8& 15.3& 10.5& 47.6& 18.0& 20.8& 17.3& 31.3& 14.1\\  
DS-7B-Base\cite{bi2024deepseek}     & 26.3& 31.1& 14.6& 16.4& 18.3& 14.3& 35.5& 2.8& 3.6& 4.1& 35.7& 6.8& 28.3& 10.7& 23.1& 9.6\\
DS-7B-Chat\cite{bi2024deepseek}     & 59.4& 31.3& 31.3& 24.5& 28.2& 17.8& 35.8& 6.0& \textbf{20.2}& {14.9}& 18.5&{ 32.2}& 44.4& 32.2& 33.9& 18.6\\

Qwen2.5-3B-Inst.\cite{qwen2.5} & 56.3& 29.3& 23.4& 3.1& 18.2& 17.7& 25.4& 0.21& 16.3& 13.8& {62.8}& 16.9& 43.2& 11.4& 35.0& 13.2\\ 

Qwen3-8B-Inst.\cite{qwen3}  & 73.7& 45.6& 50.8& 54.8 & \textbf{32.9}& \textbf{45.9} & \textbf{49.0}& 7.5& 18.2& \textbf{21.9}& \textbf{79.1}& 37.2 & 42.0& 21.1 & \textbf{49.3} & 33.4 \\
Qwen3-32B-Inst \cite{qwen3} & \textbf{75.8}& \textbf{54.4}& \textbf{61.1}& \textbf{59.2}& 31.6&44.1& 39.6& 8.6 & 9.2& 20.5& 78.9& 46.1&28.1& 25.5& 46.3& \textbf{36.9}\\

Magma-8B\cite{yang2025magmafoundationmodelmultimodal}        & 51.5& 21.4& 25.4& 20.9 &20.6 &18.7 &32.0 &5.4 &16.3 &11.5 &32.6 &21.0 &37.5 &16.1 &30.8 &16.4 \\

RP-v1-13B\cite{yuan2024robopoint}       & 45.7&20.5&24.7&22.3&24.9&14.7&37.6& 5.6&12.9&12.7&61.2&20.9&39.8&10.6&35.2&15.3\\

IntVL-14B-Inst.\cite{wang2025internvl3_5}   &63.4&37.2&37.9&48.6&30.3&21.4&26.4&7.2&4.3&15.0&94.3&\textbf{44.1}&45.8&16.0 &43.2&27.1\\
IntVL-30B-A3B-Inst.\cite{wang2025internvl3_5}    &72.1&33.9&46.6&50.0&17.4&17.7&33.3&6.8&19.3&10.0&90.1&35.0&51.1&22.2&47.1&25.1\\
\bottomrule
\end{tabular}
\label{tab: openfull-models}
\vspace{-2mm}
\end{table*}

\begin{table*}[t]
\centering
\tiny
\setlength{\tabcolsep}{1.5pt}
\renewcommand{\arraystretch}{0.80}
\caption{Comparing open and closed-source MLLMs on a randomly selected 100-sample-per-task subset of DORI. Closed-source models show improved performance, but all models still perform poorly overall.}
\vspace{-2mm}
\begin{tabular}{l
  >{\columncolor{col1!12}}c
  >{\columncolor{col1!25}}c
  >{\columncolor{col1!12}}c
  >{\columncolor{col1!25}}c
  >{\columncolor{col3!12}}c
  >{\columncolor{col3!25}}c
  >{\columncolor{col3!12}}c
  >{\columncolor{col3!25}}c
  >{\columncolor{col2!12}}c
  >{\columncolor{col2!25}}c
  >{\columncolor{col2!12}}c
  >{\columncolor{col2!25}}c
  >{\columncolor{col4!12}}c
  >{\columncolor{col4!25}}c
  c c}
\toprule
 & \multicolumn{4}{c}{\cellcolor{col1!30}\begin{tabular}[c]{@{}c@{}}Frontal\\Alignment\end{tabular}}
 & \multicolumn{4}{c}{\cellcolor{col3!30}\begin{tabular}[c]{@{}c@{}}Rotational\\Transformation\end{tabular}}
 & \multicolumn{4}{c}{\cellcolor{col2!30}\begin{tabular}[c]{@{}c@{}}Relative\\Orient.\end{tabular}}
 & \multicolumn{2}{c}{\cellcolor{col4!30}\begin{tabular}[c]{@{}c@{}}Canonical\end{tabular}}
 & \multicolumn{2}{c}{} \\

 & \multicolumn{2}{c}{\cellcolor{col1!20}\begin{tabular}[c]{@{}c@{}}View\\Parallel.\end{tabular}}
 & \multicolumn{2}{c}{\cellcolor{col1!20}\begin{tabular}[c]{@{}c@{}}Dir.\\Facing\end{tabular}}
 & \multicolumn{2}{c}{\cellcolor{col3!20}\begin{tabular}[c]{@{}c@{}}Single-\\axis Rot.\end{tabular}}
 & \multicolumn{2}{c}{\cellcolor{col3!20}\begin{tabular}[c]{@{}c@{}}Compo-\\und Rot.\end{tabular}}
 & \multicolumn{2}{c}{\cellcolor{col2!20}\begin{tabular}[c]{@{}c@{}}Inter-\\Obj. Dir.\end{tabular}}
 & \multicolumn{2}{c}{\cellcolor{col2!20}\begin{tabular}[c]{@{}c@{}}Viewer-\\scene Dir.\end{tabular}}
 & \multicolumn{2}{c}{\cellcolor{col4!20}\begin{tabular}[c]{@{}c@{}}Orient.\end{tabular}}
 & \multicolumn{2}{c}{\begin{tabular}[c]{@{}c@{}}Avg.\end{tabular}} \\

\cmidrule(lr){2-5}\cmidrule(lr){6-9}\cmidrule(lr){10-13}\cmidrule(lr){14-15}\cmidrule(lr){16-17}
 & C & G & C & G & C & G & C & G & C & G & C & G & C & G & C & G \\
\midrule

LLaVA-v1.6-34B\cite{liu2024llavanext}                                                         & 45.5& 30.0& 22.5& 23.5& 15.0& 15.5& 19.3& 7.6& 15.0& 10.5& 37.3& 34.3& 66.0& 19.0& 31.5& 20.0\\
LLava-Next-8B\cite{liu2024llavanext}                                                       & 34.0& 13.0& 18.0& 19.0& 12.5& 17.5& 40.3& 7.0& 12.0& 13.5& 64.0& 25.0& 66.0& 16.0& 35.2& 15.8\\
DS-7B-Chat\cite{bi2024deepseek}                                                       & 52.5& 28.0& 24.5& 24.5& 21.0& 15.0& 35.3& 5.0& 24.0& 14.0& 23.6& 32.0& 67.0& 2.0& 35.4& 19.7\\
Qwen3-8B-Inst.\cite{qwen3}  & 61.5& 41.0& 38.5& 36.0& 23.5& 37.0& \textbf{69.6}& 9.3& 17.5& 13& 88.3& 47& 62.0& 26.0& 51.5& 29.9 \\
Qwen3-32B-Inst\cite{qwen3}  & 60.0& 39.5& 37.5& 33.5& 20.5& 37.5& 68.3& 10.3& 16.5& 13.5& 86.6& 44.6& 45.0& 26.0& 47.7& 29.2\\
IntVL-14B-Inst.\cite{wang2025internvl3_5} &61.5&32.0&32.5&42.5&29.0&16.0&31.3&8.3&6.5&14.0&\textbf{93.3}&45.6&71.0&25.0&46.4&26.2\\        
IntVL-30B-A3B-Inst.\cite{wang2025internvl3_5}                                            &67.5&34.0&40.5&39.0&18.0&11.0&35.0&8.0&26.0&13.0&89.0&33.6&79.0&33.0&50.7&24.5\\
\midrule
Gemini 1.5 Pro                                              & {68.5}& {53.0}& {43.5}&{ 39.0}& 24.0& 37.5& {65.3}& 12.3& {25.0}& 14.5& 91.0& {47.3}& {83.0}& 28.0& 57.1& 33.0 \\
Gemini 2.0 Flash                                            & 67.0& 35.0& 37.5& 33.5& 28.5& 27.0& 52.0& 14.3& 23.0& \textbf{17.0}& {92.0}& 43.6& 78.0& 29.0& 54.0 & 28.5 \\
{Gemini 3 Flash pre.}    & \textbf{68.5}& \textbf{71.0}& \textbf{63.0}&\textbf{ 63.0}& 59.0& 18.5& {37.5}& \textbf{63.7}& \textbf{44.0}& 8.0& 91.7& \textbf{57.0}& 86.0& \textbf{51.0}& \textbf{64.2}& 42.0 \\
Claude-sonnet-4.6       & {61.5}& {41.0}& {42.5}&{ 45.5}& 26.0& 35.5& {66.0}& {21.0}& {23.5}& 16.0& 82.7& {45.0}& {90.1}& 42.9& {56.0}& 35.3 \\


{GPT-4o}                                                 &44.5&45.0&30.5&35.5&29.5&34.0&39.3&{15.3}&23.5&13.5&88.3&53.3& 88.0&45.0& 49.0&34.5\\
GPT-4-1                                                      &48.5&41.0&39.0&{40.5}&{32.5}&{42.0}&31.6&14.0&22.5&8.5&87.6&44.6&\textbf{93.0}&46.0&50.6&33.8\\
{GPT-5-mini}                                 &61.5&55.5&55.0&{60.0}&\textbf{37.5}&\textbf{47.5}&46.3&22.6&19.0&10.5&88.0&57.3&86.0&{41.0}&56.1&\textbf{42.9}\\
{GPT-5.5-Pro}                                 &67.0&65.5&61.0&{65.0}&{25.5}&{34.5}&48.7&24.7&42.5&8.5&92.7&48.7&75.0&{47.0}&58.9&41.9\\

\bottomrule
\end{tabular}
\label{tab: subset-results}
\vspace{-6mm}
\end{table*}

\begin{table}[t]
\centering
\scriptsize
\setlength{\tabcolsep}{4pt}
\renewcommand{\arraystretch}{0.95}
\caption{Model rankings across MMSI-Bench, OmniSpatial, and DORI. 
Cell color encodes rank (green = high, red = low), superscripts encode rank. DORI's coarse (C) 
and granular (G) split reveals patterns invisible to single-score benchmarks.}
\vspace{-2mm}
\begin{tabular}{lcccc}
\toprule
\textbf{Model} & \textbf{MMSI-Bench} & \textbf{OmniSpatial} & \textbf{DORI (C)} & \textbf{DORI (G)} \\
\midrule
GPT-5-mini & \cellcolor{rank1}41.9$^{1}$ & -- & \cellcolor{rank1}\textbf{56.1}$^{1}$ & \cellcolor{rank1}\textbf{42.9}$^{1}$ \\
Gemini 2.0 Flash & -- & \cellcolor{rank4}44.0$^{4}$ & \cellcolor{rank2}54.0$^{2}$ & \cellcolor{rank4}28.5$^{4}$ \\
GPT-4.1 & \cellcolor{rank2}30.9$^{2}$ & \cellcolor{rank1}\textbf{51.8}$^{1}$ & \cellcolor{rank3}50.6$^{3}$ & \cellcolor{rank3}33.8$^{3}$ \\
InternVL3-\cite{wang2025internvl3_5} & \cellcolor{rank4}26.8$^{4}$ & \cellcolor{rank3}45.9$^{3}$ & \cellcolor{rank5}46.4$^{5}$ & \cellcolor{rank5}27.1$^{5}$ \\
RoboPoint-v1-13B\cite{yuan2024robopoint} & -- & \cellcolor{rank6}34.6$^{6}$ & \cellcolor{rank6}35.2$^{6}$ & \cellcolor{rank6}15.3$^{6}$ \\
Qwen2.5-VL-3B\cite{qwen2.5} & \cellcolor{rank5}26.5$^{5}$ & \cellcolor{rank5}40.3$^{5}$ & \cellcolor{rank7}35.0$^{7}$ & \cellcolor{rank7}13.2$^{7}$ \\
GPT-4o & \cellcolor{rank3}30.3$^{3}$ & \cellcolor{rank2}47.8$^{2}$ & \cellcolor{rank4}49.0$^{4}$ & \cellcolor{rank2}34.5$^{2}$ \\
\bottomrule
\end{tabular}
\label{tab: ranking_comparison}
\vspace{-4mm}
\end{table}

We evaluate {$26$} state-of-the-art multimodal models spanning diverse architectures, parameter scales, and pretraining methodologies across both open-source and proprietary systems. The models are as follows: LLaVA-v1.6-13B~\cite{Liu_2024_CVPR}, LLaVA-v1.6-34B~\cite{Liu_2024_CVPR}, LLava-Next-8B~\cite{liu2024llavanext}, Yi-VL-6B~\cite{ai2024yi}, Yi-VL-34B~\cite{ai2024yi}, Mantis-CLIP~\cite{Jiang2024MANTISIM}, Mantis-Idfs-8B~\cite{Jiang2024MANTISIM}, DS-1.3B-Base~\cite{bi2024deepseek}, DS-1.3B-Chat~\cite{bi2024deepseek}, DS-7B-Base~\cite{bi2024deepseek}, DS-7B-Chat~\cite{bi2024deepseek}, Qwen2.5-3B-Inst.~\cite{qwen2.5}, {Qwen3-8B-Inst.~\cite{qwen3}, Qwen3-32B-Inst.~\cite{qwen3}}, Magma-8B~\cite{yang2025magmafoundationmodelmultimodal}, RP-v1-13B~\cite{yuan2024robopoint}, {IntVL-14B-Inst~\cite{wang2025internvl3_5}, and IntVL-30B-A3B-Inst.~\cite{wang2025internvl3_5}}.
Proprietary models are queried via  their official APIs; open-source models are run on 2--4 NVIDIA RTX A6000  GPUs, with models exceeding 34B parameters evaluated on a single NVIDIA  A100-80G. All models are evaluated at temperature 0 for deterministic outputs. Answer options are shuffled randomly across runs to control for position bias, and responses are parsed via a standardized extraction pipeline that handles free-form text, letter-based, and JSON-formatted outputs. We report answer accuracy separately for coarse and granular question types across all task categories.

\noindent\textbf{Granular evaluation tolerance.} For granular questions, we use task-specific angular tolerances that reflect the resolution of each orientation judgment. The main results use a strict matching criterion to evaluate whether models recover the intended metric orientation rather than only the correct coarse category. In the supplementary, we additionally report a more tolerant evaluation setting. Model rankings are largely stable under this relaxed criterion, indicating that the observed failures are not solely due to small angular miscalibrations, but often reflect larger errors in orientation reasoning.

\subsection{Results} 
\cref{tab: openfull-models} benchmarks open-source MLLMs across all DORI question types, and reveals a fundamental limitation: many open-source MLLMs perform at or near random chance. Granular questions are especially challenging. Most models rely on coarse categorical judgments rather than precise angular reasoning. Even robotics-specialized models, such as, Magma-8B~\cite{yang2025magmafoundationmodelmultimodal} and RP-V1-13B~\cite{yuan2024robopoint} - fail to exceed general-purpose architectures, indicating that domain-specific pretraining alone is insufficient. Larger model families outperform smaller ones on coarse questions, but this advantage largely disappears on fine-grained tasks. 

\cref{tab: subset-results} extends this comparison to closed-source systems. On coarse accuracy, the leading proprietary models outperform the best open-source alternatives by up to 12.7 percentage points, yet several closed-source systems fail to consistently exceed top open-source models. On granular accuracy,  the picture is more mixed: GPT-5-mini and GPT-5.5-Pro show a more meaningful lead,  while Gemini 2.0 Flash remains near the best open-source level. Across both question types, all models exhibit severe deficits on rotational transformation and relative orientation. This confirms that robust geometric understanding remains absent even in state-of-the-art proprietary systems.
We attribute this to pretraining methodology. Most MLLMs employ CLIP-style contrastive objectives that optimize for semantic alignment rather than geometric structure~\cite{zhong2022regionclip}. Tong~\etal~\cite{tong2024eyes} similarly noted that pretraining creates a ``dimensional collapse'' in the embedding space, where continuous orientation variations become compressed into discrete semantic clusters (\eg, treating ``left'' and ``right'' as opposing categorical concepts rather than points along a continuous angular spectrum). While this may be slightly mitigated by using generative objectives~\cite{li2025exploringgenerativemllmsperceive}, MLLMs lack the necessary equivalent neural inductive biases utilized by humans~\cite{cognition-emerge-frontier-models, ling2009dissociation, delhaye2018neural, goble2012neural}. Instead, MLLMs approximate neural mechanisms through suboptimal attention patterns leading to hallucinations~\cite{huang-etal-2024-visual, yamada2024evaluating, deng2020self, yang2024improving}. 


\cref{tab: ranking_comparison} 
compares model rankings against MMSI-Bench~\cite{yang2025mmsi} and 
OmniSpatial~\cite{jia2026omnispatial} - two benchmarks covering general scene-level and 
multi-image spatial reasoning. 
In contrast, DORI isolates orientation, enabling a direct test of 
whether general spatial competence transfers to geometric reasoning. The ranking shifts are substantial: GPT-4o places 2nd on OmniSpatial and 3rd on MMSI-Bench, but drops on DORI's coarse orientation split, despite remaining competitive on granular questions.
This suggests that strong performance on broader spatial reasoning benchmarks does
not necessarily imply robust performance across all forms of object-centric orientation
reasoning, a blind spot that DORI is designed to expose.

\subsection{Analysis}

\noindent \textbf{Architecture critically determines orientation capacity.}
\cref{tab: openfull-models} also reveals clear architectural hierarchies among open-source models. Token-based fusion (\eg, Mantis-Idfs-8B\cite{Jiang2024MANTISIM}) consistently outperforms linear projection, preserving richer spatial information. Instruction-tuned variants universally surpass base models: DeepSeek-7B-Chat outperforms DeepSeek-7B-Base\cite{bi2024deepseek} by $\sim$47\% relative on average coarse accuracy. Strikingly, smaller tuned models can surpass larger base ones, e.g: DeepSeek-1.3B-Chat\cite{bi2024deepseek} exceeds DeepSeek-7B-Base\cite{bi2024deepseek} on coarse questions. This confirms that architectural design, pretraining data, and training objectives matter more than parameter count~\cite{ranasinghe2024learning}. The 41\% relative coarse gain from Qwen2.5-3B\cite{qwen2.5} to Qwen3-8B\cite{qwen3} further shows that advances in vision-language alignment directly benefit geometric understanding.
\begin{table*}[t]
\centering
\scriptsize
\setlength{\tabcolsep}{1.5pt}
\renewcommand{\arraystretch}{0.85}

\caption{Performance of Qwen2.5-3B-Inst\cite{qwen2.5} model finetuned using LoRA using Linear Projection, Linear Projection with a bottleneck, \& Token-based Fusion on DORI }
\vspace{-2mm}
\label{tab: fusion_token_table_performance}
\begin{tabular}{l
  >{\columncolor{col1!12}}c
  >{\columncolor{col1!25}}c
  >{\columncolor{col1!12}}c
  >{\columncolor{col1!25}}c
  >{\columncolor{col3!12}}c
  >{\columncolor{col3!25}}c
  >{\columncolor{col3!12}}c
  >{\columncolor{col3!25}}c
  >{\columncolor{col2!12}}c
  >{\columncolor{col2!25}}c
  >{\columncolor{col2!12}}c
  >{\columncolor{col2!25}}c
  >{\columncolor{col4!12}}c
  >{\columncolor{col4!25}}c
  c c}
\toprule
 & \multicolumn{4}{c}{\cellcolor{col1!30}\begin{tabular}[c]{@{}c@{}}Frontal\\Alignment\end{tabular}}
 & \multicolumn{4}{c}{\cellcolor{col3!30}\begin{tabular}[c]{@{}c@{}}Rotational\\Transformation\end{tabular}}
 & \multicolumn{4}{c}{\cellcolor{col2!30}\begin{tabular}[c]{@{}c@{}}Relative\\Orient.\end{tabular}}
 & \multicolumn{2}{c}{\cellcolor{col4!30}\begin{tabular}[c]{@{}c@{}}Canonical\end{tabular}}
 & \multicolumn{2}{c}{} \\

 & \multicolumn{2}{c}{\cellcolor{col1!20}\begin{tabular}[c]{@{}c@{}}View\\Parallel.\end{tabular}}
 & \multicolumn{2}{c}{\cellcolor{col1!20}\begin{tabular}[c]{@{}c@{}}Dir.\\Facing\end{tabular}}
 & \multicolumn{2}{c}{\cellcolor{col3!20}\begin{tabular}[c]{@{}c@{}}Single-\\axis Rot.\end{tabular}}
 & \multicolumn{2}{c}{\cellcolor{col3!20}\begin{tabular}[c]{@{}c@{}}Compo-\\und Rot.\end{tabular}}
 & \multicolumn{2}{c}{\cellcolor{col2!20}\begin{tabular}[c]{@{}c@{}}Inter-\\Obj. Dir.\end{tabular}}
 & \multicolumn{2}{c}{\cellcolor{col2!20}\begin{tabular}[c]{@{}c@{}}Viewer-\\scene Dir.\end{tabular}}
 & \multicolumn{2}{c}{\cellcolor{col4!20}\begin{tabular}[c]{@{}c@{}}Orient.\end{tabular}}
 & \multicolumn{2}{c}{\begin{tabular}[c]{@{}c@{}}Avg.\end{tabular}} \\

\cmidrule(lr){2-5}\cmidrule(lr){6-9}\cmidrule(lr){10-13}\cmidrule(lr){14-15}\cmidrule(lr){16-17}
 & C & G & C & G & C & G & C & G & C & G & C & G & C & G & C & G \\
\midrule
Linear Proj   &55.9&39.4&19.4&0.0&36.6&28.0&55.9&5.6&25.0&17.8&77.0&26.2&18.8&11.7&41.2&19.1\\
Linear Proj+Btlneck                                               &55.6&32.4&26.1&0.0&19.1&23.8&30.1&5.0&19.0&10.8&41.6&18.8&33.5&14.1&32.1&14.8\\
Token based fusion                                             &87.2&72.4&75.0&2.5&64.9&65.6&63.1&14.9&67.7&56.5&94.1&69.5&67.6&55.8&74.3&46.6\\
\bottomrule
\end{tabular}
\vspace{-2mm}
\end{table*}
 To further isolate the contribution of fusion strategy, we finetune  Qwen2.5-3B-Inst.\ with LoRA under three conditions for 2000 steps:  standard linear projection, linear projection with a bottleneck  (hidden dim 256, GELU activation), and the model's default token-based fusion. As shown in \cref{tab: fusion_token_table_performance},  token-based fusion substantially outperforms both projection variants across all tasks (74.3\% vs.\ 41.2\% average coarse accuracy),  corroborating the pattern observed in \cref{tab: openfull-models}. Adding a bottleneck to linear projection further degrades performance  (32.1\% average coarse), indicating that information compression at the fusion stage is particularly harmful for fine-grained orientation reasoning. We note, however, that this experiment carries a confounding variables:  Qwen was pretrained with token-based fusion, so the linear projection variants are not part of their native training regime. A fully controlled comparison would require retraining comparable MLLMs from scratch under each fusion strategy, which is beyond the scope of this work. We therefore interpret these results as suggestive rather than causal: within this adaptation setting, token-based fusion performs substantially better, indicating that fusion strategy is an important design axis that should be considered alongside model scale.
\begin{figure}[t]
\begin{minipage}[c]{0.48\linewidth}
\centering
\vspace{-2mm}
\includegraphics[width=.78\linewidth]{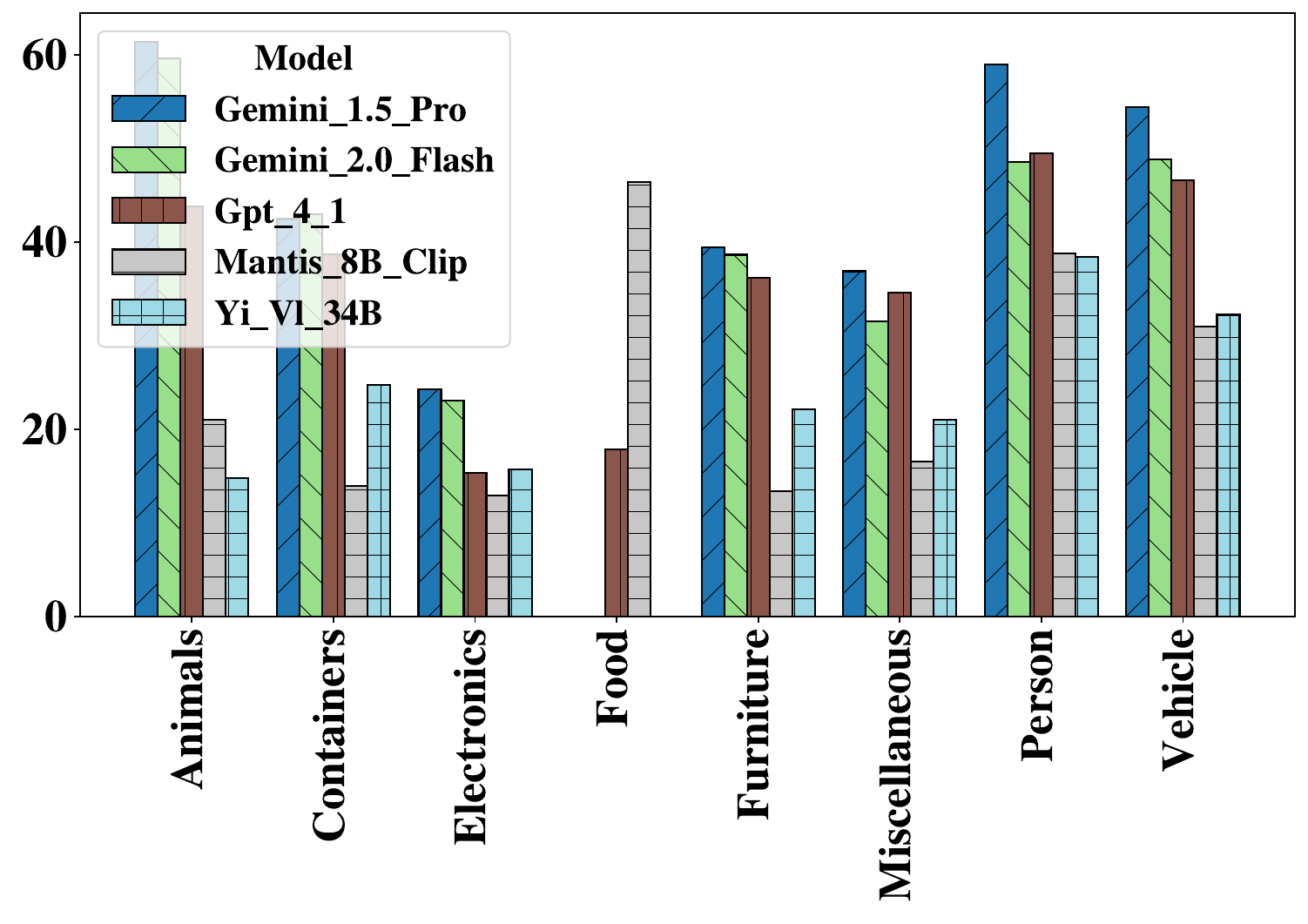}
\vspace{-3mm}
\caption{Performance of MLLMs by source category (additional models in supplementary).  Although for many categories the relative ranking of methods is relatively stable, in a few cases, like the food category, most models perform poorly.}
\label{fig:perf_by_category}
\end{minipage}%
\hfill
\begin{minipage}[c]{0.48\linewidth}
\centering
\includegraphics[width=\linewidth]{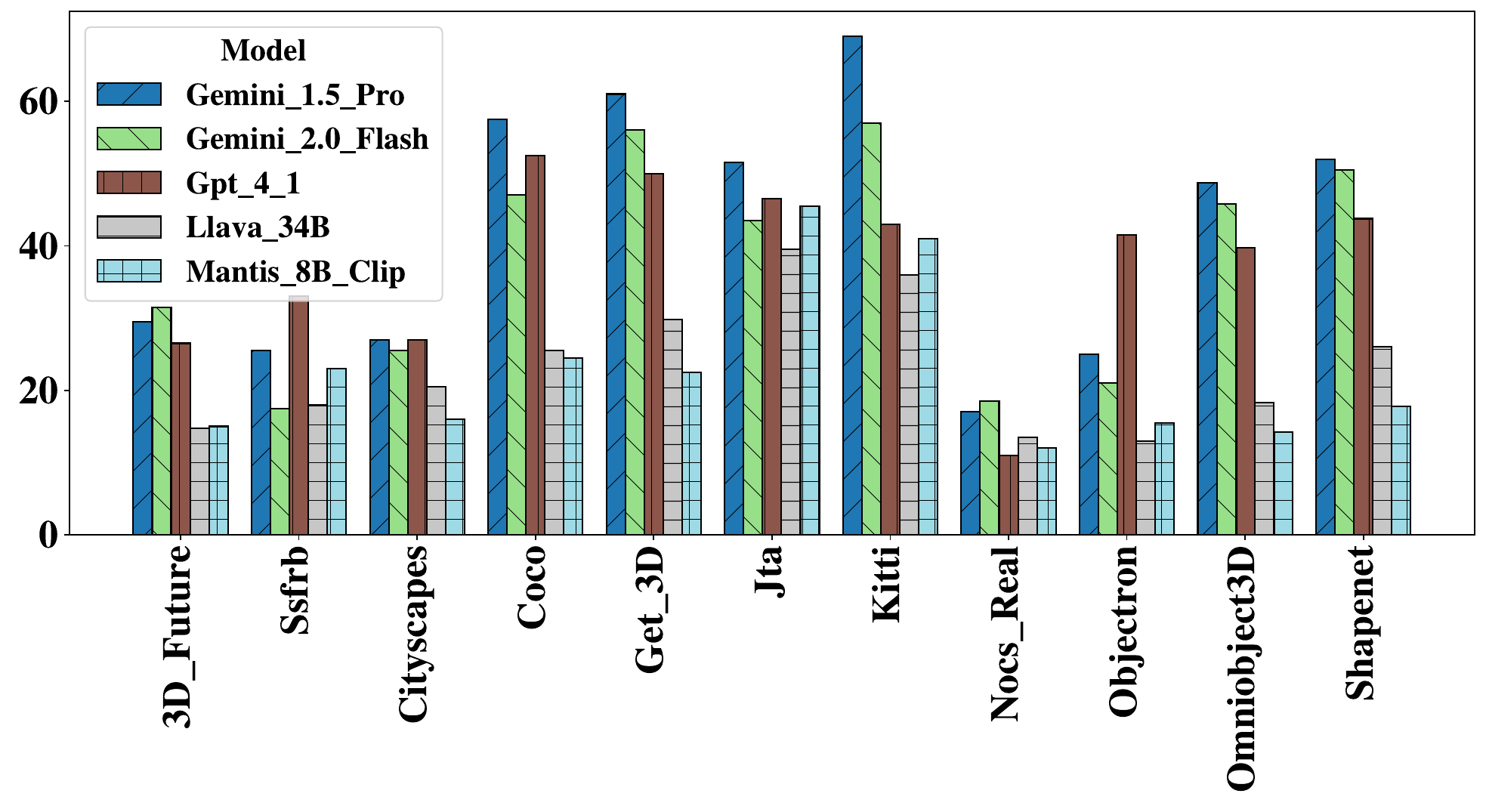}
\caption{Performance by source dataset. Models generalize better to 
simulated data and coarse questions, exposing a gap on natural images 
and fine-grained tasks (additional models in supplementary).}
\label{fig:perf_by_source}
\end{minipage}
\vspace{-4mm}
\end{figure}

\smallskip
\noindent \textbf{Real versus synthetic sources.}
Because DORI combines real and synthetic images, we separately examine performance
across source types. While performance varies across individual datasets and visual
domains, the aggregate model rankings are largely consistent between real and synthetic
subsets. Thus, our conclusions are not driven by a single data source or by synthetic
rendering artifacts alone. Instead, the mixed-source design allows DORI to combine
controlled orientation ground truth with the visual diversity of real-world scenes.
\smallskip

\noindent \textbf{Category-specific patterns expose reliance on semantic cues.}
Models perform substantially better on people and animals than on furniture or containers (Fig.~\ref{fig:perf_by_category}), that have clear front/back distinctions. This gap also reflects reliance on semantic shortcuts (\eg, recognizing faces) rather than a fundamental geometric understanding when determining object orientation. Fig.~\ref{fig:perf_by_source} reveals a complementary pattern: performance is 
markedly higher on simulated datasets than natural images, suggesting that 
real-world clutter and variability expose geometric limitations that controlled 
renders conceal. In addition, strong results on datasets like COCO raise contamination concerns. To address this, we followed  \cite{teterwak2024}, and separated into (likely) in-pretraining and out-of-pretraining sets, and find that models generally perform similarly on both sets. Thus, confirming negligible contamination effects.

\begin{table*}[t]
\centering
\scriptsize
\setlength{\tabcolsep}{1.5pt}
\renewcommand{\arraystretch}{0.85}
\caption{Comparison of a base Qwen2.5-VL-3B model and a LoRA-tuned variant trained on DORI, evaluated zero-shot on 3DSRBench~\cite{ma20243dsrbench}, Blink~\cite{blink}, and the SAT benchmark~\cite{ray2024satspatialaptitudetraining}. Finetuning on DORI yields up to a 27\% improvement over the base model. Note: the SAT Finetuning row reports results from the official SAT model repository, which uses a Qwen2.5-VL-7B backbone - substantially larger than the Qwen2.5-VL-3B models used in all other comparisons.}
\vspace{-2mm}
\begin{tabular}{lccc cccc c}
\toprule
& 
\multicolumn{3}{c}{\begin{tabular}[c]{@{}c@{}}Blink\end{tabular}} &
\multicolumn{4}{c}{\begin{tabular}[c]{@{}c@{}}3DSRBench\end{tabular}} & 
\multicolumn{1}{c}{\begin{tabular}[c]{@{}c@{}}SAT\end{tabular}} \\
\cmidrule(lr){2-4}\cmidrule(lr){5-8}\cmidrule(lr){9-9}
& 
\multicolumn{1}{c}{\begin{tabular}[c]{@{}c@{}}Multi-View\\Reasoning\end{tabular}} &
\multicolumn{1}{c}{\begin{tabular}[c]{@{}c@{}}Visual-\\Corresp.\end{tabular}} &
\multicolumn{1}{c}{\begin{tabular}[c]{@{}c@{}}Relative-\\Depth\end{tabular}} &
\multicolumn{1}{c}{\begin{tabular}[c]{@{}c@{}}Orient.\end{tabular}} &
\multicolumn{1}{c}{\begin{tabular}[c]{@{}c@{}}Multi-Obj.\\View To. Obj.\end{tabular}} &
\multicolumn{1}{c}{\begin{tabular}[c]{@{}c@{}}Multi-Obj.\\Parallel\end{tabular}} &
\multicolumn{1}{c}{\begin{tabular}[c]{@{}c@{}}Multi-Obj.\\Same Dir.\end{tabular}} &
\multicolumn{1}{c}{\begin{tabular}[c]{@{}c@{}}\end{tabular}} \\
\midrule
Base Model                                                           &42.9&25.3&64.1&32.5&11.4&11.4&46.8&51.3\\

SAT Finetuning
&40.6&66.2&66.9&33.7&24.4&50.0&50.4&50.8\\
\rowcolor{teal!20}DORI Finetuning                                                          &\textbf{45.1}&\textbf{26.5}&\textbf{65.3}&\textbf{38.6}&\textbf{14.0}&\textbf{38.1}&\textbf{50.3}&\textbf{63.3}\\
\bottomrule
\end{tabular}
\label{tab: external_datasets}
\end{table*}
\begin{table*}[t]
\centering
\scriptsize
\setlength{\tabcolsep}{1.5pt}
\renewcommand{\arraystretch}{0.85}
\caption{Performance of Base and Fine-tuned Qwen2.5-3B-Inst.\cite{qwen2.5} models across DORI Question Types}
\label{tab:lora_finetuning}
\setlength{\tabcolsep}{1.2pt}
\vspace{-4mm}
\begin{tabular}{lcccccccccccccccc}
\toprule
 & \multicolumn{4}{c}{\begin{tabular}[c]{@{}c@{}}Frontal\\Alignment\end{tabular}} &\multicolumn{4}{c}{\begin{tabular}[c]{@{}c@{}}Rotational\\Transformation\end{tabular}} & \multicolumn{4}{c}{\begin{tabular}[c]{@{}c@{}}Relative\\Orient.\end{tabular}} & \multicolumn{2}{c}{\begin{tabular}[]{@{}c@{}}\\Canonical\end{tabular}} & \multicolumn{2}{c}{} \\

& \multicolumn{2}{c}{\begin{tabular}[c]{@{}c@{}}View\\Parallel.\end{tabular}} & \multicolumn{2}{c}{\begin{tabular}[c]{@{}c@{}}Dir.\\Facing\end{tabular}} & \multicolumn{2}{c}{\begin{tabular}[c]{@{}c@{}}Single-\\axis Rot.\end{tabular}} & \multicolumn{2}{c}{\begin{tabular}[c]{@{}c@{}}Compo-\\und Rot.\end{tabular}} & \multicolumn{2}{c}{\begin{tabular}[c]{@{}c@{}}Inter-\\Obj.\ Dir.\end{tabular}} & \multicolumn{2}{c}{\begin{tabular}[c]{@{}c@{}}Viewer-\\scene Dir.\end{tabular}} & \multicolumn{2}{c}{\begin{tabular}[c]{@{}c@{}}Orient.\\\end{tabular}} & \multicolumn{2}{c}{\begin{tabular}[c]{@{}c@{}}\\Avg.\end{tabular}} \\
\cmidrule(lr){2-5}\cmidrule(lr){6-9}\cmidrule(lr){10-13}\cmidrule(lr){14-15}\cmidrule(lr){16-17}
                                                            & C & G & C & G & C & G & C & G & C & G & C & G & C & G & C & G \\
\midrule

Base Model                                         &57.7&34.8&27.2&0.0&22.9&20.4&37.7&5.6&7.3&13.7&77.7&22.0&47.0&20.2&39.6&16.6\\
SAT Fintuning                                                           &36.3&42.2&22.5&26.1&33.8&23.6&24.2&4.2&10.8&14.7&76.5&26.0&47.0&20.0&35.8&19.6\\

\rowcolor{teal!20}+ Finetuned                               &\textbf{87.8}&\textbf{72.8}&\textbf{81.9}&\textbf{2.6}&\textbf{76.8}&\textbf{76.1}&\textbf{64.5}&\textbf{17.3}&\textbf{79.2}&\textbf{60.6}&\textbf{99.0}&\textbf{84.9}&\textbf{67.6}&\textbf{37.1}&\textbf{79.5}&\textbf{50.2}\\
\hline
\end{tabular}
\vspace{-2mm}
\end{table*}
\smallskip
\noindent \textbf{Fine-tuning on DORI enables substantial transfer gains.}
Tab.~\ref{tab: external_datasets} measures zero-shot transfer to three external benchmarks (3DSRBench~\cite{ma20243dsrbench}, Blink~\cite{blink}, and SAT~\cite{ray2024satspatialaptitudetraining}) after fine-tuning on DORI. Fine-tuning Qwen2.5-VL-3B\cite{qwen2.5} on 27K DORI samples (real + synthetic) with LoRA~\cite{lora_low_rank} yields up to 27 percentage points absolute gain. \cref{tab:lora_finetuning} further reports performance on 7K held-out DORI samples, showing 40--34 percentage point absolute gains over the base model. 
Together, these results show that DORI finetuning generalizes across orientation and spatial reasoning tasks, capturing transferable orientation signals rather than prompt-specific patterns. However, transfer to non-orientation tasks is mixed, indicating orientation-centered rather than universal gains.
\begin{table*}[t]
\centering
\scriptsize
\setlength{\tabcolsep}{1.5pt}
\renewcommand{\arraystretch}{0.85}
\caption{Comparison of\textbf{ Human performance vs.\ MLLMs} on {50} randomly sampled questions of each type.  We find a significant performance gap exists between our expert annotators and MLLMs}
\vspace{-2mm}
\begin{tabular}{l
  >{\columncolor{col1!12}}c
  >{\columncolor{col1!25}}c
  >{\columncolor{col1!12}}c
  >{\columncolor{col1!25}}c
  >{\columncolor{col3!12}}c
  >{\columncolor{col3!25}}c
  >{\columncolor{col3!12}}c
  >{\columncolor{col3!25}}c
  >{\columncolor{col2!12}}c
  >{\columncolor{col2!25}}c
  >{\columncolor{col2!12}}c
  >{\columncolor{col2!25}}c
  >{\columncolor{col4!12}}c
  >{\columncolor{col4!25}}c
  c c}
\toprule
 & \multicolumn{4}{c}{\cellcolor{col1!30}\begin{tabular}[c]{@{}c@{}}Frontal\\Alignment\end{tabular}}
 & \multicolumn{4}{c}{\cellcolor{col3!30}\begin{tabular}[c]{@{}c@{}}Rotational\\Transformation\end{tabular}}
 & \multicolumn{4}{c}{\cellcolor{col2!30}\begin{tabular}[c]{@{}c@{}}Relative\\Orient.\end{tabular}}
 & \multicolumn{2}{c}{\cellcolor{col4!30}\begin{tabular}[c]{@{}c@{}}Canonical\end{tabular}}
 & \multicolumn{2}{c}{} \\

 & \multicolumn{2}{c}{\cellcolor{col1!20}\begin{tabular}[c]{@{}c@{}}View\\Parallel.\end{tabular}}
 & \multicolumn{2}{c}{\cellcolor{col1!20}\begin{tabular}[c]{@{}c@{}}Dir.\\Facing\end{tabular}}
 & \multicolumn{2}{c}{\cellcolor{col3!20}\begin{tabular}[c]{@{}c@{}}Single-\\axis Rot.\end{tabular}}
 & \multicolumn{2}{c}{\cellcolor{col3!20}\begin{tabular}[c]{@{}c@{}}Compo-\\und Rot.\end{tabular}}
 & \multicolumn{2}{c}{\cellcolor{col2!20}\begin{tabular}[c]{@{}c@{}}Inter-\\Obj. Dir.\end{tabular}}
 & \multicolumn{2}{c}{\cellcolor{col2!20}\begin{tabular}[c]{@{}c@{}}Viewer-\\scene Dir.\end{tabular}}
 & \multicolumn{2}{c}{\cellcolor{col4!20}\begin{tabular}[c]{@{}c@{}}Orient.\end{tabular}}
 & \multicolumn{2}{c}{\begin{tabular}[c]{@{}c@{}}Avg.\end{tabular}} \\

\cmidrule(lr){2-5}\cmidrule(lr){6-9}\cmidrule(lr){10-13}\cmidrule(lr){14-15}\cmidrule(lr){16-17}
 & C & G & C & G & C & G & C & G & C & G & C & G & C & G & C & G \\
\midrule

 GPT-4o                                                       &22.0&36.0&38.0&6.0&36.0&34.0&76.0&66.6&38.0&48.0&88.0&70.6&70.0&20.0&52.5&40.1\\
Gemini 1.5 Pro                                                        &64.0&72.0&76.0&80.0&56.0&54.0&86.0&66.0&64.0&62.0&98.0&80.0&50.0&64.0&70.5&68.2\\
 Human                                                       &82.0&92.0&92.0&82.0&80.0&78.0&86.0&86.0&86.0&74.0&98.0&90.0&92.0&92.0&88.0&84.8\\
\bottomrule
\end{tabular}
\label{tab: human-eval}
\end{table*}
\smallskip
\noindent \textbf{Human evaluation exposes a performance ceiling.}
Tab.~\ref{tab: human-eval} compares expert human performance against the best closed-source model across all task types. We recruited 14 experts to evaluate 50 examples per task type ($700$ samples total), each with three answer options: the correct answer, ``Cannot be determined,'' and a random distractor. Humans achieved up to 98\% accuracy, validating annotation quality. The best closed-source model trails by roughly 17 percentage points, revealing substantial headroom for MLLM improvement.
\smallskip

\noindent \textbf{Prompt structure scaffolds orientation understanding.}
We ablated key prompt components: formatting, examples, task descriptions, and step-by-step reasoning (\cref{fig: prompt_strructure}), across 8 models. \cref{tab: prompt_ablate} compares performance without structured prompting. Most models degrade sharply without structure: LLaVA-v1.6-13B \cite{liu2023llava} collapses to near-zero on Direction Facing granular, Compound Rotation granular, and Canonical Orientation granular, while LLaVA-Next-8B\cite{liu2023llava} suffers a 93\% relative decline on View Parallelism granular. For these architectures, structured prompts yield up to 50--100$\times$ improvements on fine-grained tasks. Prompt sensitivity is not uniform, however. Mantis-Idfs-8B\cite{Jiang2024MANTISIM} shows counterintuitive \textit{gains} under ablation: Compound Rotation coarse surges 141\% relative, suggesting rigid templates interfere with its interleaved architecture's native multi-image reasoning. DS-7B-Chat\cite{bi2024deepseek} is similarly non-monotonic: Inter-Object Direction coarse improves 145\% relative, yet Compound Rotation coarse collapses 97\%. 
These divergences show that prompts can disambiguate some spatial transformations while disrupting others. Notably, Canonical Orientation remains difficult across all models, indicating that prompt engineering alone cannot replace geometric priors.

\section{Conclusion}
\begin{table*}[t]
\centering
\scriptsize
\setlength{\tabcolsep}{1.5pt}
\renewcommand{\arraystretch}{0.85}
\caption{Models perform much worse for most of the tasks when the prompts are unstructured and have no clarification examples. The degradation is particularly prominent for Compound rotation and Canonical orientation related tasks}
\vspace{-2mm}
\label{tab: prompt_ablate}
\begin{tabular}{l
  >{\columncolor{col1!12}}c
  >{\columncolor{col1!25}}c
  >{\columncolor{col1!12}}c
  >{\columncolor{col1!25}}c
  >{\columncolor{col3!12}}c
  >{\columncolor{col3!25}}c
  >{\columncolor{col3!12}}c
  >{\columncolor{col3!25}}c
  >{\columncolor{col2!12}}c
  >{\columncolor{col2!25}}c
  >{\columncolor{col2!12}}c
  >{\columncolor{col2!25}}c
  >{\columncolor{col4!12}}c
  >{\columncolor{col4!25}}c
  c c}
\toprule
 & \multicolumn{4}{c}{\cellcolor{col1!30}\begin{tabular}[c]{@{}c@{}}Frontal\\Alignment\end{tabular}}
 & \multicolumn{4}{c}{\cellcolor{col3!30}\begin{tabular}[c]{@{}c@{}}Rotational\\Transformation\end{tabular}}
 & \multicolumn{4}{c}{\cellcolor{col2!30}\begin{tabular}[c]{@{}c@{}}Relative\\Orient.\end{tabular}}
 & \multicolumn{2}{c}{\cellcolor{col4!30}\begin{tabular}[c]{@{}c@{}}Canonical\end{tabular}}
 & \multicolumn{2}{c}{} \\

 & \multicolumn{2}{c}{\cellcolor{col1!20}\begin{tabular}[c]{@{}c@{}}View\\Parallel.\end{tabular}}
 & \multicolumn{2}{c}{\cellcolor{col1!20}\begin{tabular}[c]{@{}c@{}}Dir.\\Facing\end{tabular}}
 & \multicolumn{2}{c}{\cellcolor{col3!20}\begin{tabular}[c]{@{}c@{}}Single-\\axis Rot.\end{tabular}}
 & \multicolumn{2}{c}{\cellcolor{col3!20}\begin{tabular}[c]{@{}c@{}}Compo-\\und Rot.\end{tabular}}
 & \multicolumn{2}{c}{\cellcolor{col2!20}\begin{tabular}[c]{@{}c@{}}Inter-\\Obj. Dir.\end{tabular}}
 & \multicolumn{2}{c}{\cellcolor{col2!20}\begin{tabular}[c]{@{}c@{}}Viewer-\\scene Dir.\end{tabular}}
 & \multicolumn{2}{c}{\cellcolor{col4!20}\begin{tabular}[c]{@{}c@{}}Orient.\end{tabular}}
 & \multicolumn{2}{c}{\begin{tabular}[c]{@{}c@{}}Avg.\end{tabular}} \\

\cmidrule(lr){2-5}\cmidrule(lr){6-9}\cmidrule(lr){10-13}\cmidrule(lr){14-15}\cmidrule(lr){16-17}
 & C & G & C & G & C & G & C & G & C & G & C & G & C & G & C & G \\
\midrule
LLaVA-v1.6-13B&57.9&33.0&25.4&0.01&20.2&16.5&27.7&0.01&16.7&12.2&60.4&19.8&43.1&0.0& 35.9& 11.6 \\
Yi-VL-6B          &52.4&37.8&23.9&20.3&36.8&15.9&19.7&21.6&11.0&16.5&78.6&18.6&22.3&12.9& 34.9& 20.5\\
Mantis-CLIP                     &40.5&9.1&6.9&6.8&6.2&2.6&0.9&1.8&0.2&0.0&57.9&1.9&41.6&46.7&22.0&6.7\\
Mantis-Idfs-8B               &51.2&29.9&29.7&13.6&15.3&10.5&62.5&5.0&5.2&18.4&85.9&32.1&32.6&46.1&40.3&19.9\\
LLava-Next-8B  & 56.9& 1.8& 25.5& 29.9& 22.2& 11.7& 34.1& 6.6& 24.3& 10.0& 66.0&26.4& 38.7& 11.0& 38.2& 13.9\\
DS-1.3B-Chat & 40.7& 30.2& 22.6& 21.8& 12.2& 19.8& 35.6& 5.6& 16.2& 14.2& 31.1&21.6& 14.8& 9.8& 24.7& 17.5 \\
DS-7B-Base   & 45.2& 35.0& 21.6& 21.2& 25.4& 18.0& 2.8& 5.7& 15.0& 11.7& 33.6& 25.7& 3.6& 16.5& 21.0& 19.1 \\
DS-7B-Chat  & 48.1& 32.4& 27.9& 25.7& 47.0& 19.5& 1.0& 5.1& 49.6& 15.8& 36.6& 22.7& 17.3& 10.7& 32.5& 18.8\\
\bottomrule
\end{tabular}
\vspace{-4mm}
\end{table*}

We introduce DORI, a cognitively informed benchmark isolating orientation understanding in MLLMs. Across 26 models, a consistent pattern emerges: systems competent on general spatial benchmarks struggle severely on orientation tasks, trailing expert humans by roughly 18 percentage points. DORI's coarse-to-granular design reveals why models rely on categorical heuristics rather than continuous geometric reasoning, a limitation that broad benchmarks cannot detect. Canonical Orientation is particularly revealing: while some models improve on coarse canonical judgments, granular canonical reasoning remains difficult and highly inconsistent, suggesting that robust orientation understanding likely requires stronger architectural or pretraining-level geometric priors rather than scaling alone. Encouragingly, LoRA fine-tuning on DORI transfers broadly to held-out spatial benchmarks, showing that targeted orientation training yields genuine geometric gains. Further discussions, error analysis, and additional experiments are provided in the supplementary.

\begingroup
  \let\clearpage\relax
  \let\newpage\relax
  \renewcommand{\contentsname}{}
  \tableofcontents
\endgroup
\clearpage
\appendix
\section{Benchmark Construction and Statistics}
\subsection{Task-wise Dataset Creation}
\label{app: dataset_creation}
\textbf{View parallelism perception.} Evaluates a model's ability to determine whether an object's front-facing surface is oriented toward, away from, or perpendicular to the camera plane. We constructed this dataset using images from the JTA~\cite{jta} and KITTI\cite{kitti} datasets (specifically the subset used in OmniNOCS \cite{omninocs}). For JTA, which contains 3D human pose annotations, we calculated orientation by analyzing shoulder positions relative to the camera and head angle to precisely determine facing direction. For KITTI, we leveraged the available rotation matrices to categorize vehicles and pedestrians based on their orientation relative to the camera. This task is critical for fundamental scene understanding, where determining which objects are facing an agent is essential for interaction and navigation decisions.
\smallskip

\noindent\textbf{Frontal alignment perception.}  Extends orientation assessment to cardinal directions, requiring models to identify if objects are facing toward, away, leftward, or rightward relative to the camera. We developed this dataset using images from COCO \cite{mscoco} and Cityscapes \cite{cityscapes} (from the OmniNOCS~\cite{omninocs} subset). For COCO images, which lack orientation annotations, we conducted expert manual labeling of object orientations. For Cityscapes, we utilized rotational matrices to precisely determine directional orientation, limiting images to contain at most three objects to ensure assessment clarity. This directional understanding is vital for spatial navigation and object manipulation tasks where agents must understand not just if objects face them, but their specific directional orientation.
\smallskip

\noindent\textbf{Single-axis rotation.} Assesses understanding of rotational transformations around a vertical axis by asking models to determine the optimal rotation direction and precise angular adjustment needed for objects to face the camera. We constructed this dataset using 3D-Future \cite{3d_future}, which provides high-resolution 3D furniture models with known 6-DoF parameters. We focused primarily on chair variants with distinctive front/back features, calculating the exact rotational adjustment needed for the object to face the camera directly. This capability forms the foundation for computational manipulation planning and scene reconfiguration understanding. Furthermore, we utilized the Objectron~\cite{ahmadyan2021objectron} subset of the OmniNOCS~\cite{omninocs} dataset. This also included 6 DoF information which we used to determine the angle of the object with respect to the camera, for this dataset we utilized bikes, chairs and bottles. 
\smallskip

\noindent\textbf{Compound rotation.} Evaluates comprehension of complex rotations involving sequential transformations around multiple axes, where the rotation order impacts the final orientation. We developed this dataset using 3D-rendered objects from Get3D \cite{get3d}, ShapeNet \cite{shapenet}, and OmniObject3D,~\cite{wu2023omniobject3d} implementing a controlled rendering pipeline in Blender. For each object, we rendered an initial third-person view, then applied precise rotations along horizontal and vertical axes in varying sequences, rendering the transformed state. This task assesses the sophisticated mental rotation capabilities required for complex object manipulation and orientation reasoning across multiple dimensions.
\smallskip

\noindent\textbf{Inter-object direction perception.} Evaluates understanding of relative orientation between multiple objects from their own perspectives rather than the camera view. Using the 3D Future \cite{3d_future} and NOCS REAL~\cite{wang2019normalized} datasets, we leveraged 6 DoF parameters to calculate precise angular relationships between object pairs. The task requires models to determine if objects face the same direction, opposite directions, or have perpendicular orientations, progressing to granular assessment of the exact rotation needed for objects to align. This capability is essential for understanding agent-object and object-object relationships in complex scenes, particularly for collaborative robotic tasks or scene arrangement planning.
\smallskip

\noindent\textbf{Viewer-scene direction perception.} Evaluates perception of rotational changes between two images of the same object. Using Get3D~\cite{get3d}, ShapeNet~\cite{shapenet}, and OmniObject3D~\cite{wu2023omniobject3d} datasets, we rendered objects with a ground plane reference, then created corresponding images with the object rotated by specific angles around a vertical axis. Models must determine whether rotation occurred and, at the granular level, specify the exact degree of rotation. This assessment examines the ability to track orientation changes across views - a crucial capability for video understanding, temporal reasoning, and object tracking applications.
\smallskip

\noindent\textbf{Canonical orientation reasoning.} Evaluates models' understanding of normal object orientations and their ability to identify when objects appear in non-canonical positions. Using a subset of COCO\cite{mscoco} images with clear orientation expectations (\eg, people standing upright, vehicles with wheels on the ground), we created variations with systematic flips and rotations. Models must first identify whether images appear in their canonical orientation, then determine the specific transformations (rotation, flipping, or both) needed to restore proper orientation. This capability assesses world knowledge about typical object positioning, which is critical for anomaly detection, image correction, and understanding intentional vs. unintentional orientation deviations.

\begin{figure}[t]
    \centering
    \includegraphics[width=\linewidth]{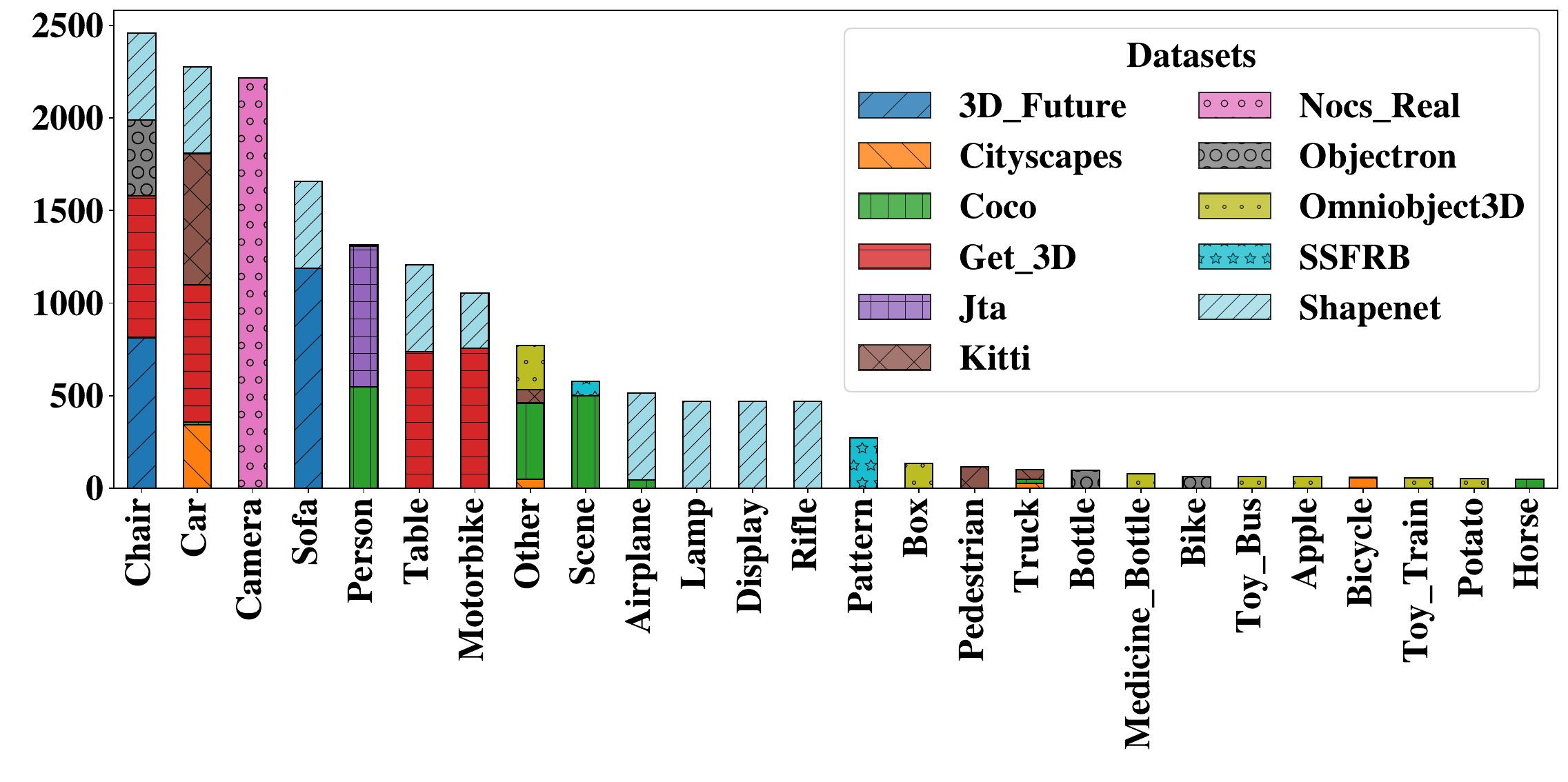}
    \caption{Sample category distribution of top 25 object categories showing their sources}    \label{fig:sample_cat_distrib}    
\end{figure}

\subsection{Compositional Diversity}
\label{sec:cat_dist}
Aforementioned 14 datasets span various natural and simulated domains, capturing varied object instances, backgrounds, occlusion levels, lighting conditions, and viewpoints. Such breadth of distribution ensures that the evaluation probes both the generalization ability of MLLMs and their robustness to contextual, visual, and category shifts across domains. 

\cref{fig:sample_cat_distrib} further complements this by presenting the distribution of the top 25 most frequent object categories across the datasets. While a few object classes, such as \textit{car}, \textit{person}, \textit{camera}, and \textit{chair} dominate in frequency, the distribution spans a broad range of object types and source domains. This heterogeneity plays a critical role in evaluating the performance of models not only on popular categories but also on long-tail classes, thereby encouraging more balanced and comprehensive assessment.

\cref{fig:sample_cat_distrib_with_ques} presents the distribution of the top 25 most frequent object categories in DORI, annotated with their corresponding orientation task types. This visualization highlights not only which categories are most prevalent, but also how they are utilized across the various tasks in DORI. 
While commonly occurring objects such as \textit{chair}, \textit{car}, and \textit{sofa} span several task types, others are more narrowly concentrated. This overlap and separation across tasks reflect the intentional diversity of DORI, designed to evaluate model capabilities across both general-purpose and category-specific orientation challenges. The distribution also underscores the need for models to generalize effectively across unevenly represented categories and task combinations.
\begin{figure}[t]
    \centering
    \includegraphics[width=\linewidth]{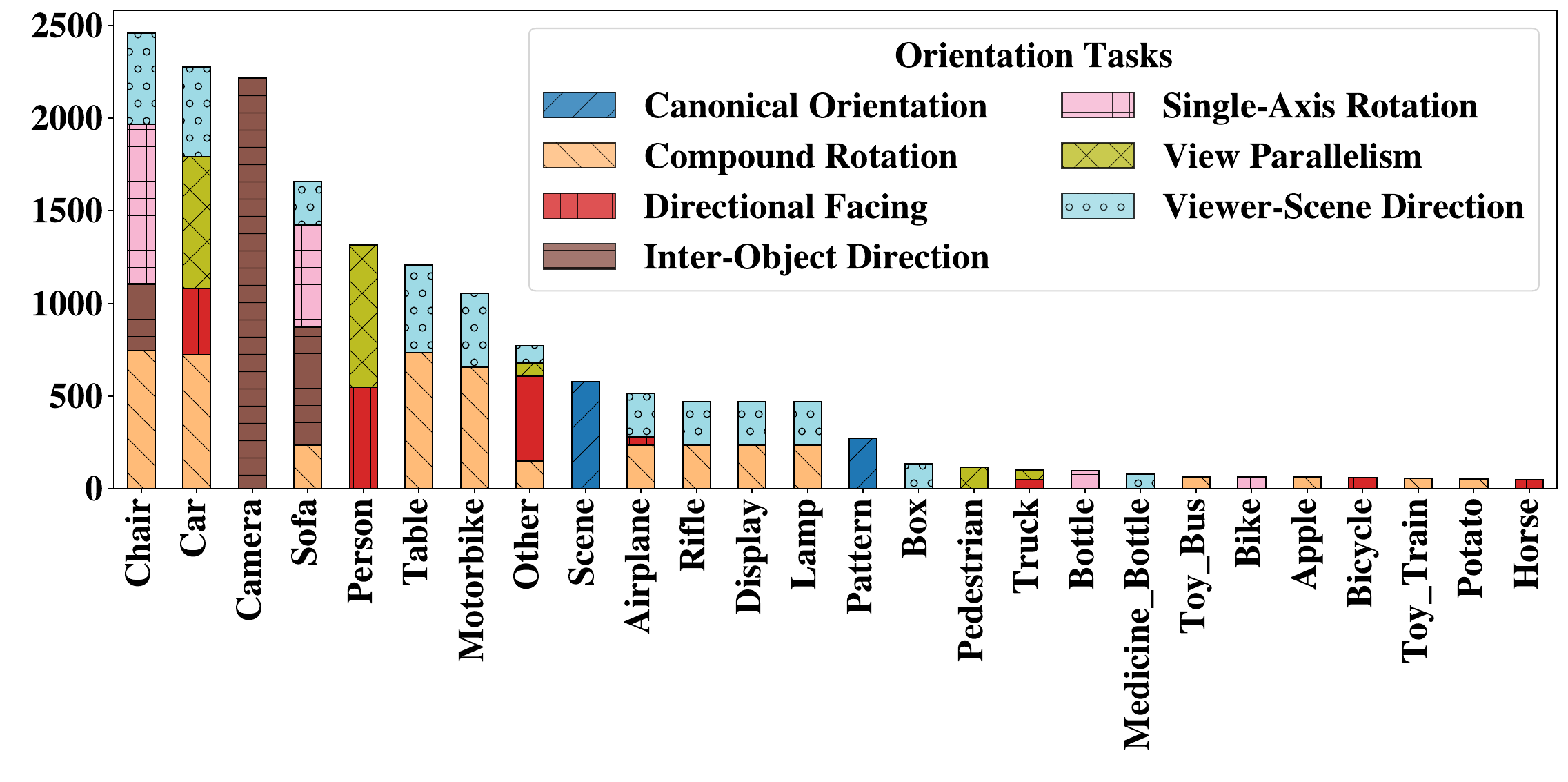}
    \caption{Sample category distribution of top 25 object categories showing their associated orientation tasks}
    \label{fig:sample_cat_distrib_with_ques}    

\end{figure}
\subsection{Mapping Object Categories to Broad Classes}
\begin{figure}[t]
    \centering
    \includegraphics[width=0.8\linewidth]{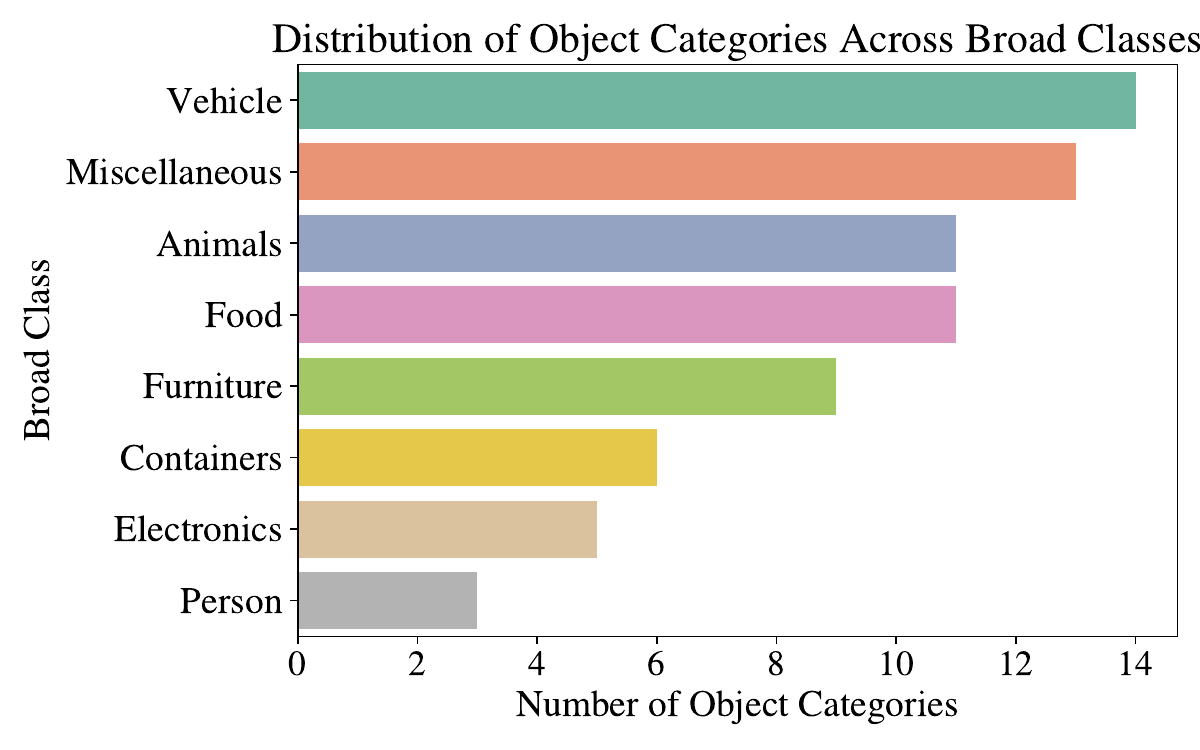}
    \caption{Distribution of fine-grained object categories mapped to broader semantic classes. Most categories fall under \textit{Vehicle}, \textit{Miscellaneous}, and \textit{Animals}, reflecting common object types in the evaluated datasets.}    
    \label{fig:obj_class_distribution}    
\end{figure}
\begin{table}[h]
\centering
\caption{Mapping from object categories to broad semantic classes.}
\vspace{1em}
\label{tab:category_mapping}
\begin{tabular}{ll}
\toprule
\textbf{Object Category} & \textbf{Broad Class} \\
\midrule
car, truck, tram, van, bus, train, airplane, boat, & Vehicle \\
 motorcycle, bicycle,  coach, bike, motorbike, trailer& Vehicle \\
pedestrian, cyclist, person & Person \\
zebra, dog, elephant, giraffe, cat, horse, sheep, & Animals \\
bird, cow, bear, starfish & Animals \\
pizza, broccoli, donut, orange, apple, tomato, potato & Food \\
cake, egg, wine glass, bowl, bottle, onion  & Food \\
bottle, cup, wine glass, bowl, medicine\_bottle, box & Containers \\
chair, sofa, table, lamp, bench, sofa-chair, soft-sofa & Furniture \\
laptop, tv, display, camera, laptop-camera & Electronics \\
clock, vase, toilet, umbrella, teddy bear, rifle & Miscellaneous \\
stop sign, toy\_train, toy\_bus, house, ball & Miscellaneous \\
scene, pattern & Miscellaneous \\
\bottomrule
\end{tabular}
\end{table}

\label{section:broad_classes}
To enable category-level analysis, we group the fine-grained object labels in DORI into a smaller set of semantically meaningful broad classes. This abstraction facilitates more interpretable evaluation across heterogeneous datasets and reduces sparsity in underrepresented categories. The mapping spans common categories such as \texttt{Vehicle}, \texttt{Person}, \texttt{Animals}, \texttt{Food}, \texttt{Containers}, \texttt{Furniture}, \texttt{Electronics}, and \texttt{Miscellaneous}, and is illustrated in \cref{tab:category_mapping} and \cref{fig:obj_class_distribution}. 

\subsection{VQA Examples in DORI}
\label{app:vqa_examples}
To further contextualize model behavior, we present a curated selection of Visual Question Answering (VQA) examples drawn from the DORI benchmark, covering each of the seven distinct orientation reasoning tasks. These qualitative illustrations shed light on the nuanced challenges faced by state-of-the-art Multimodal Large Language Models (MLLMs), beyond aggregate metrics.

Each example is carefully chosen to represent either a canonical failure or, in rarer cases, a surprising success. The samples span both coarse-level and granular-level questions, reflecting the dual axes of abstraction and visual complexity within DORI. Despite confident language in many model predictions, we observe frequent dissonance between answer correctness and the underlying rationale, especially for tasks requiring precise spatial alignment or inter-object reasoning.
For instance, in the \textbf{View Parallelism} and \textbf{Directional Facing} tasks, models often misjudge subtle orientation cues, such as limb articulation or gaze direction as we show in \cref{fig:example_question_1_coarse_fail},\cref{fig:example_question_1_granular_fail}, \cref{fig:example_question_2_coarse_fail}, and \cref{fig:example_question_2_coarse_fail2}, leading to confidently incorrect predictions. Likewise, for \textbf{Single-Axis} and \textbf{Compound Rotation} scenarios, we show in \cref{fig:example_question_3_granular_fail},\cref{fig:example_question_3_granular_fail2}, \cref{fig:example_question_4_coarse}, and \cref{fig:example_question_4_granular_fail}, that even top-performing models struggle with mentally simulating object motion, resulting in angular miscalculations or overly generic justifications.

Notably, \textbf{Inter-object Direction} and \textbf{Viewer-Scene Direction} tasks as we show in \cref{fig:example_question_5_coarse_fail},\cref{fig:example_question_5_granular_fail}, \cref{fig:example_question_6_granular}, and \cref{fig:example_question_6_granular_fail}, expose limitations in relational orientation reasoning, with models frequently underestimating angular disparities or misrepresenting the directional frame of reference. The final task, \textbf{Canonical Orientation} as seen in \cref{fig:example_question_7_coarse_fail}, underscores a broader epistemic gap: models often assert certainty in inherently ambiguous scenarios, revealing an overconfidence not grounded in the visual evidence.

Together, these examples highlight persistent limitations in visual-spatial grounding, even among the most capable contemporary MLLMs. They underscore the need for further architectural innovations and training strategies to imbue models with a deeper, more structured understanding of object orientation dynamics.
\subsection{Enumerated List of Questions}
\label{app: enumerated_qs}
\label{qlists}
\noindent \paragraph{Coarse Questions}
\begin{itemize}
    \item \textbf{Q1 - View Parallelism:} Determine which way Object A's front is facing relative to the camera.
    \item \textbf{Q2 - Directional Facing:} Determine which direction Object A's front-facing surface is oriented from the camera's viewpoint.
    \item \textbf{Q3 - Single-axis Rotation:} Determine the shortest direction of rotation for Object A to face the camera.
    \item \textbf{Q4 - Compound Rotation:} Determine what type of rotation the object has undergone between the two images.
    \item \textbf{Q5 - Inter-object Direction:} Determine if objects A and B are facing each other from their own perspectives.
    \item \textbf{Q6 - Viewer-Scene Direction:} Determine if the object has rotated between the two images.
    \item \textbf{Q7 - Canonical Orientation:} Determine if the image is in its canonical orientation.
\end{itemize}
\noindent \paragraph{Fine-grained/Granular Questions}
\begin{itemize}
    \item \textbf{Q1 - View Parallelism:} Determine how much Object A's front surface deviates from being parallel to the camera plane.
    \item \textbf{Q2 - Directional Facing:} Identify the precise orientation of Object A's front-facing surface from the camera's viewpoint
    \item \textbf{Q3 - Single-axis Rotation:} Determine the closest clockwise rotation needed for Object A to face the camera.
    \item \textbf{Q4 - Compound Rotation:} Determine the exact rotation angles the object has undergone between the two images.
    \item \textbf{Q5 - Inter-object Direction:} Determine how much Object B would need to rotate to face Object A.
    \item \textbf{Q6 - Viewer-Scene Direction:} Determine how many degrees clockwise the object has rotated between the two images.
    \item \textbf{Q7 - Canonical Orientation:} Determine how the image can be restored to its canonical orientation.
\end{itemize}
\subsection{Confounding Factors}
\label{section:confounding_facts}
In our dataset we have carefully designed our questions for both the prompt and image to reduce the amount of different confounding factors that are present in past datasets. We include aspects like: defining axes, defining what the canonical view is, and we direct the readers to section \cref{app:vqa_examples} \& \cref{ref: glossary} which showcases these elements present in our questions.

We can see in examples like \cref{fig:example_question_1_granular_fail} we utilize a bounding box which helps the model/rater determine which object to focus on; this helps reduce object recognition difficulty. For scene clutter we can look at \cref{fig:example_question_6_granular_fail} which focuses on particular objects without any objects being present in the image. \cref{fig:example_question_2_coarse_fail2} showcases an example of reducing linguistic ambiguity, by describing what the MLLM should look at and identify its front facing surface, additionally objects that have ambiguous front facing surface like a table we have labelled those as “Cannot be determined”. For contextual distractions we can also look to \cref{fig:example_question_4_coarse} which contains backgrounds that do not influence the exact orientation of the object.

\begin{figure}[t]
   \begin{flushleft}
    \includegraphics[page=1, width=\linewidth]{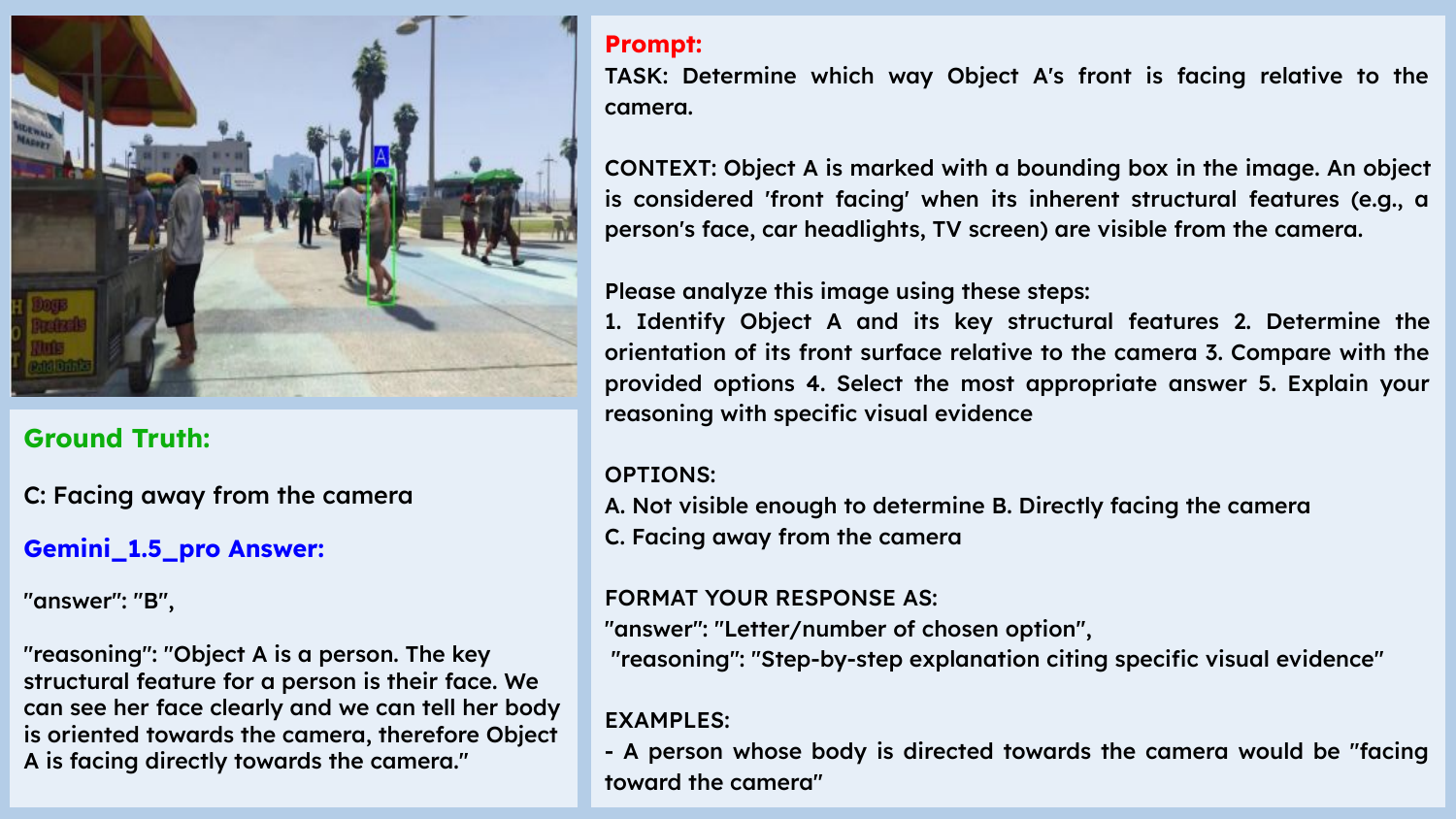}
  \end{flushleft}
  \caption{An example VQA from the \textbf{View Parallelism orientation task} illustrating a failure case on a coarse-level question in DORI, with a prediction from Gemini 1.5 Pro. Many models failed this question, highlighting a common challenge among MLLMs in understanding fundamental scene geometry. Specifically, in reasoning whether an object's front-facing surface is oriented toward, away from, or perpendicular to the camera plane.}
  \label{fig:example_question_1_coarse_fail}
\end{figure}

\begin{figure}[t]
   \begin{flushleft}
    \includegraphics[page=2, width=\linewidth]{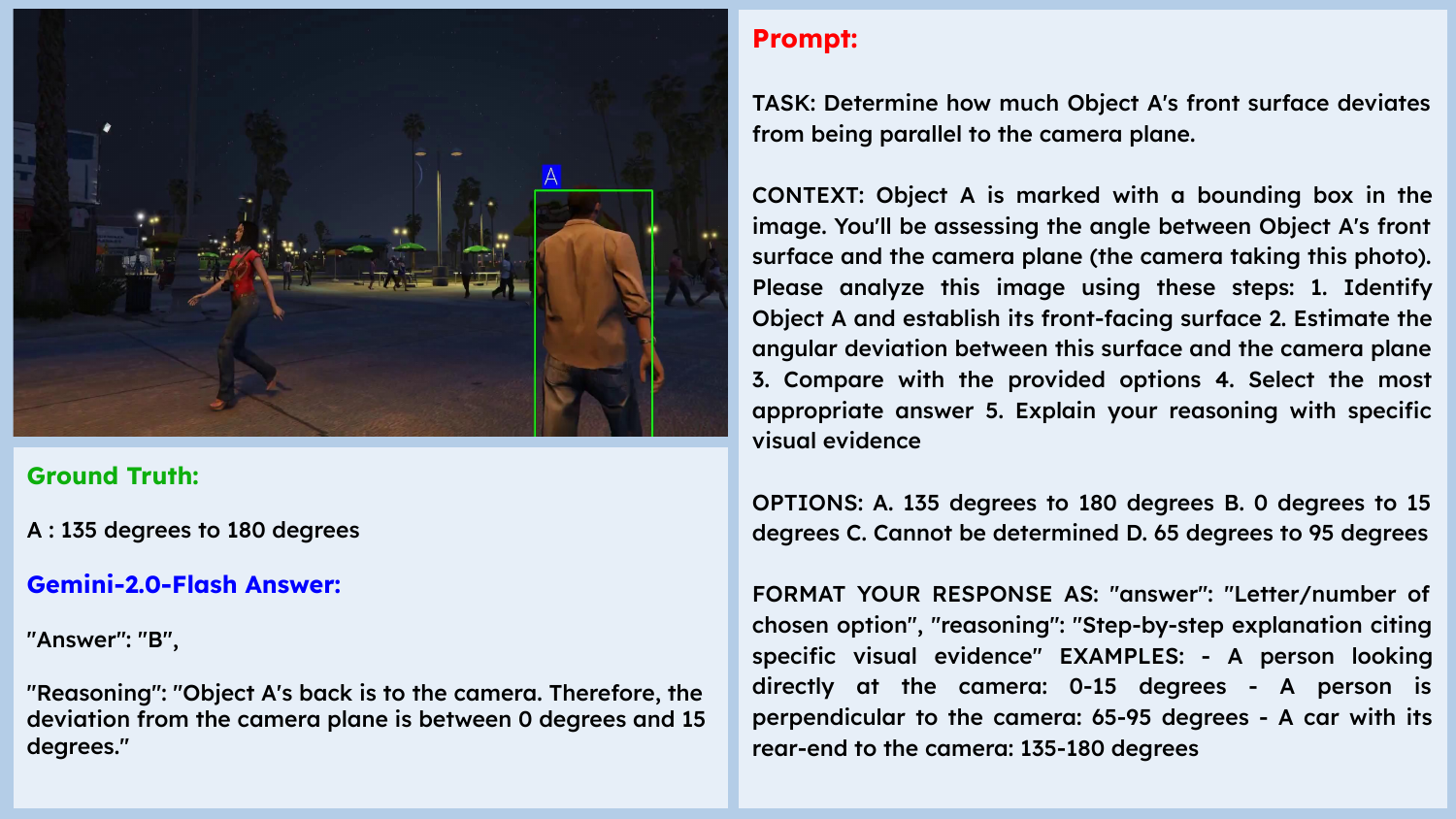}
  \end{flushleft}
  \caption{An example VQA from the \textbf{View Parallelism orientation task} highlighting a failure case on a granular-level question in DORI, with a response from Gemini 2.0 Flash. Although the model confidently selects an answer, its reasoning reflects a fundamental misunderstanding of the object's orientation relative to the camera. While the ground truth indicates the object is turned between 135° and 180° away, the model incorrectly infers a near-parallel alignment, underscoring limitations in fine-grained spatial perception among current MLLMs.}
  \label{fig:example_question_1_granular_fail}
\end{figure}
\begin{figure}[t]
   \begin{flushleft}
        \includegraphics[page=3, width=\linewidth]{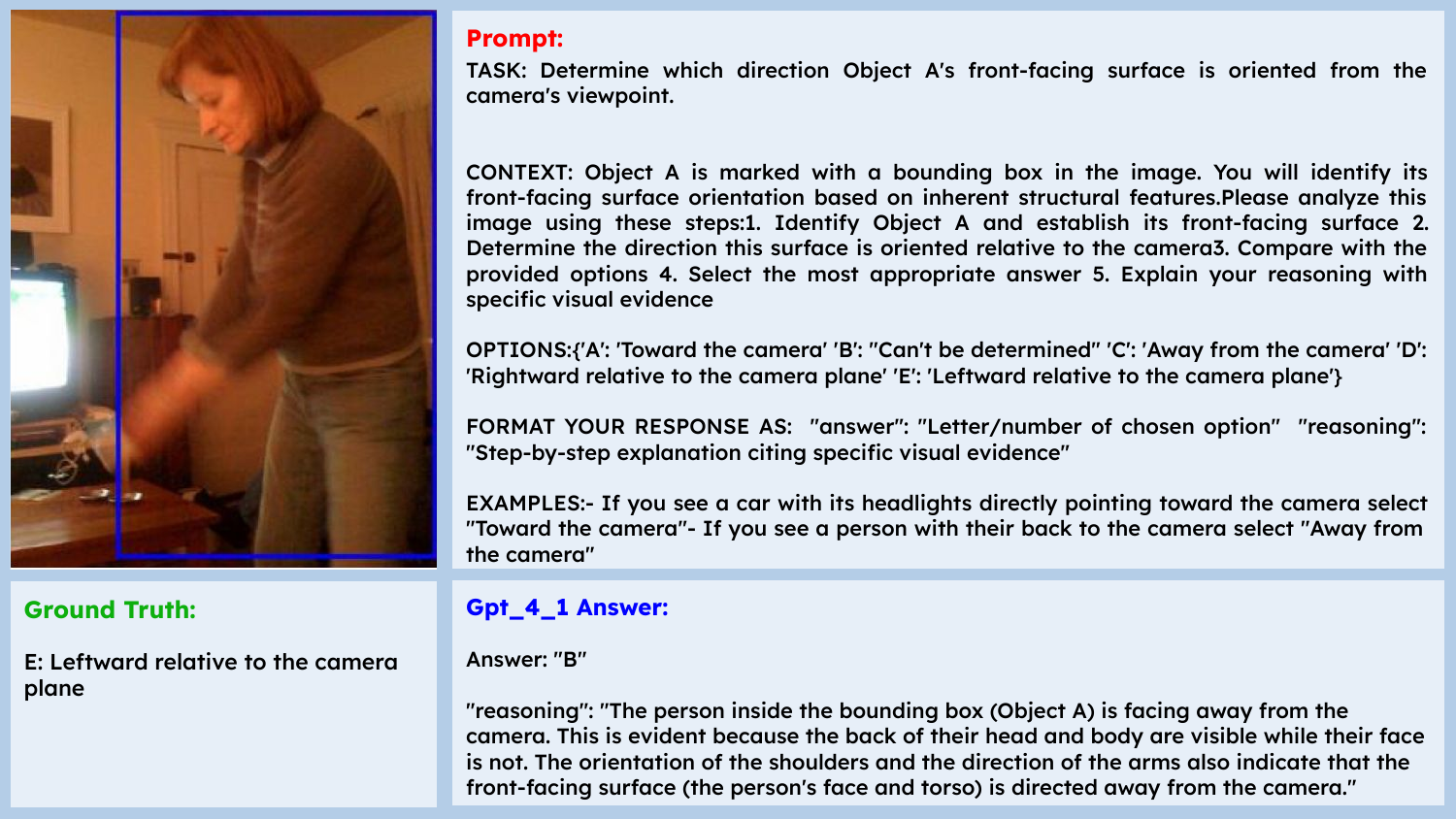}
  \end{flushleft}
  \caption{A coarse-level VQA from the \textbf{Directional Facing task} in DORI, illustrating a failure case by GPT-4-1. While the ground truth indicates a leftward orientation, the model predicts ‘away,’ misreading body posture and head direction. Notably, 13 out of 15 models failed this question, underscoring a widespread struggle with coarse directional inference.}
  \label{fig:example_question_2_coarse_fail}
\end{figure}

\begin{figure}[ht]
   \begin{flushleft}
        \includegraphics[page=4, width=\linewidth]{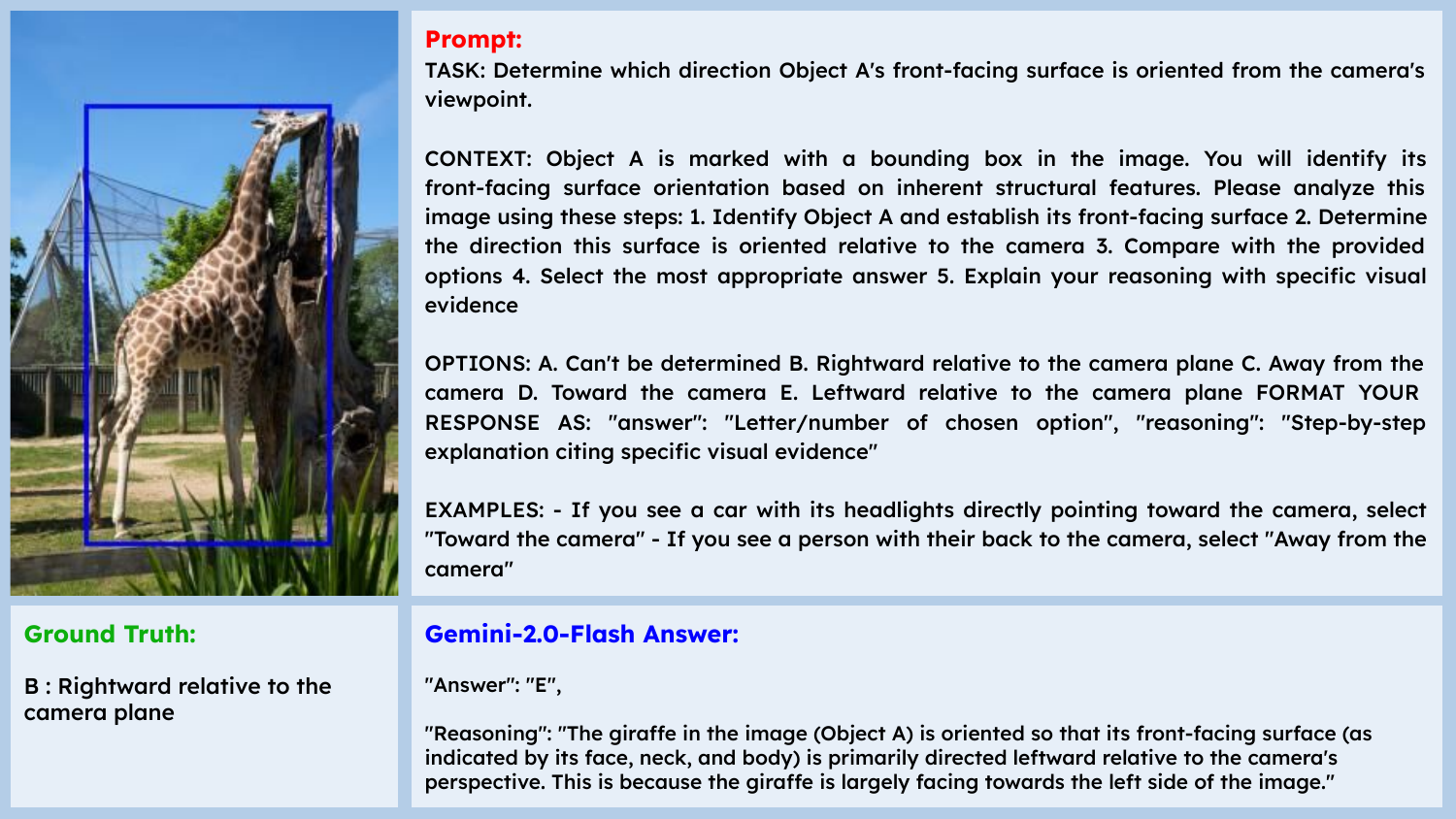}
  \end{flushleft}
  \caption{A coarse-level VQA from the \textbf{Directional Facing task} in DORI, showing a failure case by Gemini-2.0-Flash. While the ground truth identifies the giraffe’s orientation as rightward, the model incorrectly infers a leftward direction. This highlights the difficulty in interpreting animal pose and orientation cues, a challenge shared by the majority of models in this example.}
  \label{fig:example_question_2_coarse_fail2}
\end{figure}

\begin{figure}[t]
   \begin{flushleft}
        \includegraphics[page=5, width=\linewidth]{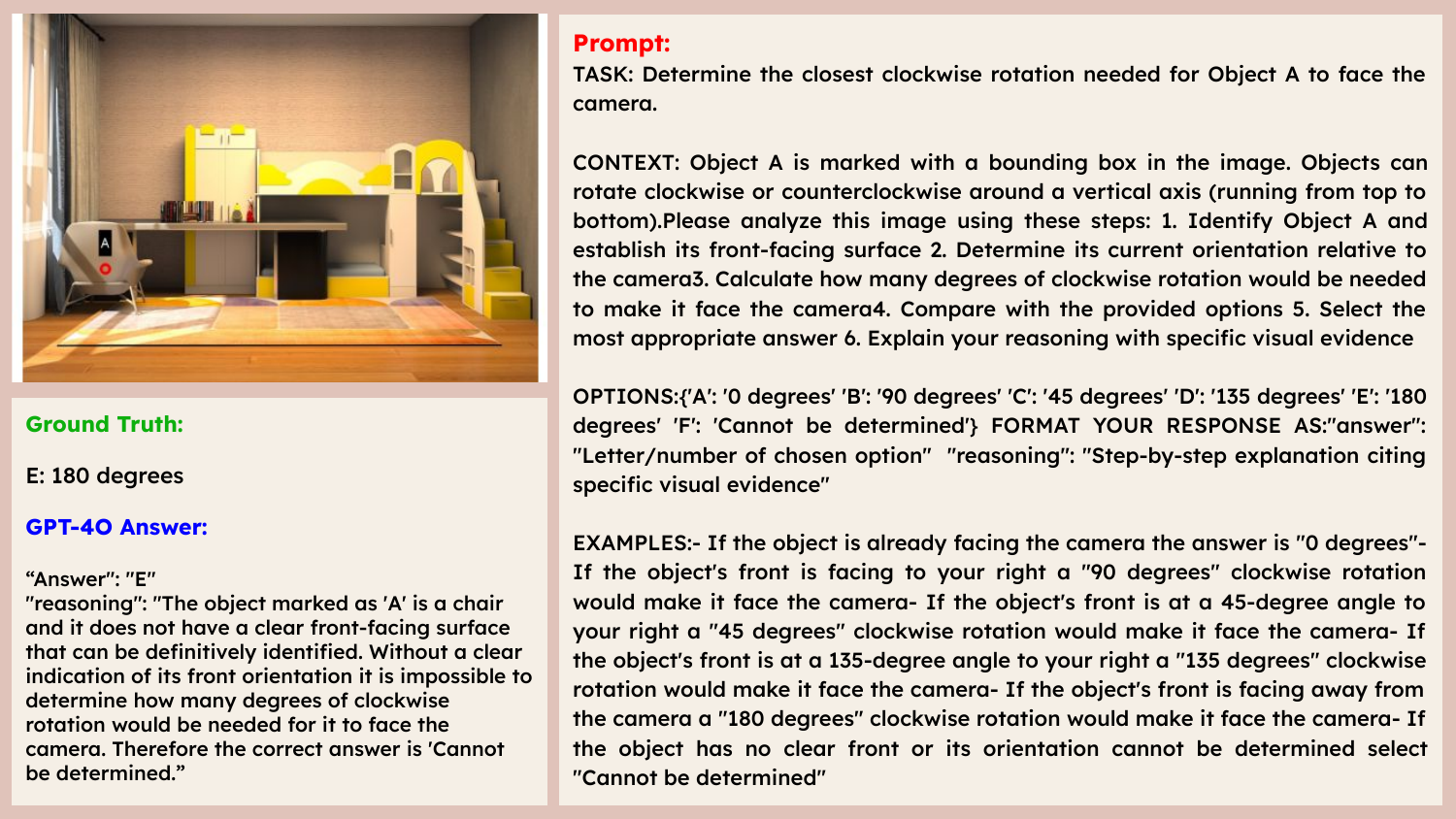}
  \end{flushleft}
  \caption{A granular-level VQA from the \textbf{Single-Axis Rotation} task in DORI, featuring a case where GPT-4o selects the correct answer (180° rotation) but provides flawed reasoning. While the model concludes with the correct choice, its explanation suggests uncertainty due to perceived ambiguity in object orientation. This mismatch between prediction and rationale underscores gaps in visual reasoning, with only 2 out of 15 models answering correctly.}
  \label{fig:example_question_3_granular_fail}
\end{figure}

\begin{figure}[t]
   \begin{flushleft}
        \includegraphics[page=6, width=\linewidth]{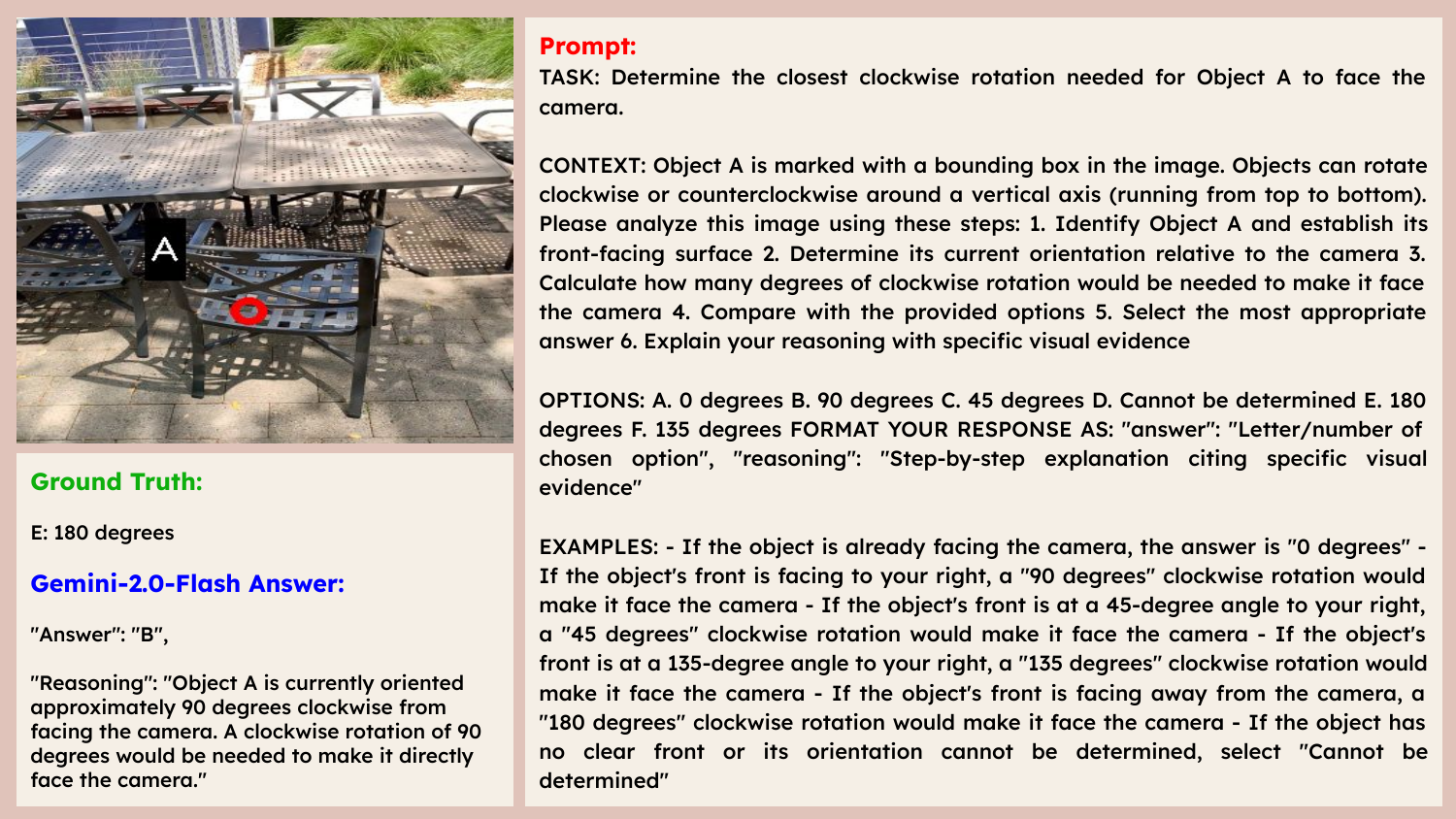}
  \end{flushleft}
  \caption{A granular-level VQA from the \textbf{Single-Axis Rotation} in DORI, showing a failure case with a prediction from Gemini 2.0 Flash. While the correct answer is a 180° rotation, the model incorrectly selects 90° (option B), suggesting a misjudgment of the object’s current orientation relative to the camera. This highlights challenges models face in reasoning about precise rotational alignment.}
  \label{fig:example_question_3_granular_fail2}
\end{figure}

\begin{figure}[t]
   \begin{flushleft}
       \includegraphics[page=7, width=\linewidth]{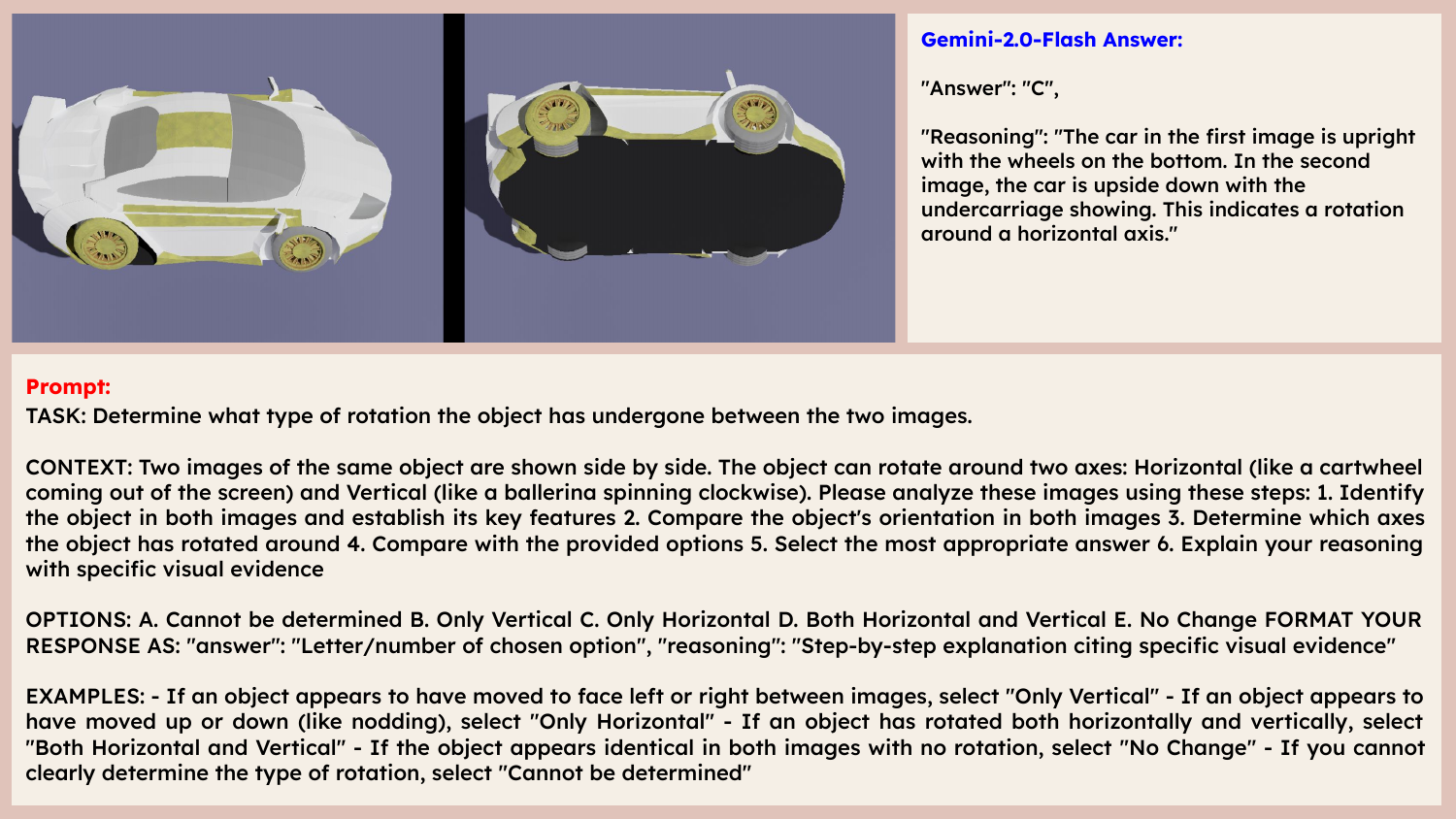}
  \end{flushleft}
  \caption{A coarse-level VQA from the \textbf{Compound Rotation task} in DORI, with a correct prediction from Gemini 2.0 Flash. The model identifies a horizontal axis rotation, but its reasoning overstates the change, describing the car as ‘upside down’ when it is only partially inverted.}
  \label{fig:example_question_4_coarse}
\end{figure}

\begin{figure}[t]
   \begin{flushleft}
        \includegraphics[page=8, width=\linewidth]{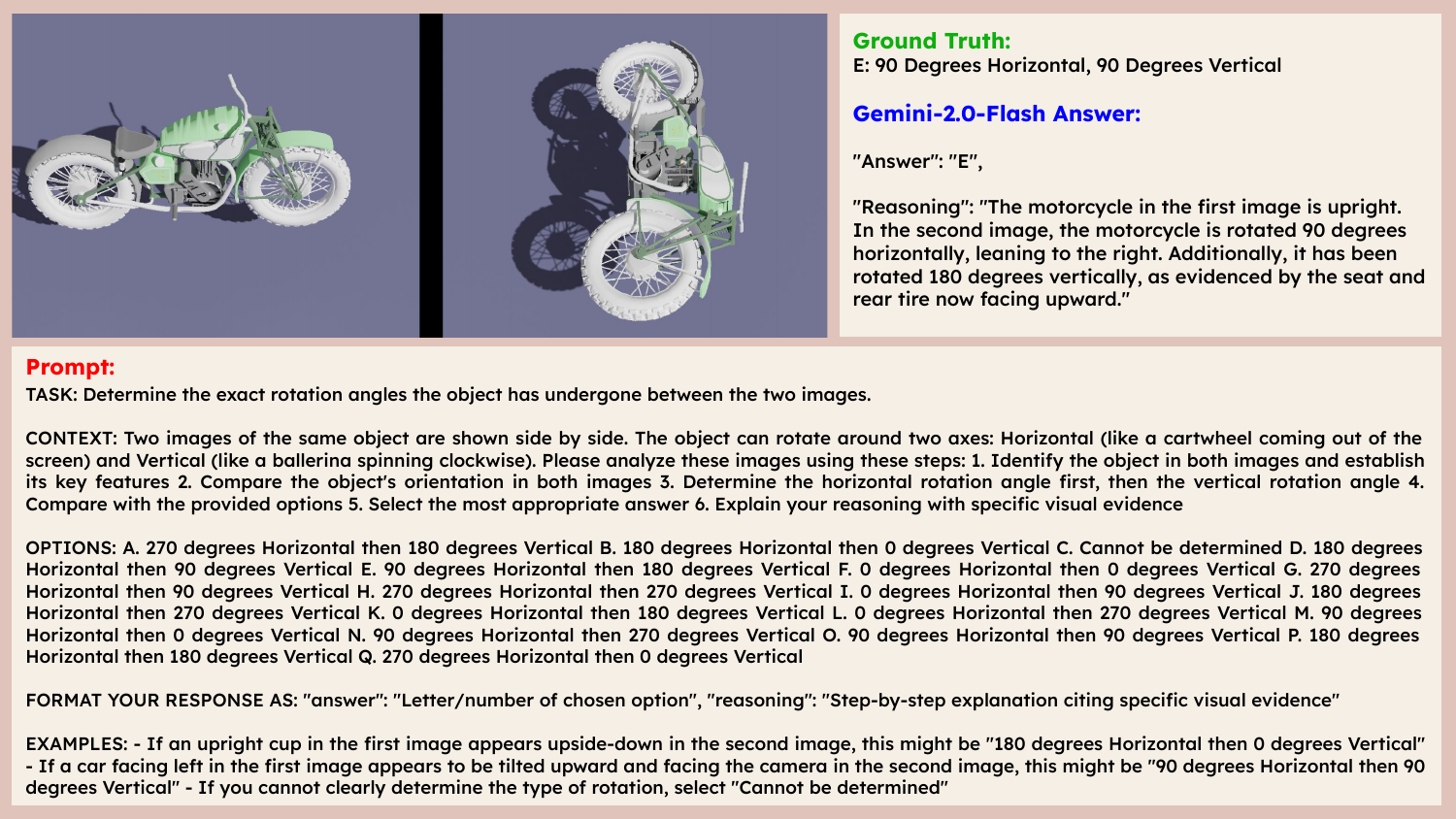}
  \end{flushleft}
  \caption{A granular-level VQA from the \textbf{Compound Rotation task} in DORI showing a failure case from Gemini 2.0 Flash. While the model selects the correct option, its reasoning misidentifies the vertical rotation, describing a 180° flip when the ground truth indicates a 90° transformation.}
  \label{fig:example_question_4_granular_fail}
\end{figure}

\begin{figure}[t]
   \begin{flushleft}
        \includegraphics[page=10, width=\linewidth]{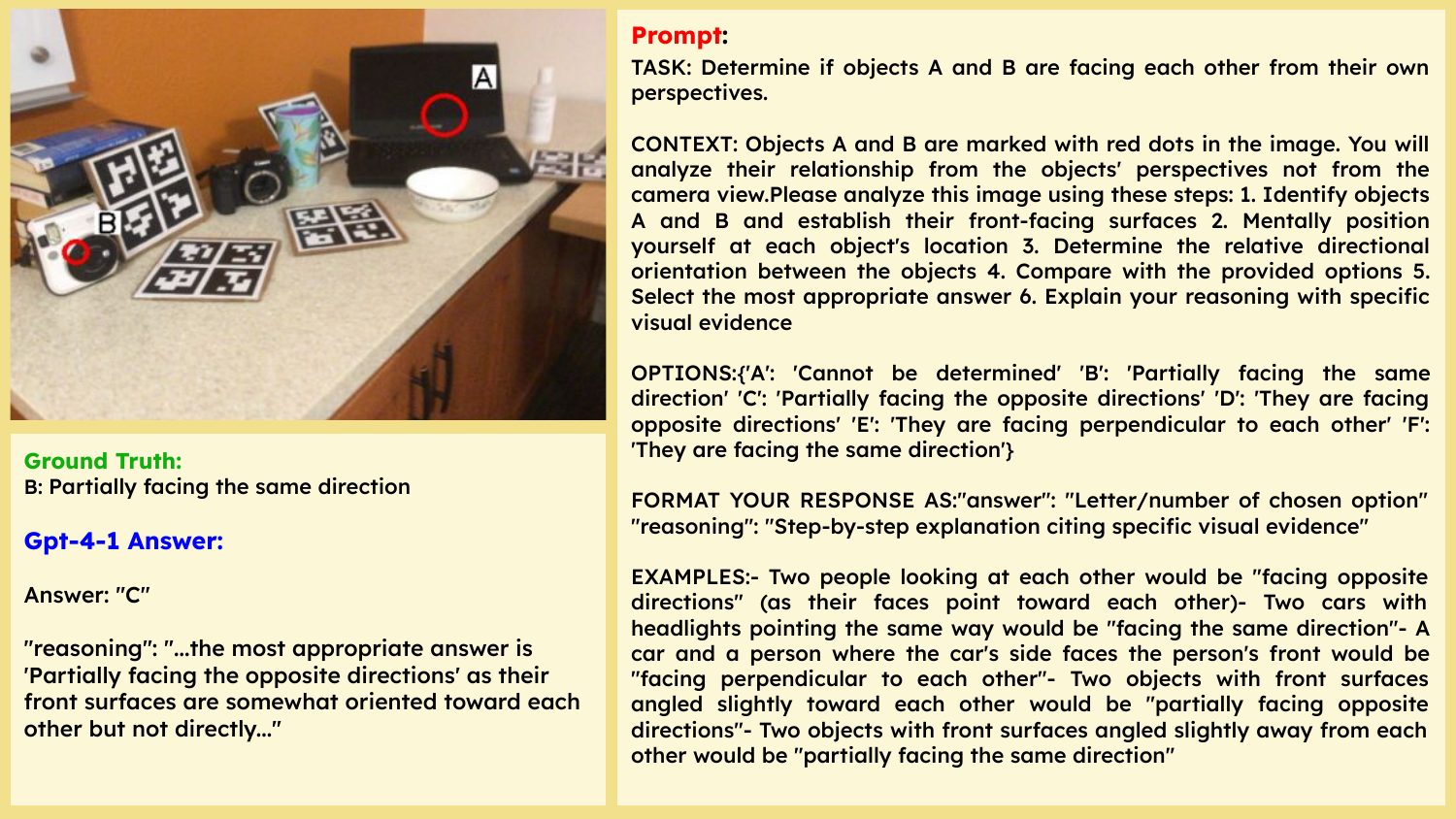}
  \end{flushleft}
  \caption{A coarse-level VQA from the \textbf{Inter-object Direction task} in DORI showing a failure case with GPT-4-1. Although the correct answer is 'Partially facing the same direction,' the model incorrectly selects 'Partially facing opposite directions,' misjudging the relative orientations. Notably, 13 out of 15 models failed this example, highlighting a shared difficulty in reasoning about partially aligned object directions.}
  \label{fig:example_question_5_coarse_fail}
\end{figure}

\begin{figure}[t]
   \begin{flushleft}
        \includegraphics[page=9, width=\linewidth]{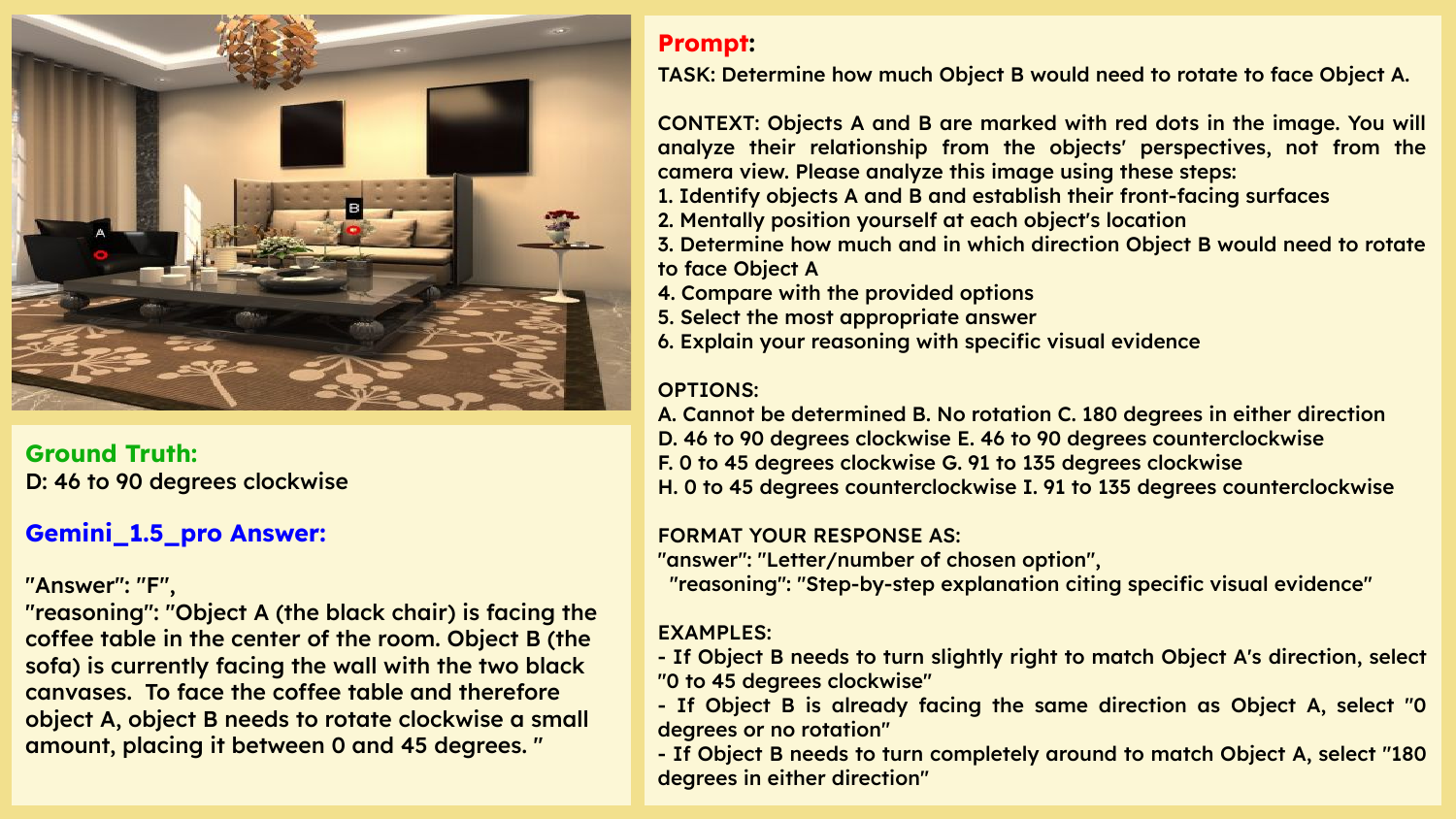}
  \end{flushleft}
  \caption{A granular-level VQA from the \textbf{Inter-object Direction task} in DORI illustrating a failure case with Gemini 1.5 Pro. While the ground truth indicates a required clockwise rotation between 46° and 90° for alignment, the model underestimates this angle. Its reasoning misinterprets the spatial alignment between the chair and sofa. All 15 models failed this example, underscoring the challenge of inter-object directional understanding.}
  \label{fig:example_question_5_granular_fail}
\end{figure}


\begin{figure}[t]
   \begin{flushleft}
        \includegraphics[page=11, width=\linewidth]{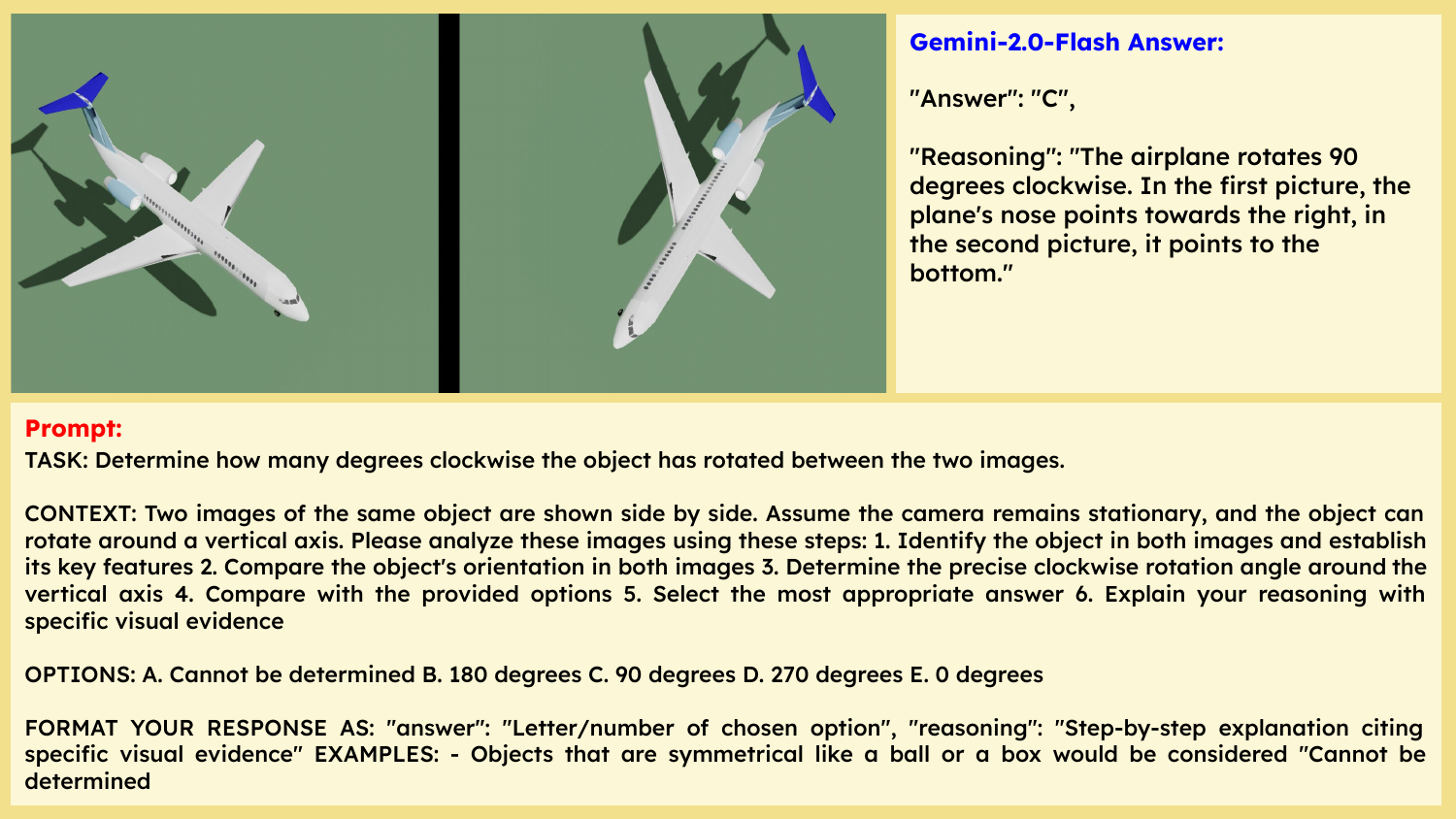}
  \end{flushleft}
  \caption{\textbf{Viewer-Scene Direction.} Example success case on a granular-level DORI question with Gemini-2.0-Flash. While the model correctly selects the 90-degree rotation, it inaccurately describes the directional shift as 'right' to 'bottom' instead of the more precise 'bottom-right' to 'bottom-left,' reflecting a partial misunderstanding of fine-grained orientation.}
  \label{fig:example_question_6_granular}
\end{figure}

\begin{figure}[t]
   \begin{flushleft}
       \includegraphics[page=12, width=\linewidth]{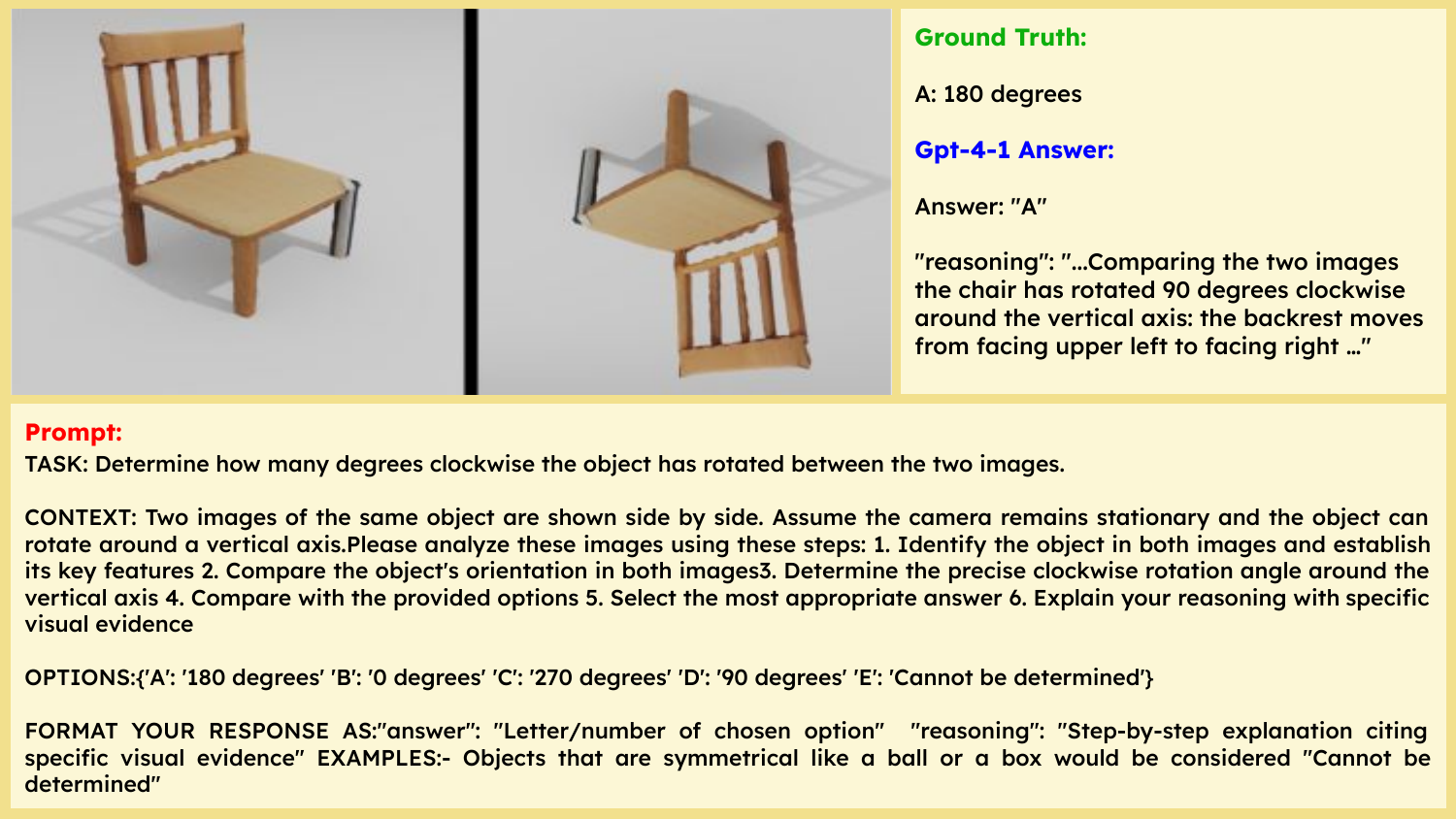}
  \end{flushleft}
  \caption{\textbf{Viewer-Scene Direction.} Example failure case on a granular-level DORI question with GPT-4-1. Although the model selects the correct answer of 180 degrees, its reasoning mistakenly describes a 90-degree rotation, highlighting a disconnect between answer selection and spatial understanding. Notably, 13 out of 15 models failed this case.}
  \label{fig:example_question_6_granular_fail}
\end{figure}

\begin{figure}[t]
   \begin{flushleft}
        \includegraphics[page=14, width=\linewidth]{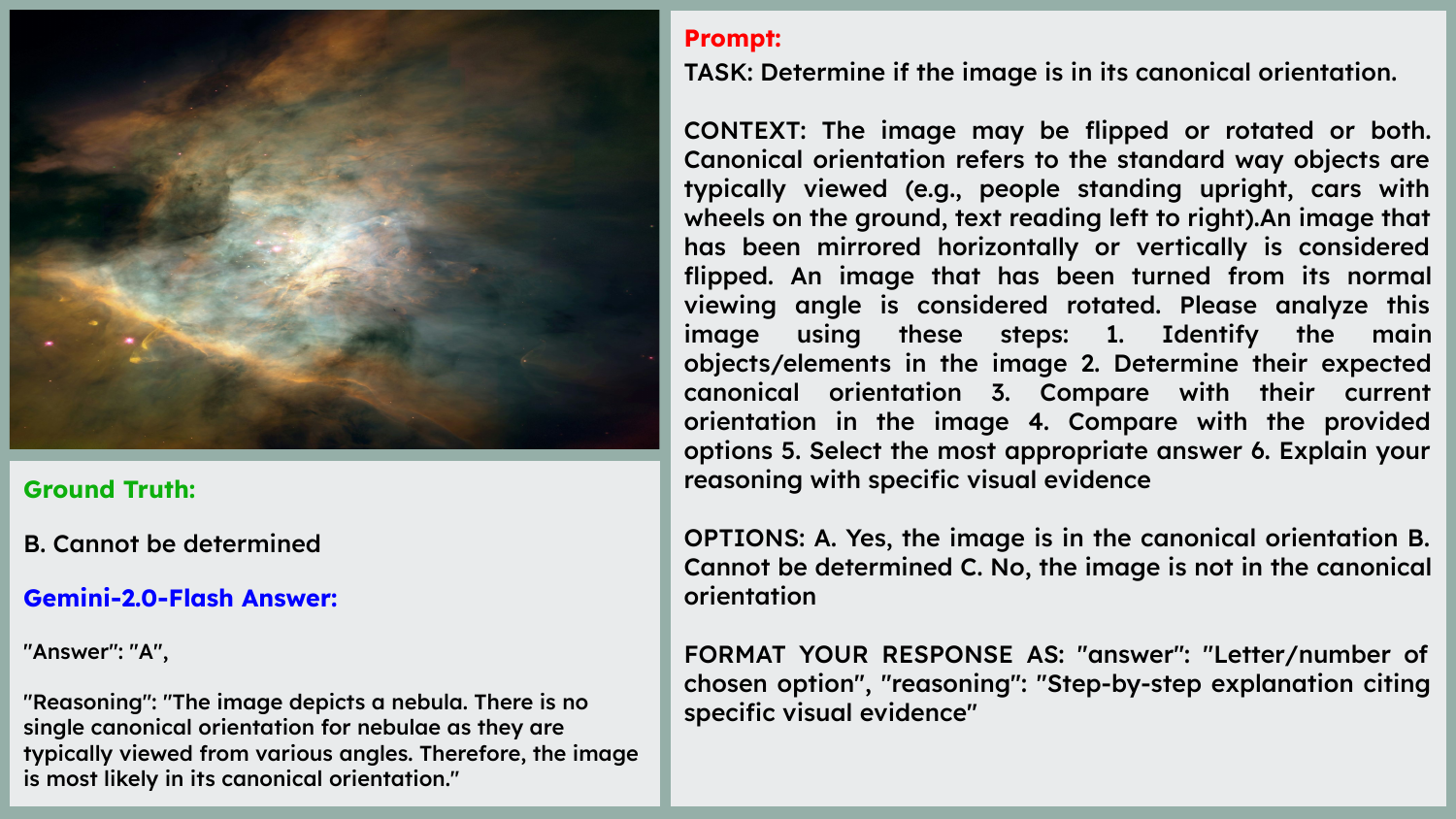}
  \end{flushleft}
  \caption{\textbf{Canonical Orientation.} Example failure case on a coarse-level DORI question with Gemini-2.0-Flash. While the ground truth is 'Cannot be determined,' the model incorrectly selects a definitive orientation. Its reasoning contradicts the inherent ambiguity it acknowledges, exposing uncertainty in handling objects with no fixed canonical pose.}
  \label{fig:example_question_7_coarse_fail}
\end{figure}
\label{ref: clockwise_counter_clockwise}
\begin{figure}[t]
    \centering
    \includegraphics[width=\linewidth]{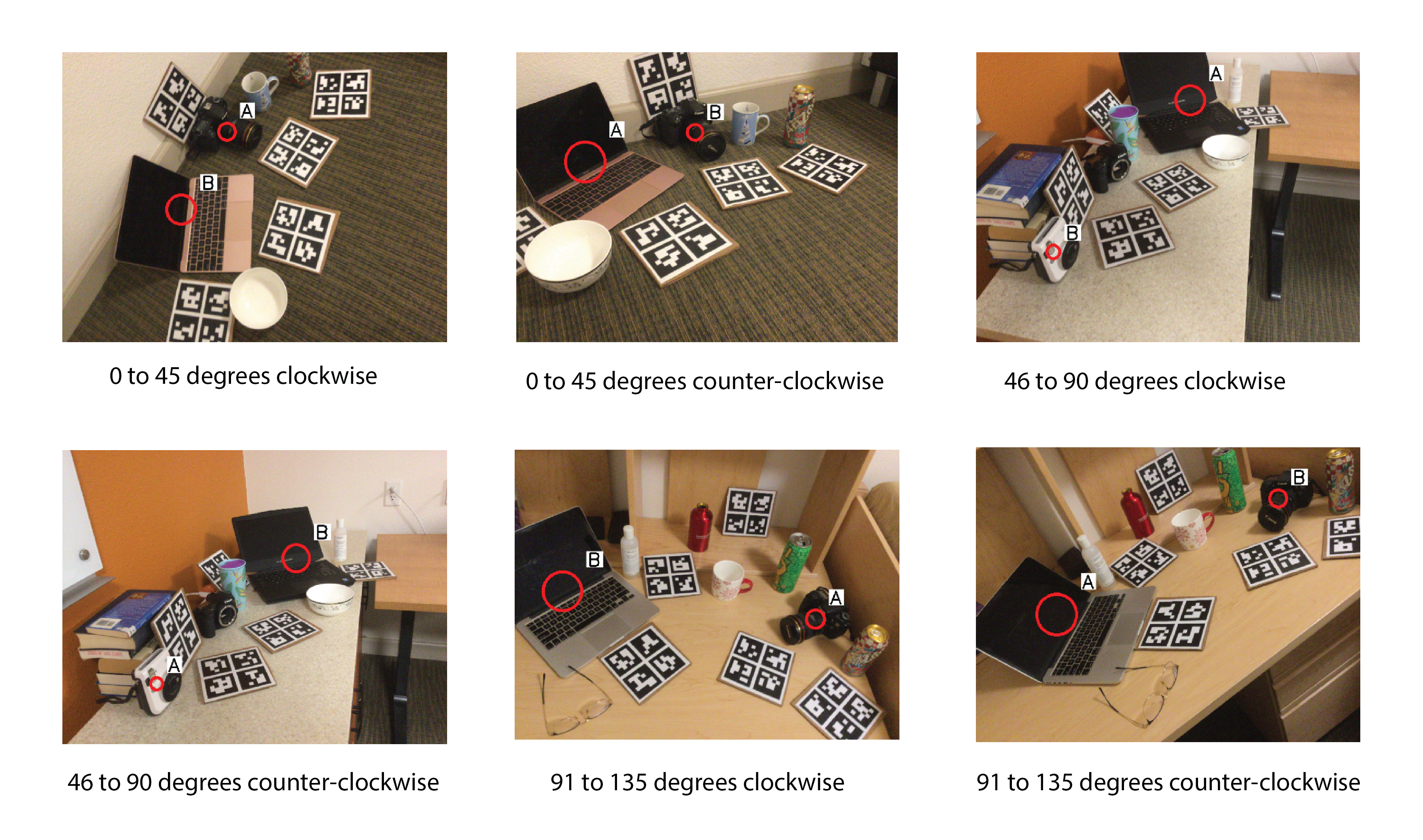}
    \caption{{An example of clockwise and counter-clockwise rotations for the Inter-object direction perception task, we utilized samples from the NOCS REAL dataset}}
    \label{fig:clockwise_and_counter_clockwise}
\end{figure}
\clearpage
\section{Additional Analyses of Model Performance}
\label{app:more_model_analysis}
To deepen our understanding of current Multimodal Large Language Models (MLLMs) on DORI, we present a series of additional analyses that dissect performance across key axes of task structure and model behavior.

\cref{fig:perf_by_num_options} examines the relationship between answer set size and model accuracy. We observe a peak in performance at 3-option questions (42.5\%), with accuracy declining markedly as the number of candidate answers increases. At 5 and 6 options, performance drops to 26\% and 19\%, respectively, with a further decline to 13\% at 9 options, and a minimum of 6\% at 16 options. This degradation illustrates the difficulty MLLMs face when navigating more complex decision spaces, suggesting that increasing output space size strains their orientation reasoning capabilities.

\begin{figure}[ht]
    \centering
    
    \includegraphics[width=0.8\linewidth]{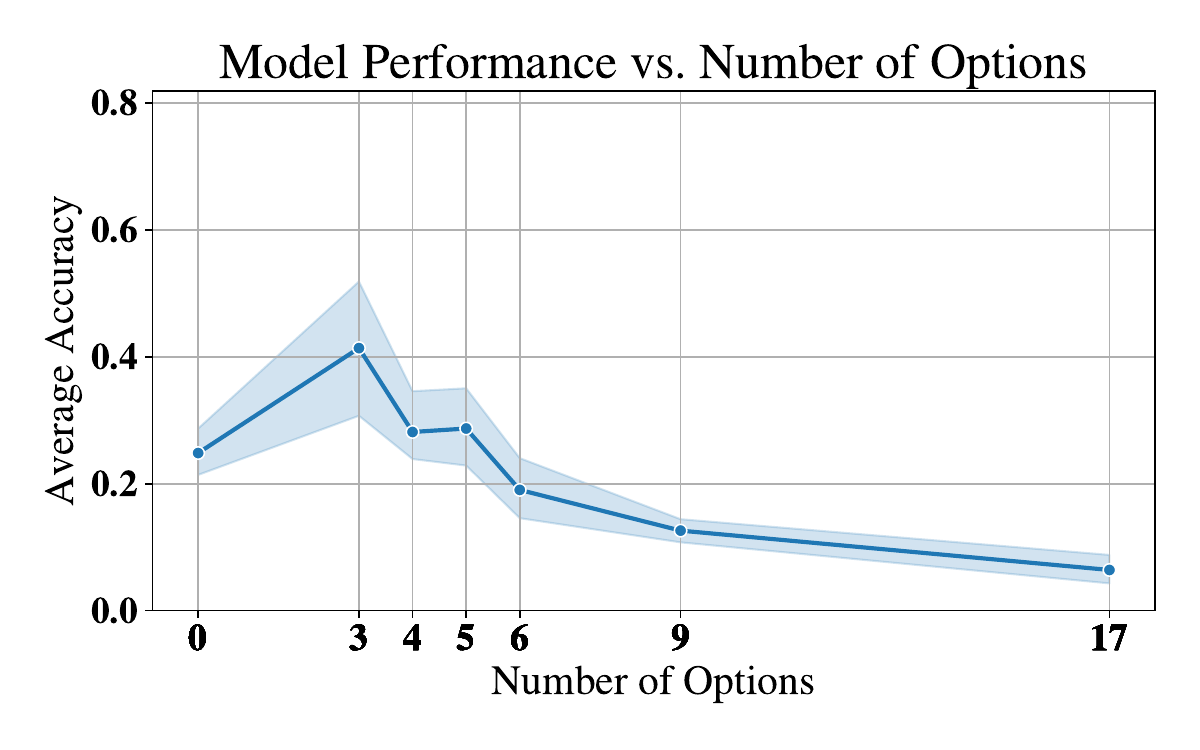}
    \caption{Model accuracy as a function of answer set size. Accuracy peaks at 42.5\% for 3-option questions, then declines steadily, dropping to about 25\% at 5 options, 19\% at 6, about 13\% at 9 options, and reaching a low of about 6\% at 16 options, indicating increasing difficulty with larger candidate sets.}
    \label{fig:perf_by_num_options}

  \end{figure}

In \cref{fig:model_performance_overview}, we break down model performance across simulated vs. natural imagery and annotation granularity levels. Gemini and GPT variants dominate across the board. Yet, all models exhibit significantly higher accuracy on simulated datasets and coarse-type questions, revealing a persistent challenge in transferring orientation understanding to more fine-grained and realistic visual contexts.

\begin{figure}[ht]
  \centering

  \begin{subfigure}[b]{0.48\linewidth}
    \centering
    \includegraphics[width=\linewidth]{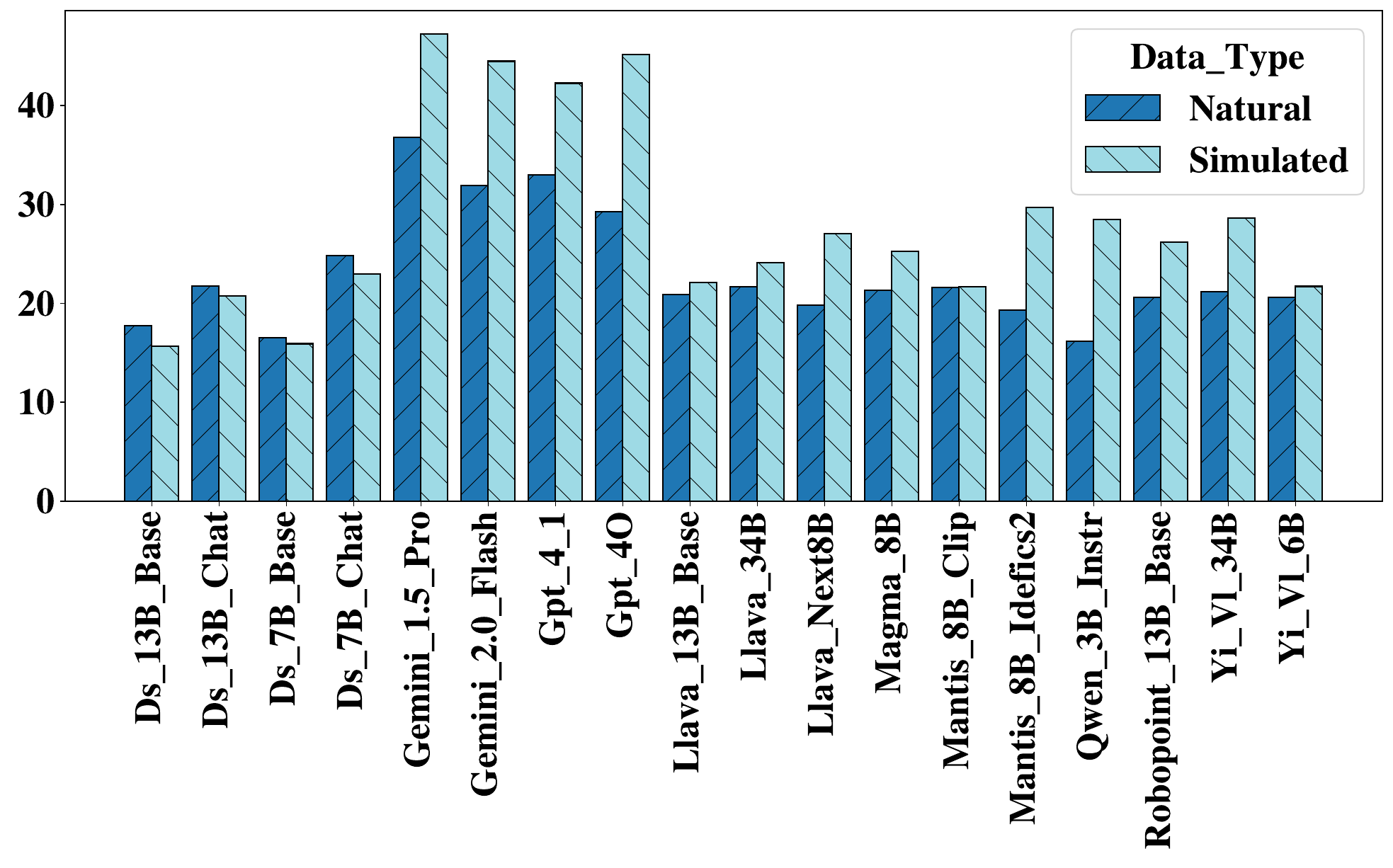}
    \caption{Performance grouped by data type.}
    \label{fig:by_data_type}
  \end{subfigure}
  \hfill
  \begin{subfigure}[b]{0.48\linewidth}
    \centering
    \includegraphics[width=\linewidth]{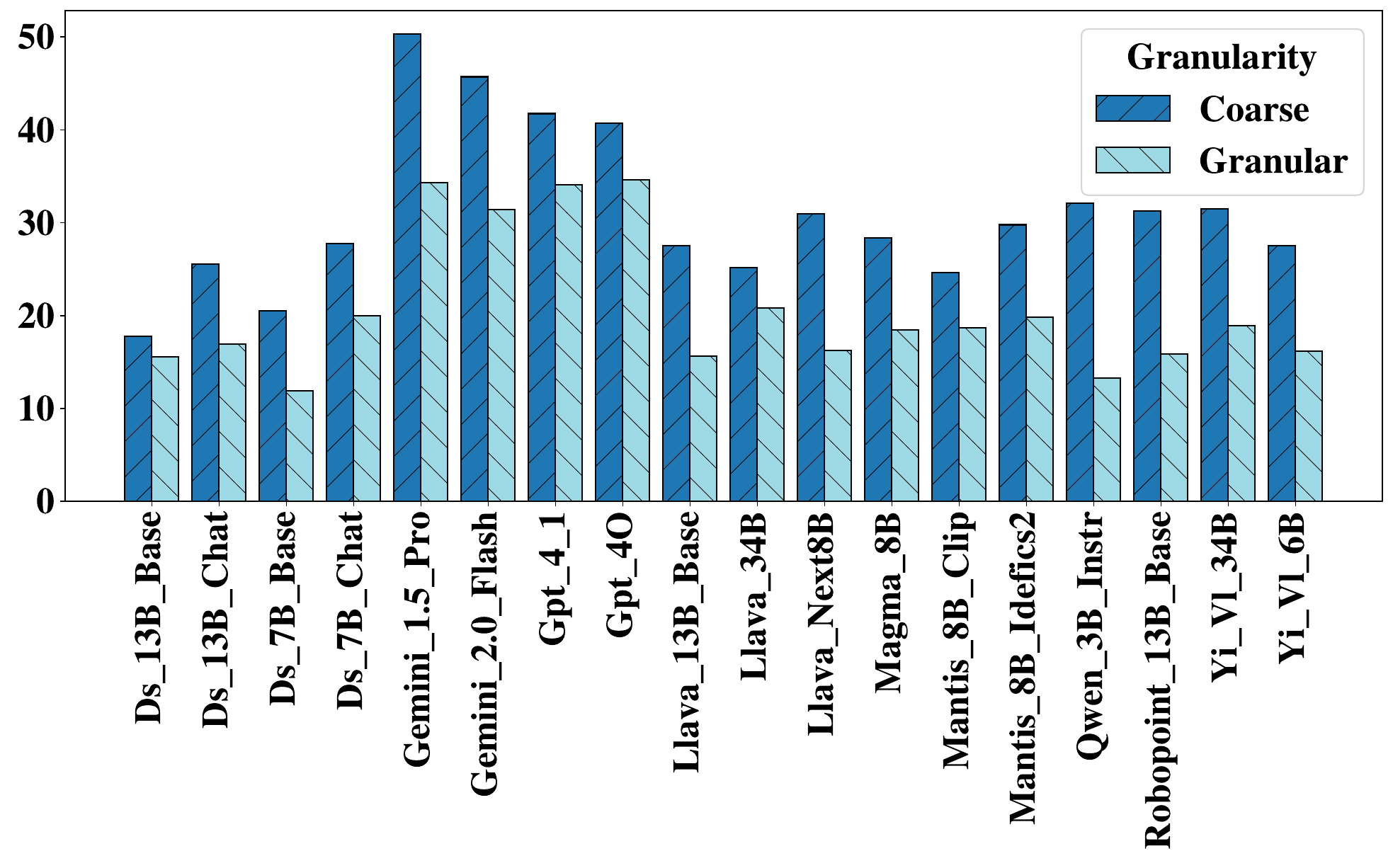}
    \caption{Performance grouped by question granularity.}
    \label{fig:by_granularity}
  \end{subfigure}

  \caption{Performance of 18 leading Multimodal Large Language Models (MLLMs) across data types \textbf{(a)} and annotation granularity levels \textbf{(b)}. \textbf{Gemini} and \textbf{GPT} models lead overall. Models perform noticeably better on simulated datasets and coarse-type questions, revealing a gap in generalizing to natural images and more fine-grained questions.}
  \label{fig:model_performance_overview}
\end{figure}

\subsection{Performance by Common Categories}
\label{section:common_categories}
We can see in \cref{fig:model_perf_by_category} we look at the performance for each of the models based on common categories. We see a similar trend for each model across each of the different categories with Food being the one that appears to standout. In this instance it seems that only the GPT variants perform better with GPT-4o performing the best. In terms of datasets that contain food in them, this would be attributed to the COCO images for the Directional Facing question and OmniObject-3D images for Compound Rotation and View-scene direction questions. Where some of these questions tend to have "Cannot be determined" since it involves items like Pizza or Cake, GPT-4o appears to be able to determine these sort of questions. This could be attributed to the training data that was used for the GPT family of models since GPT-4-1 also is shown to do well for the Food category. For the person category, however, most models tend to perform best as is expected as most models would have been exposed to persons in their training data, for instance if we look at the most popular category in COCO~\cite{mscoco} it is persons.

\begin{figure}[ht]
    \centering
    
    \includegraphics[width=01.0\linewidth]{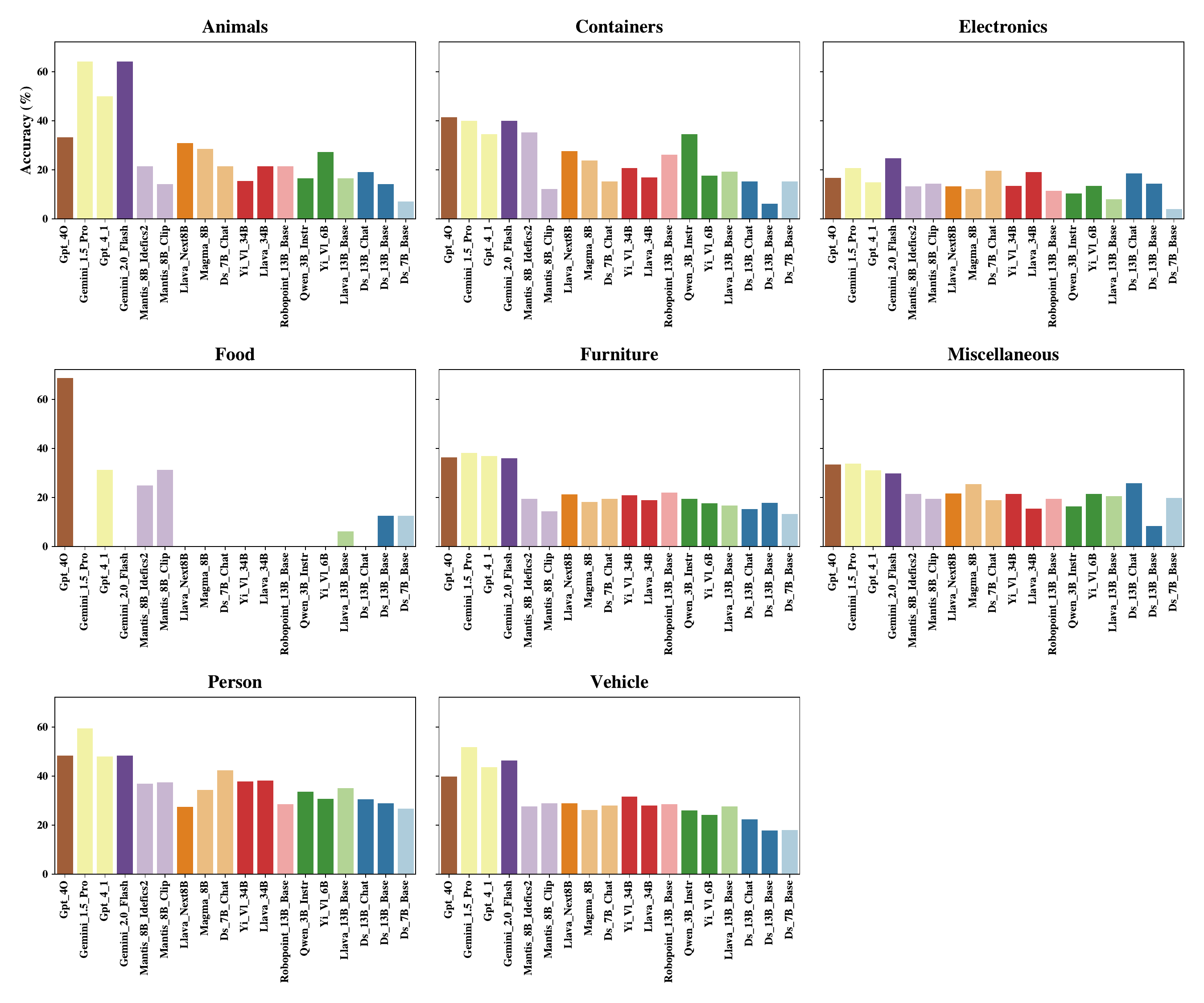}
    \caption{Model accuracy for the 18 models across the broad categories described in Section\cref{section:broad_classes} with most categories tending to performing similarly except for Food which tends to have lower performances except for GPT-4o. Which highlights a gap for orientation  related questions related to food.}
    \label{fig:model_perf_by_category}

  \end{figure}

\subsection{Performance by Data-Source}
\label{section:model_performance_all_datasets}
When looking at \cref{fig:model_perf_by_source_category} we see the main data-sources present in our dataset. We see that for more simple synthetic data models tended to perform better like in JTA, ShapeNet and Get3D; however for natural datasets like Objectron and NOCS\_Real, both the open and closed source models tended to have a drop in performance. This could be first attributed to the dense nature of the number of objects found in the NOCS\_Real dataset which includes a number of objects like cars and persons, the same can be said for Cityscapes which is also densely populated. For Objectron which can look at an object at different angles this can also prove difficult for most open-source models, with the closed source models tending to performing better which could be attributed to the pretraining data as these closed-source models are exposed to a large set of data with object potentially at different views making it more robust to object rotational views. 

\subsection{Task Complexity Hierarchy}

Our multi-dimensional analysis reveals a consistent difficulty pattern across orientation
reasoning tasks. This trend is broadly aligned with prior cognitive literature, which
distinguishes basic orientation perception from more demanding spatial transformation
and reference-frame reasoning processes~\cite{10.1093/acprof:oso/9780195381634.003.0004,
alma991032026461708041,vasilyeva2012development}. We use this literature to motivate
DORI's task organization, rather than to claim that model behavior directly mirrors
human cognitive development. In our experiments, models perform most competently on
frontal alignment tasks, particularly view parallelism: the top open-source models
(Qwen3-32B-Inst.\ and Qwen3-8B-Inst.\ in Tab.~3 of the main paper) achieve up to
75.8\% coarse accuracy, while strong models in Tab.~4 reach 67.5--68.5\%. Performance
generally drops as tasks require more complex transformations. For compound rotation,
open-source models peak at 49.0\% coarse accuracy (Qwen3-8B-Inst., Tab.~3), while
many models score substantially lower on granular compound-rotation questions; for
example, GPT-5-mini reaches 22.6\% on this split (Tab.~4 of the main paper), with
several models scoring well below 15

The models' difficulty with compound rotations suggests that current MLLMs do not
reliably preserve or manipulate the geometric structure needed to track objects through
multi-step transformations. This is further supported by the component-wise error
decomposition in \cref{tab:compound_rotation_decomposition}, which shows that
approximately 60\% of predictions fail on \emph{both} rotation components
simultaneously. Rather than making small errors on one axis, models often fail to
recover the combined transformation, indicating that compound rotation remains a
particularly challenging form of object-centric spatial reasoning.


A second large performance gap appears in tasks requiring different forms of relative
orientation reasoning, specifically viewer-scene direction perception and inter-object
direction perception in Tabs.~3 and 4 of the main paper. Viewer-scene direction
perception elicits relatively strong performance across model families, with top
open-source models reaching up to 94.3\% coarse accuracy (IntVL-14B-Inst., Tab.~3)
and strong models in Tab.~4 reaching around 90\% or higher. By contrast,
inter-object direction tasks are substantially more difficult: the strongest
open-source models in Tab.~3 achieve only around 20\% coarse accuracy, and many
models in Tab.~4 remain far below their viewer-scene direction performance. This
disparity suggests that models are better at recognizing orientation relative to a
viewer or scene than at reasoning about how two objects are oriented relative to one
another. This limitation is important for embodied AI applications such as robotics
and navigation, where agents must reason about object relationships from multiple
viewpoints.

Canonical orientation understanding, which requires determining whether objects appear in their expected or ``natural'' orientation, shows highly variable performance across models (Tabs.~3 and 4 of the main paper). Open-source models span a wide range, from near-chance performance (DS-1.3B-Base at 1.8\% coarse accuracy, Tab.~3) to stronger performance (Mantis-Idfs-8B at 51.2\% coarse accuracy, Tab.~3). Closed-source models also vary substantially, with several achieving high coarse canonical accuracy while still showing large coarse-to-granular gaps. This variability suggests that recognizing canonical object orientations depends on category-level visual cues, world knowledge, and training data coverage that are unevenly captured across architectures and training regimes.


\begin{figure}[ht]
    \centering
    
    \includegraphics[width=01.0\linewidth]{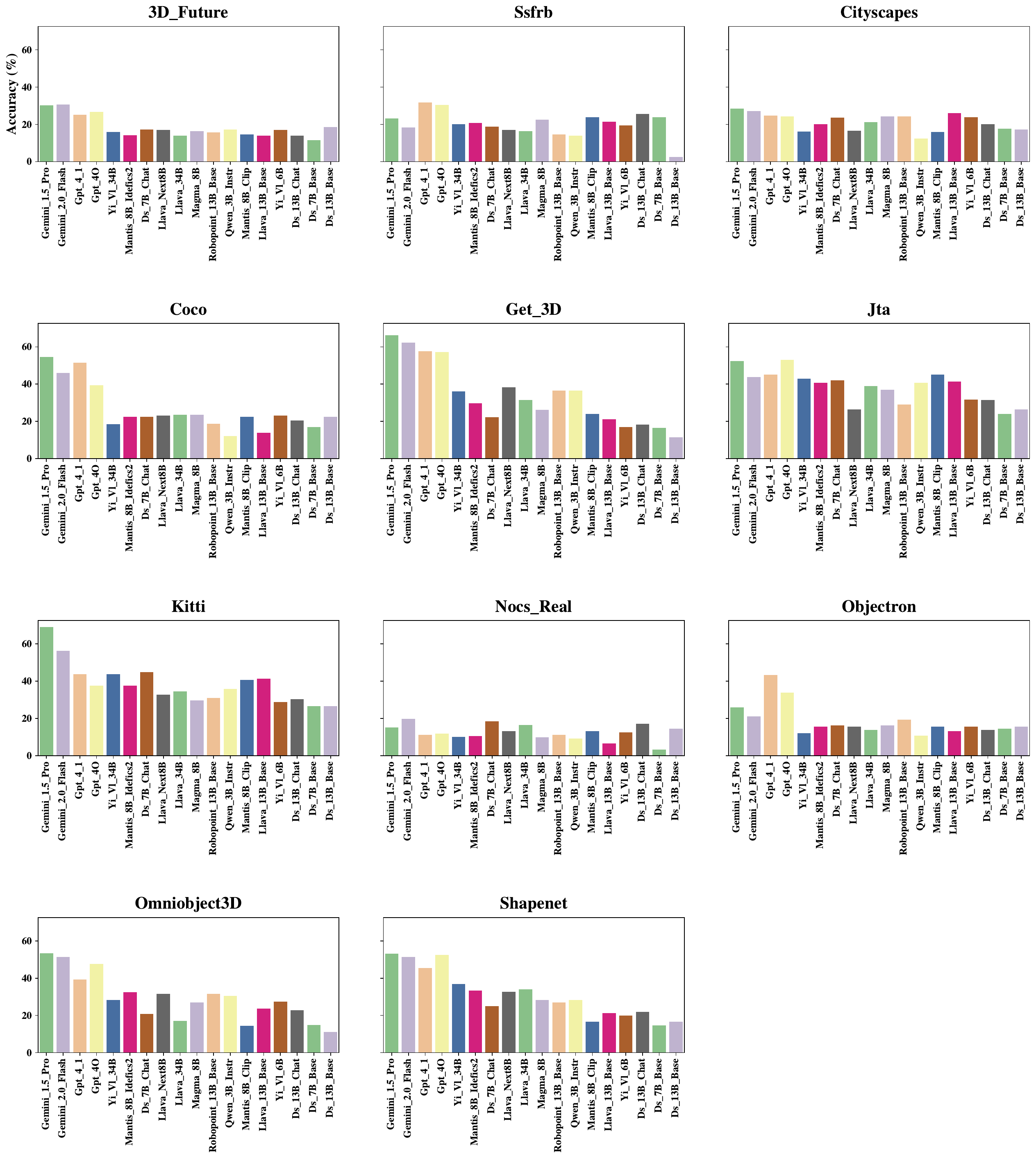}
    \caption{Model accuracy for the 18 models across the data-sources present in DORI. We see that overall the closed source models tend to perform better than the open source models. With more real and smaller datasets like NOCS\_Real and Objectron proving to be challenging for most of the models.}
    \label{fig:model_perf_by_source_category}

\end{figure}

\begin{figure}[ht]
  \centering

  \begin{subfigure}[b]{0.49\linewidth}
    \centering
    \includegraphics[width=\linewidth]
    {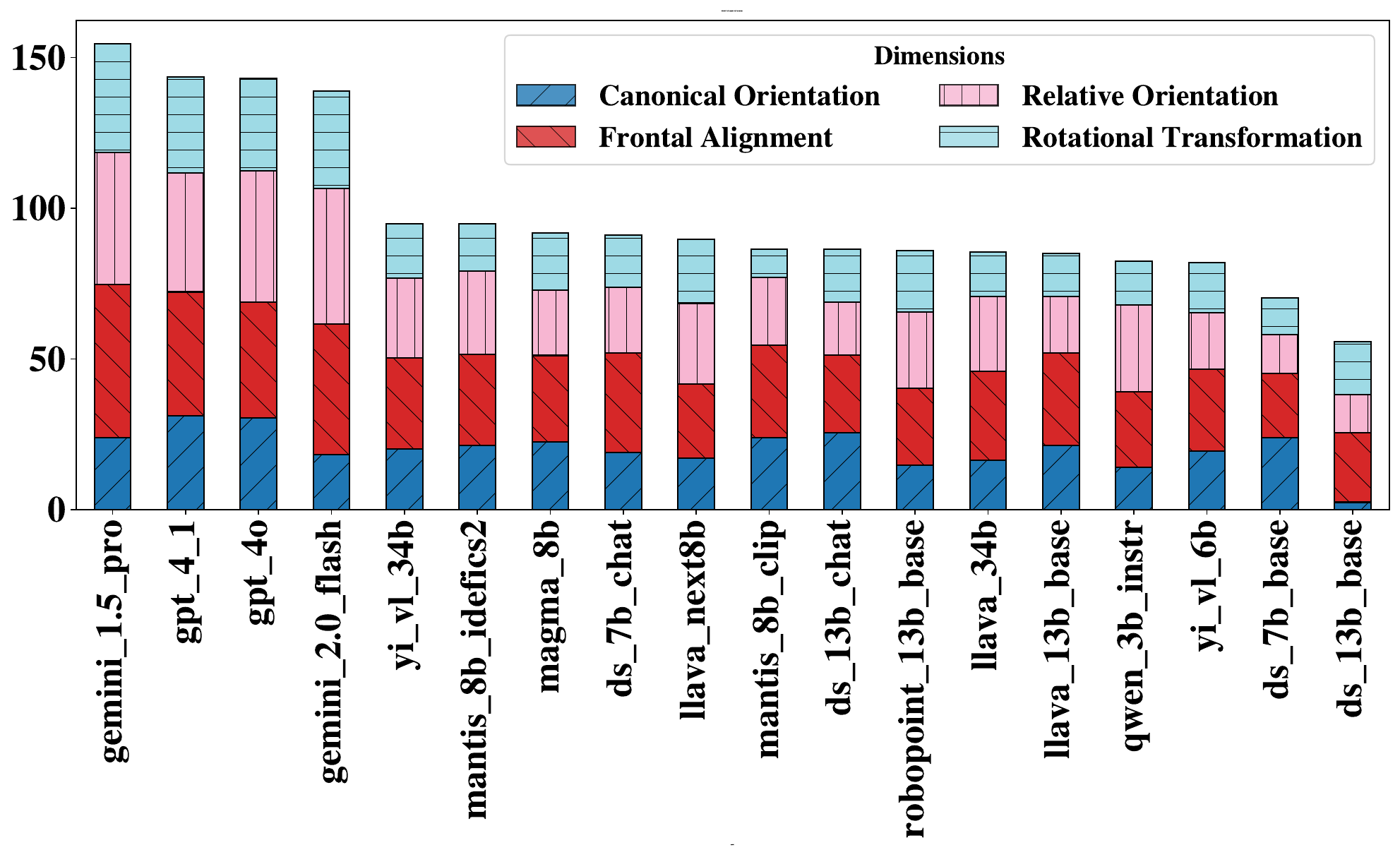}
    \caption{Performance per model across the four core dimensions of orientation.}
    \label{fig:by_dimensions}
  \end{subfigure}
  \hfill
  \begin{subfigure}[b]{0.49\linewidth}
    \centering
    \includegraphics[width=\linewidth]
    {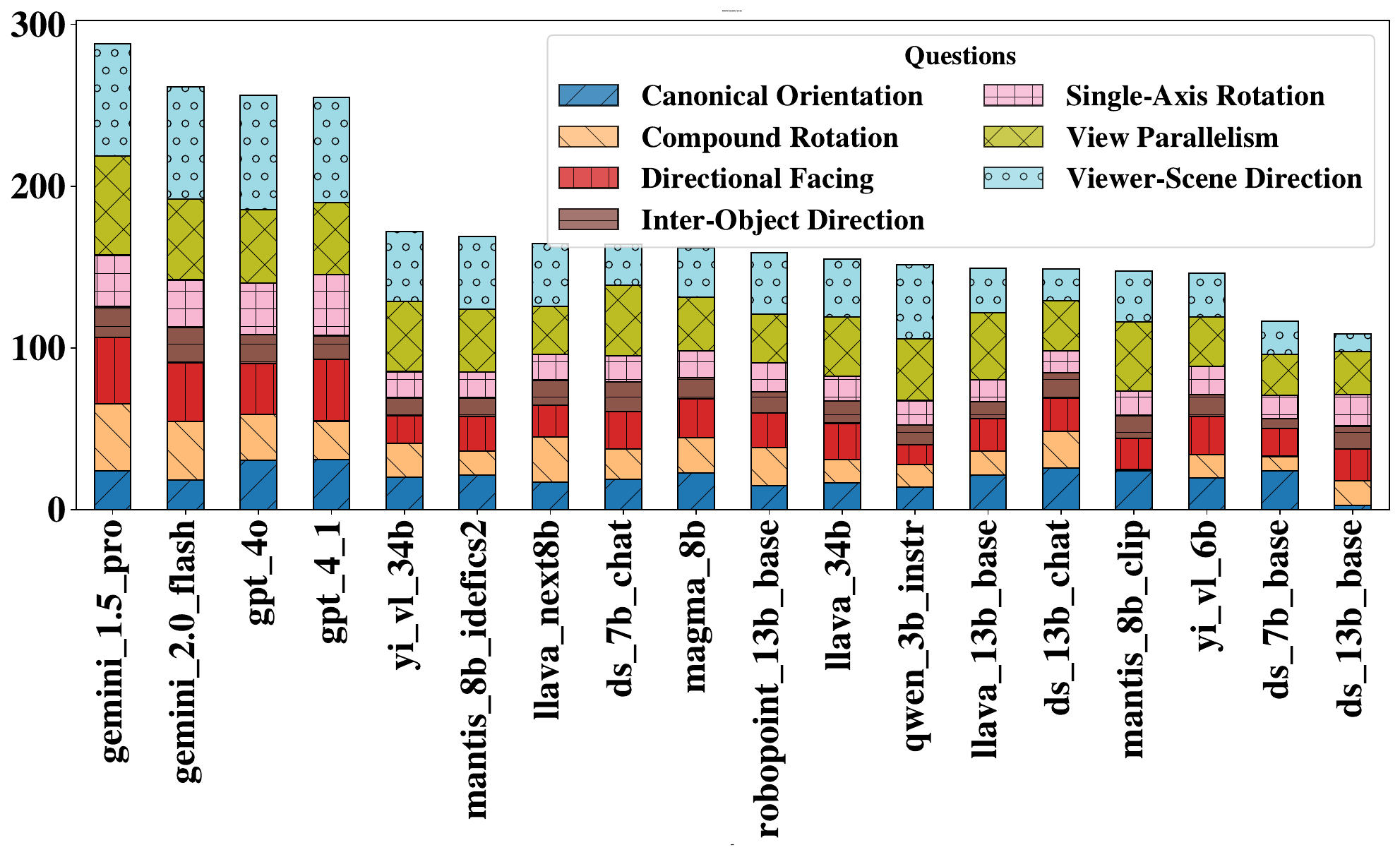}
    \caption{Performance per model across the different questions present in DORI.}
    \label{fig:by_questions}
  \end{subfigure}

  \caption{Performance of 18 leading Multimodal Large Language Models (MLLMs) different levels of our questions hierachy with \textbf{(left)} being the being the four core dimensions of orientation and \textbf{(right)} being the different questions present in DORI}
  \label{fig:model_performance_by_dimensions}
\end{figure}
\clearpage
\section{Additional Error Analysis}
\label{sec:error_analysis}

We employ a systematic approach to identify consistent failure patterns rather than random prediction mistakes by analyzing the geometric relationship between predicted and ground truth orientations. Our methodology focuses on identifying repeatable, model-agnostic spatial reasoning deficits that indicate fundamental architectural limitations.
\subsection{Systematic Error Pattern Identification} 
We define systematic failure patterns as specific ``Ground Truth $\rightarrow$ Predicted Answer'' confusion pairs that satisfy following criteria: (1) occur in more than 5\% of a model's incorrect predictions, and (2) are observed across at least 3 different model architectures. This threshold-based approach filters out infrequent, stochastic mistakes to focus on consistent failure modes that indicate underlying spatial reasoning deficits rather than random prediction errors.
We categorize confusion patterns by their geometric and cognitive properties:
\begin{itemize}
    \item \textbf{Perpendicular Confusion}: Systematic misclassification between parallel (0°--15°) and perpendicular (65°--95°) orientations, indicating categorical spatial representation
    \item \textbf{Directional Reversal}: Confusion between opposite directions (\eg, left$\leftrightarrow$right, toward$\leftrightarrow$away), suggesting directional processing failures
    \item \textbf{Angle Compression}: Tendency to predict intermediate angles when extreme angles are correct, indicating quantization of continuous spatial information
    \item \textbf{Frontal Bias}: Over-prediction of frontal/canonical orientations regardless of true orientation, likely reflecting training data distribution bias
    \item \textbf{Rotational Symmetry Confusion}: Systematic confusion between rotational equivalents (\eg, 90°$\leftrightarrow$270°, quarter vs. three-quarter turns)
    \item \textbf{Uncertainty Cascade}: Inappropriate uncertainty responses when spatial ambiguity should trigger systematic reasoning attempts
\end{itemize}
Below, \cref{tab:systematic_errors} presents the complete systematic error analysis across all spatial reasoning dimensions evaluated in DORI. The patterns reveal fundamental limitations in how current MLLMs process and represent spatial information.

\begin{table}[!htbp]
\centering
\caption{Systematic Error Patterns Across All DORI Questions}
\label{tab:systematic_errors}
\resizebox{\textwidth}{!}{%
\begin{tabular}{@{}llrrp{4cm}@{}}
\toprule
\textbf{Question Type} & \textbf{True Answer} & \textbf{Predicted Answer} & \textbf{Error (\%)} & \textbf{Pattern Type} \\
\midrule
View Parallelism & 0°--15° & 65°--95° & 20.5 & Perpendicular Confusion \\
View Parallelism & 65°--95° & 0°--15° & 10.3 & Perpendicular Confusion \\
\midrule
Directional Facing & 30° left & 30° right & 8.3 & Directional Reversal \\
Directional Facing & 180° (away) & 0° (facing camera) & 14.7 & Directional Reversal \\
Inter-object Direction & 46--90° CCW & 46--90° CW & 5.7 & Directional Reversal \\
\midrule
View Parallelism & 135°--180° & 65°--95° & 15.8 & Angle Compression \\
Single-axis Rotation & 90° & 45° & 7.6 & Angle Compression \\
\midrule
Directional Facing & 30° left & 0° (facing camera) & 15.6 & Frontal Bias \\
Directional Facing & 30° right & 0° (facing camera) & 13.7 & Frontal Bias \\
Canonical Orientation & Cannot determine & No change needed & 19.5 & Frontal Bias \\
\midrule
Viewer-scene Rotation & 270° & 90° & 13.1 & Rot. Symmetry Confusion \\
Viewer-scene Rotation & 180° & 90° & 11.6 & Rot. Symmetry Confusion \\
Single-axis Rotation & 0° & 90° & 9.2 & Rot. Symmetry Confusion \\
\midrule
View Parallelism & 0°--15° & Cannot determine & 12.3 & Uncertainty Cascade \\
Canonical Orientation & Cannot determine & UNKNOWN & 23.5 & Uncertainty Cascade \\
Viewer-scene Rotation & 270° & Cannot determine & 12.0 & Uncertainty Cascade \\
\bottomrule
\end{tabular}%
}
\end{table}
\paragraph{Results: } Our analysis reveals six systematic error categories affecting all evaluated models. The most severe is perpendicular confusion, where models systematically misclassify parallel (0°--15°) and perpendicular (65°--95°) orientations with 20.5\% and 10.3\% error rates respectively, indicating coarse categorical rather than continuous angular encoding. Models also demonstrate consistent directional processing failures including left-right confusion (8.3\%), front-back reversal (14.7\%), and clockwise-counterclockwise errors (5.7\%), suggesting fundamental directional encoding limitations. Additional patterns include angle compression where extreme positions are systematically predicted as intermediate values (7.6--15.8\% error rates), frontal bias reflecting training data distribution effects (13.7--19.5\% over-prediction of frontal orientations), rotational symmetry confusions particularly between quarter and three-quarter turns (9.2--13.1\% error rates), and uncertainty cascade failures where models inappropriately handle spatial ambiguity (12.0--23.5\% error rates). These systematic patterns across diverse architectures indicate fundamental limitations in current MLLM spatial processing mechanisms rather than model-specific deficits.
\subsection{Component-wise Error Decomposition} Compound rotation tasks (Q4 in DORI) present the most cognitively demanding spatial reasoning challenge, requiring models to track sequential 3D transformations around multiple axes. To understand the specific failure modes, we decompose compound rotation errors into orthogonal components that isolate different aspects of spatial transformation understanding.We analyze three distinct error categories that provide insight into different failure modes:
\begin{enumerate}
    \item \textbf{Order Swap Errors}: Models correctly identify both rotation components but reverse their sequence (\eg, ground truth ``90° Horizontal then 180° Vertical'' predicted as ``180° Horizontal then 90° Vertical''). This isolates sequence understanding from content understanding.
    
    \item \textbf{Component Accuracy}: Percentage of predictions where individual rotation components (horizontal or vertical) are correctly identified regardless of the accuracy of the other component. This measures partial understanding capabilities.
    
    \item \textbf{Complete Joint Failure}: Percentage of predictions where both rotation components are incorrectly predicted, indicating total breakdown of 3D spatial reasoning.
\end{enumerate}
\cref{tab:compound_rotation_decomposition} presents the complete component-wise error decomposition across all evaluated models, revealing distinct failure patterns and architectural effects.

\begin{table}[tbp]
\centering
\caption{Component-wise Error Decomposition for Compound Rotations (Q4)}
\label{tab:compound_rotation_decomposition}
\resizebox{\textwidth}{!}{%
\begin{tabular}{@{}lrrrrrr@{}}
\toprule
\textbf{Model} & \textbf{Order Swap} & \textbf{Horizontal Acc.} & \textbf{Vertical Acc.} & \textbf{Both Wrong} \\
 & \textbf{(\%)} & \textbf{(\%)} & \textbf{(\%)} & \textbf{(\%)} & \\
\midrule
Qwen-3B-instr & 0.2 ± 0.1 & 22.7 ± 2.1 & 21.4 ± 1.9 & 55.9 ± 3.2  \\
LLaVA-13B-base & 3.8 ± 0.7 & 21.3 ± 1.8 & 19.5 ± 1.6 & 59.2 ± 2.9  \\
DeepSeek-7B-chat & 4.6 ± 0.8 & 20.4 ± 1.7 & 20.5 ± 1.7 & 59.1 ± 2.8  \\
DeepSeek-1.3B-chat & 4.2 ± 0.7 & 20.4 ± 1.6 & 19.7 ± 1.5 & 59.9 ± 2.7   \\
LLaVA-Next-8B & 4.9 ± 0.9 & 20.7 ± 1.8 & 19.6 ± 1.6 & 59.6 ± 2.9\\
Magma-8B & 4.6 ± 0.8 & 20.1 ± 1.6 & 20.0 ± 1.6 & 59.9 ± 2.8  \\
DeepSeek-1.3B-base & 4.7 ± 0.8 & 19.9 ± 1.5 & 20.0 ± 1.6 & 60.1 ± 2.8   \\
DeepSeek-7B-base & 1.2 ± 0.3 & 19.7 ± 1.5 & 20.1 ± 1.6 & 60.3 ± 2.8   \\
\midrule
\textbf{Average} & 3.5 ± 1.7 & 20.6 ± 0.9 & 20.1 ± 0.6 & 59.3 ± 1.4   \\
\bottomrule
\end{tabular}%
}
\end{table}
\paragraph{Results: }The component-wise decomposition reveals a fundamental dissociation between sequence processing and spatial reasoning in MLLMs. While order swap errors remain low across all models (mean: 3.5\%), indicating competent instruction following, approximately 60\% of predictions fail on both rotation components, representing near-complete breakdown of 3D spatial processing. . Models demonstrate axis-agnostic processing (horizontal vs. vertical accuracy differential: 0.7\%), unlike human embodied cognition, suggesting identical mechanisms for all rotational transformations rather than specialized processing pathways. Notably, instruction-tuned models significantly outperform larger base models. Qwen-3B-instr achieves superior performance across all metrics despite smaller parameter count, and DeepSeek-1.3B-chat outperforms DeepSeek-7B-base with 5.4x fewer parameters. These findings indicate three critical limitations: (1) models can manipulate rotation symbols but lack geometric transformation mechanisms, (2) the consistent ~60\% joint failure rate across diverse architectures suggests fundamental rather than model-specific deficits, and (3) training methodology appears more crucial than parameter scaling for spatial reasoning capabilities, suggesting targeted training approaches may be more effective than architectural scaling.

\subsection{Soft Accuracy Calculation}
\label{sec:soft_accuracy}
Soft accuracy metrics were introduced to provide a more nuanced evaluation of models' orientation understanding capabilities and help distinguish between models that are completely wrong versus those that have an approximate understanding of orientation concepts. Unlike standard binary accuracy that only awards points for exact matches, soft accuracy awards half points $(0.5)$ for answers that are partially correct or adjacent to the ground truth. To calculate such accuracies, we implement carefully designed spatial tolerance thresholds and logical equivalences. Soft accuracy is only calculated for the fine-grained questions that typically demand a precise metric response.
\begin{itemize}
    \item For \textbf{View Parallelism}, Predictions within $\pm{45}^{\circ}$ of the ground truth angle receive partial credit. This threshold captures predictions in adjacent sectors while excluding opposed orientations. 
    \item For \textbf{Directional Facing}, half points are awarded exclusively for mirror-image confusions between "30 degrees left" and "30 degrees right". We intentionally do not extend partial credit to other angular errors, preserving the specificity of directional understanding assessment.
    \item For \textbf{Single-axis Rotatoin}, a ${45}^{\circ}$ tolerance window applies to predicted rotations $({0}^{\circ}, {45}^{\circ}, {90}^{\circ}, {135}^{\circ}, {180}^{\circ})$.  This allows credit for adjacent discrete positions while maintaining distinction between major orientation categories. For instance, predicting ${45}^{\circ}$ when the correct answer is ${90}^{\circ}$ would not qualify, but a ${135}^{\circ}$ prediction for ${180}^{\circ}$ ground truth would receive partial credit.
    \item In \textbf{Compound Rotation}, partial credit is awarded if either the horizontal or vertical rotation component is correct in multi-axis transformations. In this "X then Y" rotation sequence response, we parse both components separately. For example, prediction of "${90}^{\circ}$ horizontal then ${0}^{\circ}$ vertical" would receive 0.5 points for either correct component when compared to the ground truth "${90}^{\circ}$ horizontal then ${180}^{\circ}$ vertical"
    \item For \textbf{Inter-object Direction}, half points are given for adjacent magnitude ranges, but only if the direction (clockwise vs. counterclockwise) matches. For instance, if the ground truth was "0 to 45 degrees clockwise" and the prediction was "46 to 90 degrees clockwise", we award 0.5 points. However, if the prediction was "46 to 90 degrees counterclockwise," it would earn 0 points despite similar magnitude
    \item For \textbf{Viewer-scene direction}, the soft accuracy specifically addresses confusion between opposite rotational directions. For example, if an object has rotated ${90}^{\circ}$ clockwise between images, but the model reports ${270}^{\circ}$ clockwise (which is equivalent to ${90}^{\circ}$ counterclockwise), it receives $0.5$ points. No partial credit is given for other angle confusions.
    \item For \textbf{Canonical Orientation}, the soft accuracy addresses confusion in the order of operations. For example, if an image requires rotation followed by flipping to restore its canonical orientation, but the model suggests flipping followed by rotation, it receives 0.5 points.
\end{itemize}

\textbf{Results.} \cref{fig:q1_standard_vs_soft_accuracy},\cref{fig:q2_standard_vs_soft_accuracy},\cref{fig:q3_standard_vs_soft_accuracy},\cref{fig:q4_standard_vs_soft_accuracy},\cref{fig:q5_standard_vs_soft_accuracy},\cref{fig:q6_standard_vs_soft_accuracy},\cref{fig:q7_standard_vs_soft_accuracy}, compares standard (hard) vs.\ soft accuracy on DORI questions. 

\cref{fig:q1_standard_vs_soft_accuracy} and \cref{fig:q2_standard_vs_soft_accuracy} reveal that for View Parallelism and Directional Facing tasks, soft accuracy provides no benefit over standard accuracy, with identical performance metrics across all models. This indicates that when models err on these fundamental orientation tasks, they tend to make categorical mistakes rather than near-miss approximations. The lack of improvement suggests that errors in these tasks stem from fundamental misunderstandings rather than subtle misjudgments.

In contrast, the\textbf{ Single-axis Rotation} task (\cref{fig:q3_standard_vs_soft_accuracy}) shows the most substantial gains under soft accuracy metrics, with improvements ranging from 8.6\% to 17.7\% across models. DeepSeek-7B-Chat achieves the most dramatic improvement, a remarkable 17.7\% gain. LLaVA-Next-8B improves of  +11.5\%, while Qwen-3B-Instruct shows improvement from + 16.4\%. These substantial gains suggest that while models often fail to identify the exact rotational angle, they frequently select adjacent angular categories, demonstrating partial understanding of rotational relationships.

For Compound Rotation ( \cref{fig:q4_standard_vs_soft_accuracy}), all models except Qwe2.5-3B-Instruct (0.5\% improvement) show notable improvements under soft accuracy metrics. Both LLaVA-7B-Chat and LLava-Next-8B improves from by +14.6\%, indicating that models often correctly identify one of the two rotation components (horizontal or vertical) while missing the other. 
This partial success highlights both the inherent complexity of multi-axis rotations and the models' fragmentary grasp of compound transformations.

The Inter-object Direction task ( \cref{fig:q5_standard_vs_soft_accuracy}) shows similar amount of soft accuracy gains across all models, with improvements ranging from 7.7\% to 17.1\%. DeepSeek-7B-Chat improves from 12.7\% to 29.8\%, more than doubling its effective performance. This suggests that models often select directionally appropriate answers that fall into adjacent angular ranges, indicating a coarse understanding of relative orientations despite lacking precise angular discrimination.

For Viewer-scene Direction ( \cref{fig:q6_standard_vs_soft_accuracy}), soft accuracy provides moderate improvements, with DeepSeek-7B-Chat showing the largest gain. The comparatively smaller improvements here suggest that models correctly identify rotational changes, with fewer "near miss" responses than in other tasks.

Canonical Orientation ( \cref{fig:q7_standard_vs_soft_accuracy}) shows minimal improvement under soft accuracy metrics. The small gains observed suggest that models are rarely confused in the order of operations. When they fail on canonical orientation tasks, they typically misidentify the necessary operations entirely rather than simply reversing their order. This indicates a more fundamental gap in understanding canonical object positioning rather than mere sequencing errors.
\begin{figure}[ht]
    \centering
    \includegraphics[width=0.8\linewidth]{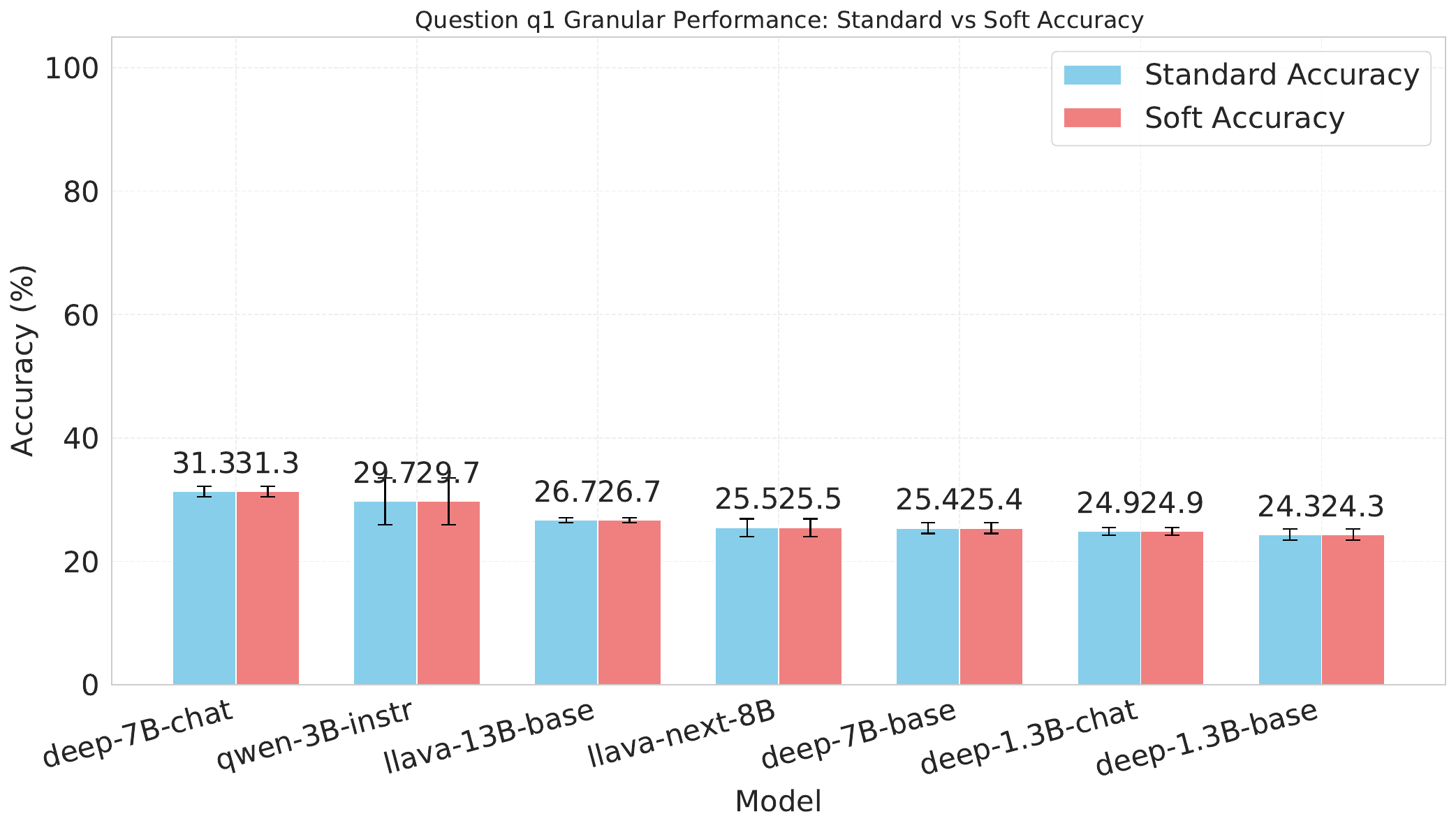}
    \caption{Comparing the mean and standard deviation of soft vs.\ standard (hard) accuracy on View Parallelism task.  See discussion Sec.\cref{sec:soft_accuracy}.}
    \label{fig:q1_standard_vs_soft_accuracy}
\end{figure}

\begin{figure}[ht]
    \centering
    \includegraphics[width=0.8\linewidth]{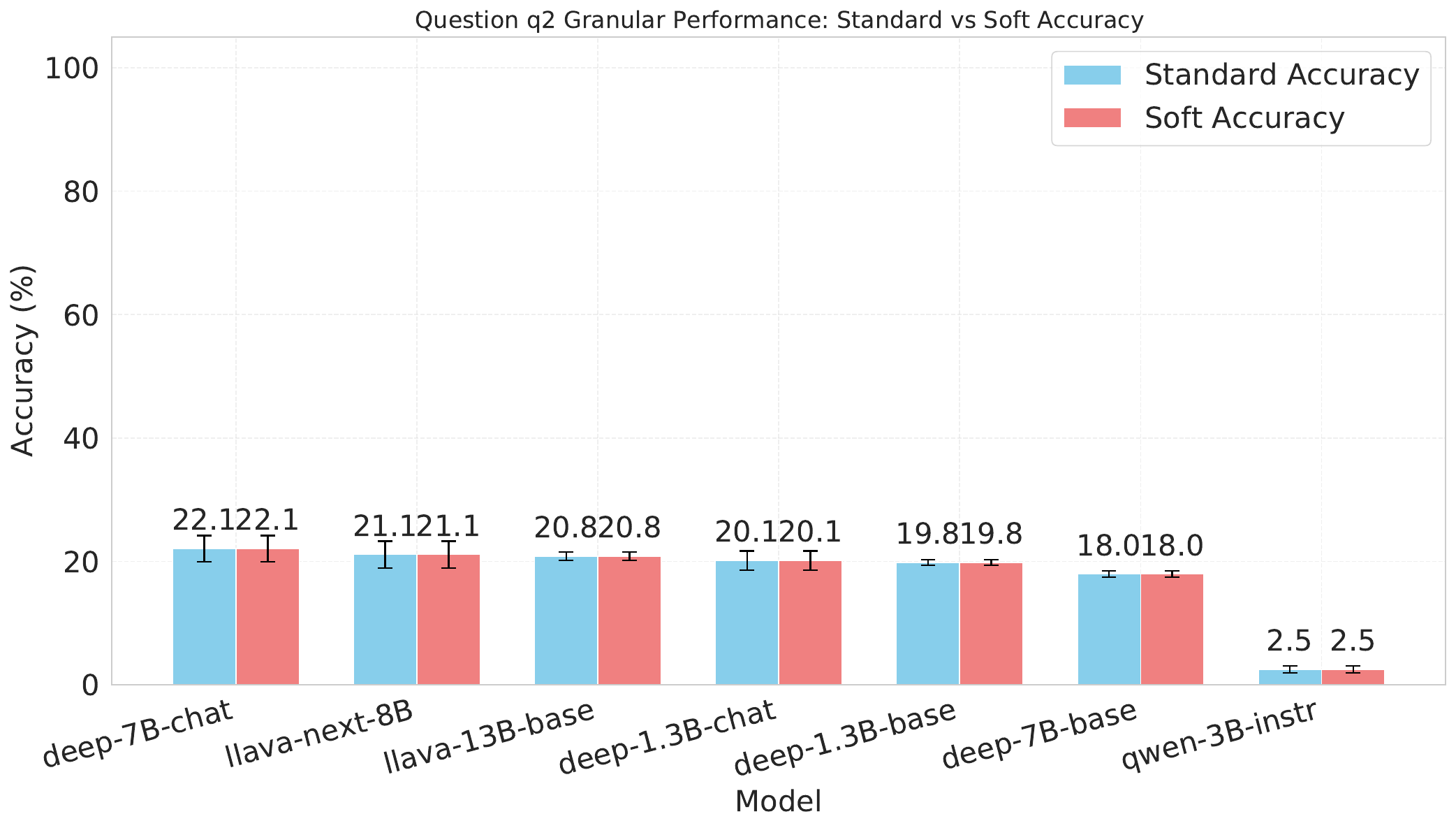}
    \caption{Comparing the mean and standard deviation of soft vs.\ standard (hard) accuracy on Directional Facing task.  See discussion Sec.\cref{sec:soft_accuracy}.}
    \label{fig:q2_standard_vs_soft_accuracy}
\end{figure}
\begin{figure}[ht]
    \centering
    \includegraphics[width=0.8\linewidth]{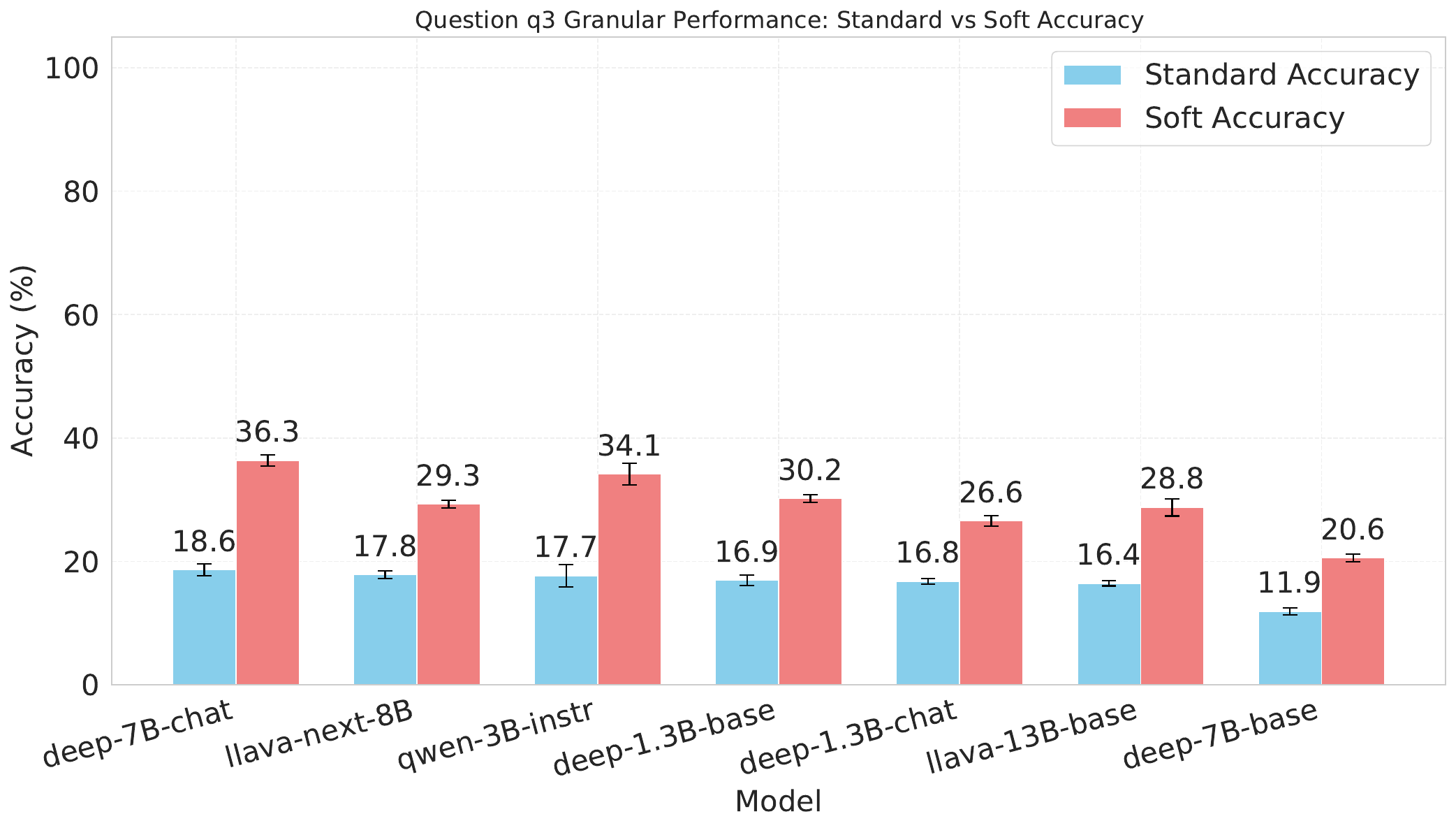}
    \caption{Comparing the mean and standard deviation of soft vs.\ standard (hard) accuracy on Single-axis Rotation task.  See discussion Sec.\cref{sec:soft_accuracy}.}
    \label{fig:q3_standard_vs_soft_accuracy}
\end{figure}
\begin{figure}[ht]
    \centering
    \includegraphics[width=0.8\linewidth]{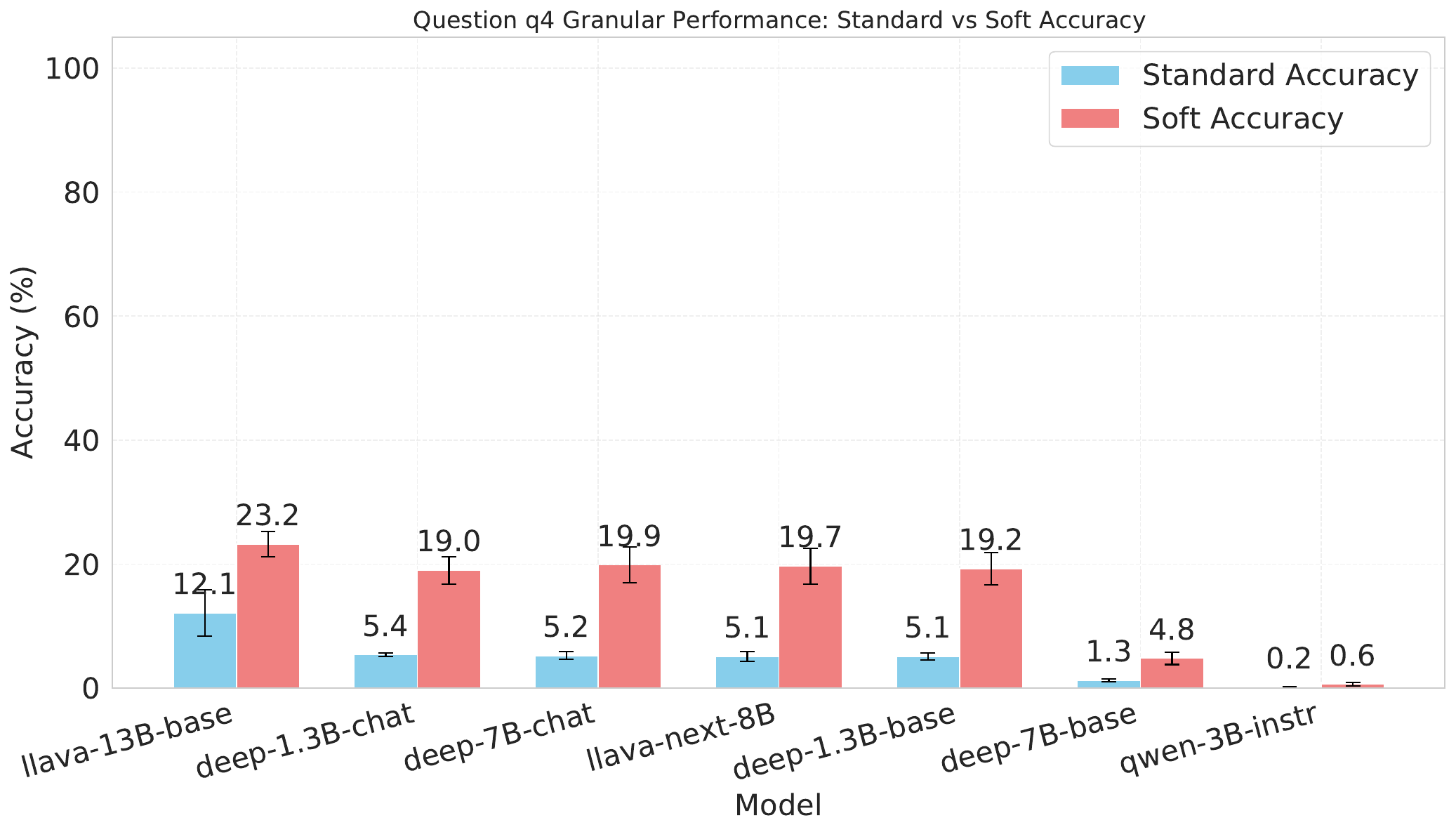}
    \caption{Comparing the mean and standard deviation of soft vs.\ standard (hard) accuracy on Compound Rotation task.  See discussion Sec.\cref{sec:soft_accuracy}.}
    \label{fig:q4_standard_vs_soft_accuracy}
\end{figure}

\begin{figure}[ht]
    \centering
    \includegraphics[width=0.8\linewidth]{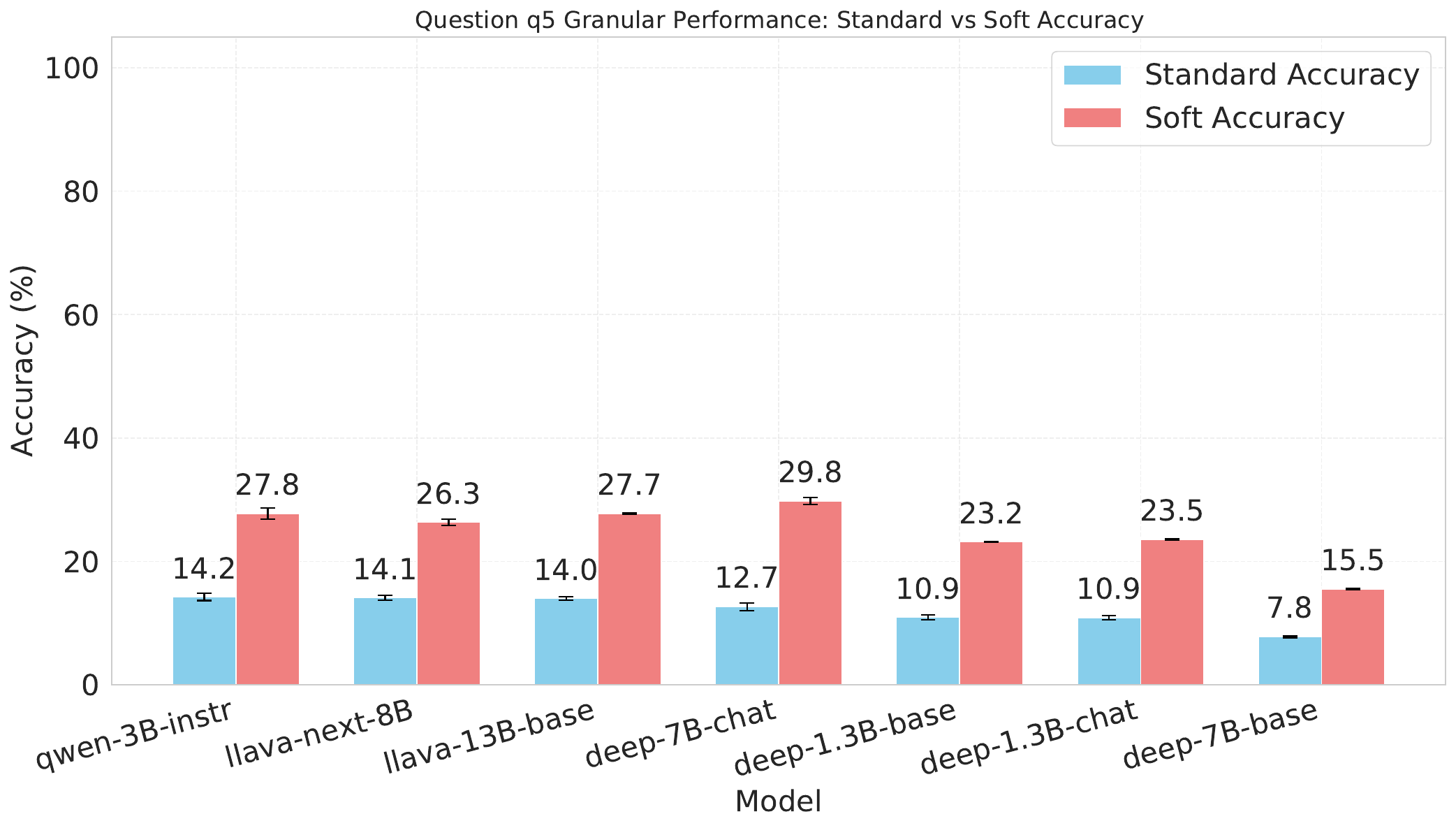}
    \caption{Comparing the mean and standard deviation of soft vs.\ standard (hard) accuracy on  Inter-object Direction task.  See discussion Sec.\cref{sec:soft_accuracy}.}
    \label{fig:q5_standard_vs_soft_accuracy}
\end{figure}

\begin{figure}[ht]
    \centering
    \includegraphics[width=0.8\linewidth]{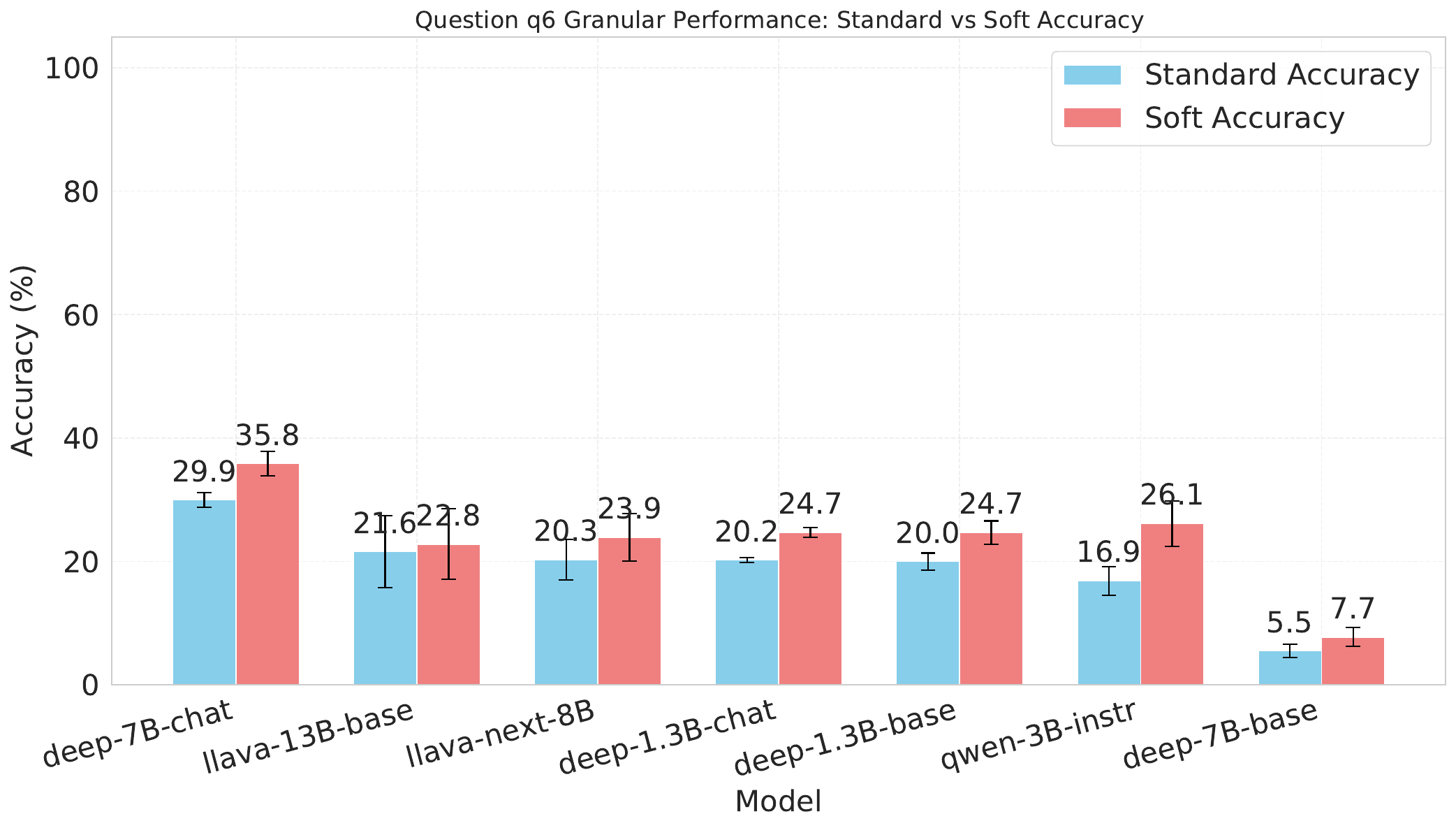}
    \caption{Comparing the mean and standard deviation of soft vs.\ standard (hard) accuracy on  Viewer-scene Direction task.  See discussion Sec.\cref{sec:soft_accuracy}.}
    \label{fig:q6_standard_vs_soft_accuracy}
\end{figure}

\begin{figure}[ht]
    \centering
    \includegraphics[width=0.8\linewidth]{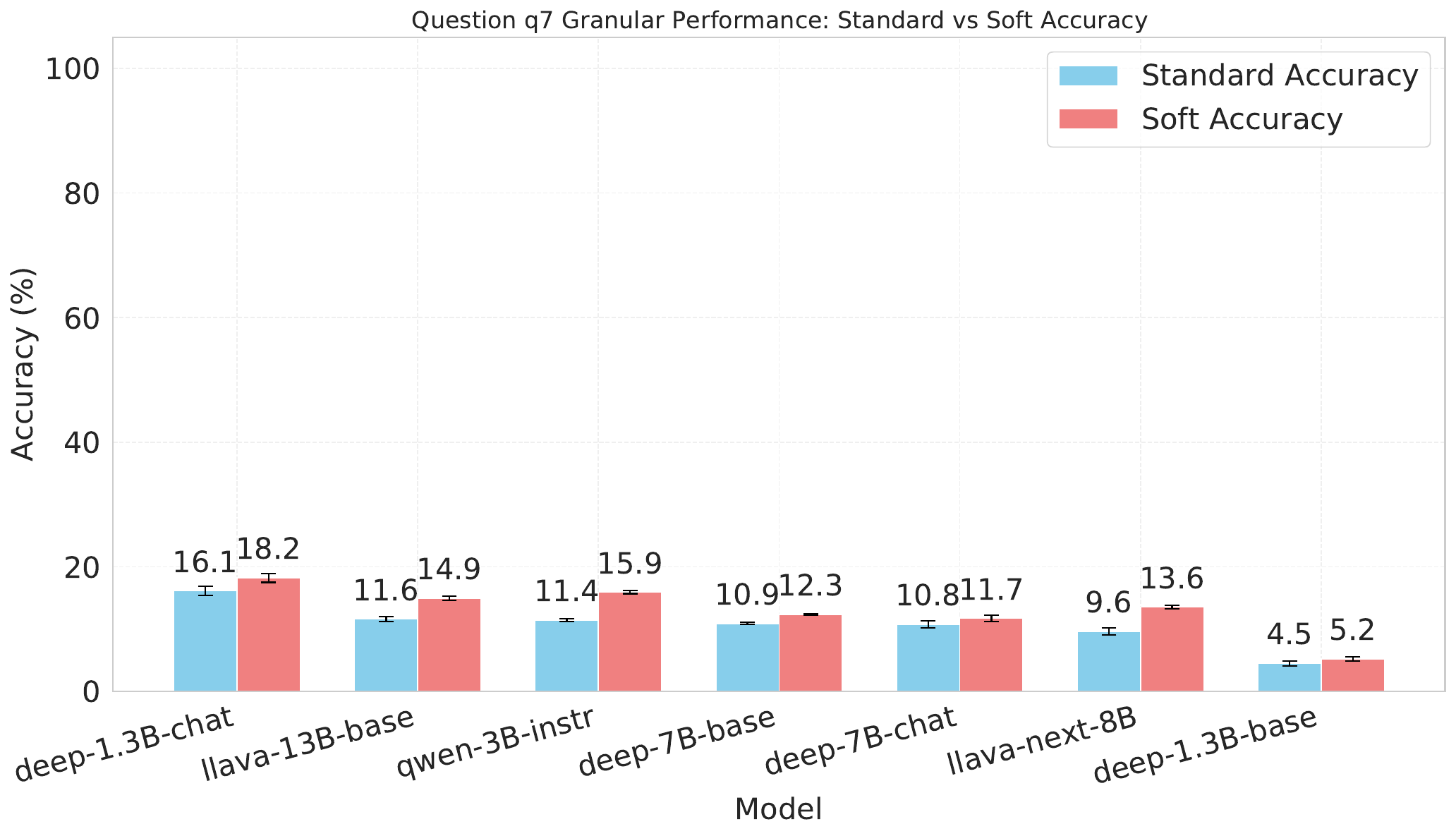}
    \caption{Comparing the mean and standard deviation of soft vs.\ standard (hard) accuracy on Canonical Orientation task.  See discussion Sec.\cref{sec:soft_accuracy}.}
    \label{fig:q7_standard_vs_soft_accuracy}
\end{figure}
\cref{fig:accuracy_improvement_heatmap}
summarizes the relative gains of soft accuracy across all tasks and models. Notably, this more lenient metric only helps some questions ( \cref{fig:q3_standard_vs_soft_accuracy} to \cref{fig:q7_standard_vs_soft_accuracy}). The heatmap reveals that Single-axis Rotation, Compound Rotation, and Inter-object Direction tasks benefit most from soft accuracy metrics, with improvements frequently exceeding 10\%. This pattern suggests that rotational and relational orientation understanding in current MLLMs exists on a spectrum rather than in binary states of correctness.  In addition, the highest gain reported was for Single-axis Rotation, but is significantly below human performance using standard accuracy (18\% vs.\ avg of about 30\%).  This shows that even when given an advantage, these models still fall significantly below human ability.

\begin{figure}[ht]
    \centering
    \includegraphics[width=\linewidth]{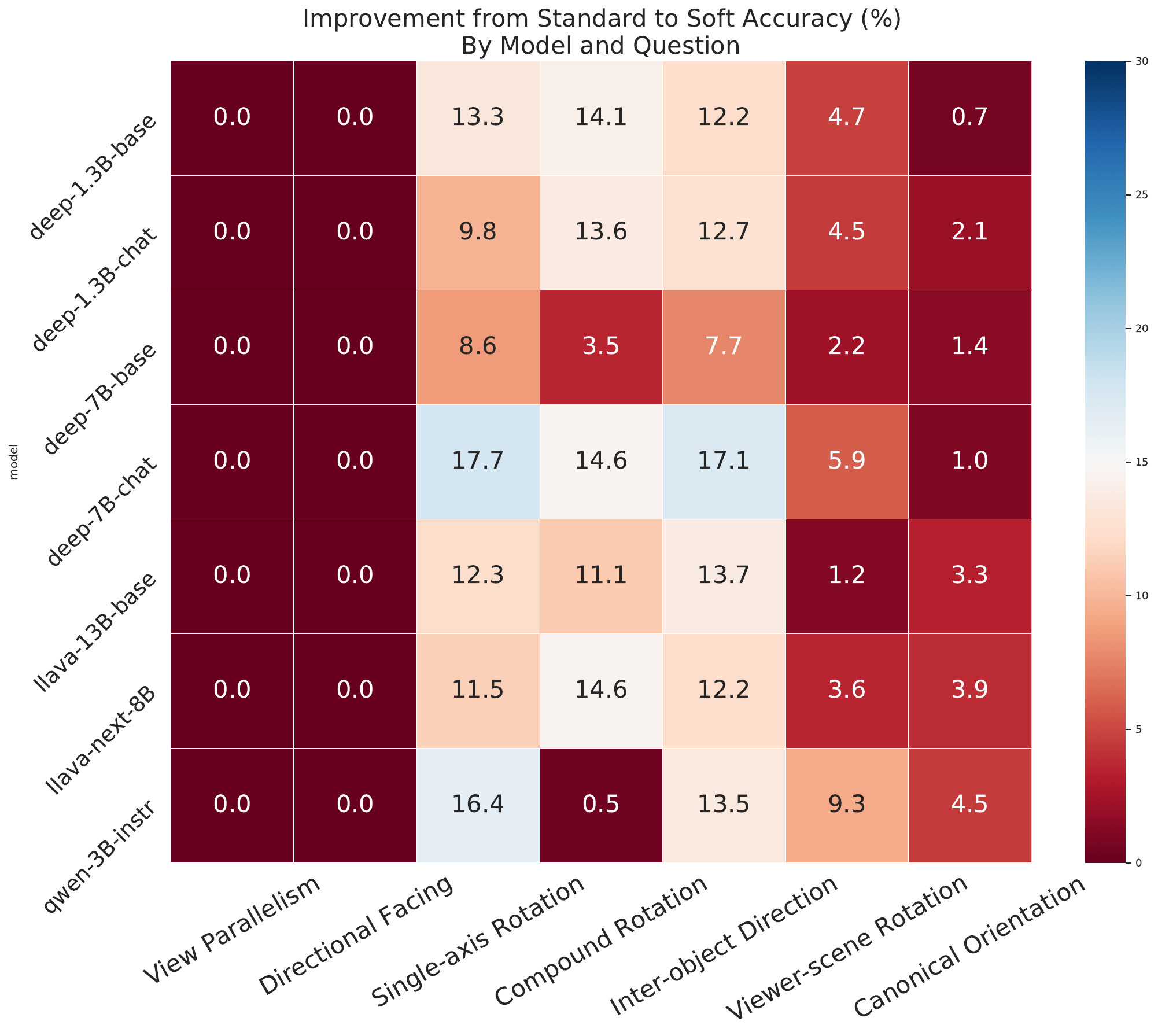}
    \caption{Relative gain from using Soft accuracy rather than Standard (hard) accuracy per question.  See\cref{sec:soft_accuracy} for discussion}
    \label{fig:accuracy_improvement_heatmap}
\end{figure}

\subsection{Performance Error Bars}
\label{sec:model_var}
The error bars in \cref{fig:q1_performance},\cref{fig:q2_performance},\cref{fig:q3_performance},\cref{fig:q4_performance},\cref{fig:q5_performance},\cref{fig:q6_performance}, and \cref{fig:q7_performance} report the mean and standard deviation of various models on DORI questions. For each model and question type combination, the formula $error = std\_accuracy / sqrt(seed\_count)$ was applied. We used $3$ different seeds: $[42, 1998, 107983]$.

Looking at the \textbf{View Parallelism} task (\cref{fig:q1_performance}), we observe relatively narrow error bars for most models, indicating consistent performance across different seeds. However, both DeepSeek base models and the LLaVa-13 B-base models display wider error bars on coarse questions (approximately $\pm3-4\%$), suggesting that their performance comes with greater variability. Despite LLaVa-13B-base demonstrating better performance than the other 2 base models, it's clear that all three base models' performance is more sensitive to initialization conditions. Qwen2.5-3B-instruct model demonstrates notable initialization variability for both coarse and granular questions.

The \textbf{Directional Facing} task (\cref{fig:q2_performance}) reveals generally smaller error bars across all models, with most variations under $\pm2\%$, indicating that performance on cardinal direction assessment remains relatively stable regardless of initialization. However, all models perform notably worse on this task compared to View Parallelism, with even the best model (DeepSeek-7B-Chat) achieving only 32.5\% accuracy on coarse questions. LLaVA-Next-8B shows consistent but lower performance, while Qwen2.5-3B-Instruct has low performance with tight error bars.

The \textbf{Single-axis Rotation} task (\cref{fig:q3_performance}) is more uniform across models, with most showing variations of $\pm2-3\%$. The comparable error bar sizes across models suggest that this task presents similar levels of difficulty for all architectures, with no model demonstrating significantly more stable performance than others. This uniformity in variability indicates that improvements in this task may require fundamental architectural innovations rather than just parameter tuning.


The \textbf{Compound Rotation} task (\cref{fig:q4_performance}) shows relatively narrow error bars across most models, indicating that performance on this task is stable across initialization seeds rather than highly variable. However, the coarse and granular results reveal a large coarse-to-granular gap: several models can recognize whether an image is broadly canonical, but still struggle to infer the precise transformations needed to restore canonical orientation. This suggests that models may learn category-level canonicality cues while lacking robust representations of the geometric operations underlying non-canonical object poses.

For \textbf{Inter-object Direction} (\cref{fig:q5_performance}), we observe slightly asymmetric performance patterns between coarse and granular questions. While DeepSeek-7B-Chat achieves the highest coarse accuracy, all models show substantially lower performance on granular questions (generally below 15\%). The error bars appear to be tight for coarse and granular questions, indicating consistent performance across different trials. The performance gap between coarse and granular questions suggests that while models can sometimes succeed at basic relational orientation tasks (determining if objects face the same/opposite directions), they systematically fail when asked to make precise angular judgments about inter-object relationships. Interestingly, for LLaVA-13B-base model, the granular performance slightly exceeds its coarse performance, running counter to the typical pattern observed in other tasks and models. This anomaly may indicate that LLaVA's training regime potentially encodes some specific features that assist with fine-grained angular estimations between objects, though its overall performance remains well below human capabilities on these tasks.

Similarly, the \textbf{Viewer-Scene Direction} task (\cref{fig:q6_performance}) reveals intriguing performance inversions between coarse and granular questions for certain models. Qwen-3B-Instruct shows the highest coarse accuracy with poor granular performance, while DeepSeek-7B-Chat demonstrates the opposite pattern. These inversions, coupled with wide error bars, indicate that different model architectures encode rotation perception in fundamentally different ways.  This task exposes fundamental inconsistencies in how current MLLMs process orientation changes, suggesting that rotation tracking may rely on different computational mechanisms than static orientation perception, with these mechanisms developing unevenly across model architectures and training regimes.

The \textbf{Canonical Orientation} task (\cref{fig:q7_performance}) doesn't exhibit any high variability among the error bars across coarse and granular performance. The error bars remain relatively narrow for most models ($\pm1-3\%$), indicating that performance limitations on this task are consistent across initialization seeds rather than highly variable. This consistency, coupled with generally poor performance, suggests that canonical orientation understanding, which requires both world knowledge about natural object positions and spatial transformation reasoning, represents a fundamental capability gap in current MLLMs. The results indicate that models lack robust internal representations of how objects "should" appear in the world, a crucial component for embodied navigation and manipulation tasks where recognizing and correcting non-canonical orientations is essential.

\noindent \paragraph{Cross-Task Insights} Across all tasks, our error bar analysis reveals several critical insights about MLLMs' orientation reasoning capabilities. First, model performance stability varies substantially across tasks, with simpler perception tasks (View Parallelism, Directional Facing) showing more consistent performance across initializations compared to complex reasoning tasks (Compound Rotation, Viewer-scene direction, Canonical Orientation). Moreover, the consistently tight error bars on poor-performing granular questions, particularly for rotational tasks, indicate systematic limitations rather than chance variability, suggesting architectural rather than parametric constraints.  We also observe that Chat-tuned models generally demonstrate more stable performance than their base counterparts, suggesting that instruction tuning not only improves accuracy but also reduces initialization sensitivity. Finally, the performance inversions observed between coarse and granular questions for some models highlight the disconnect between categorical and precise metric orientation understanding, a fundamental challenge that persists across all model families and architectures evaluated.
\begin{figure}[ht]
    \centering
    \includegraphics[width=0.8\linewidth]{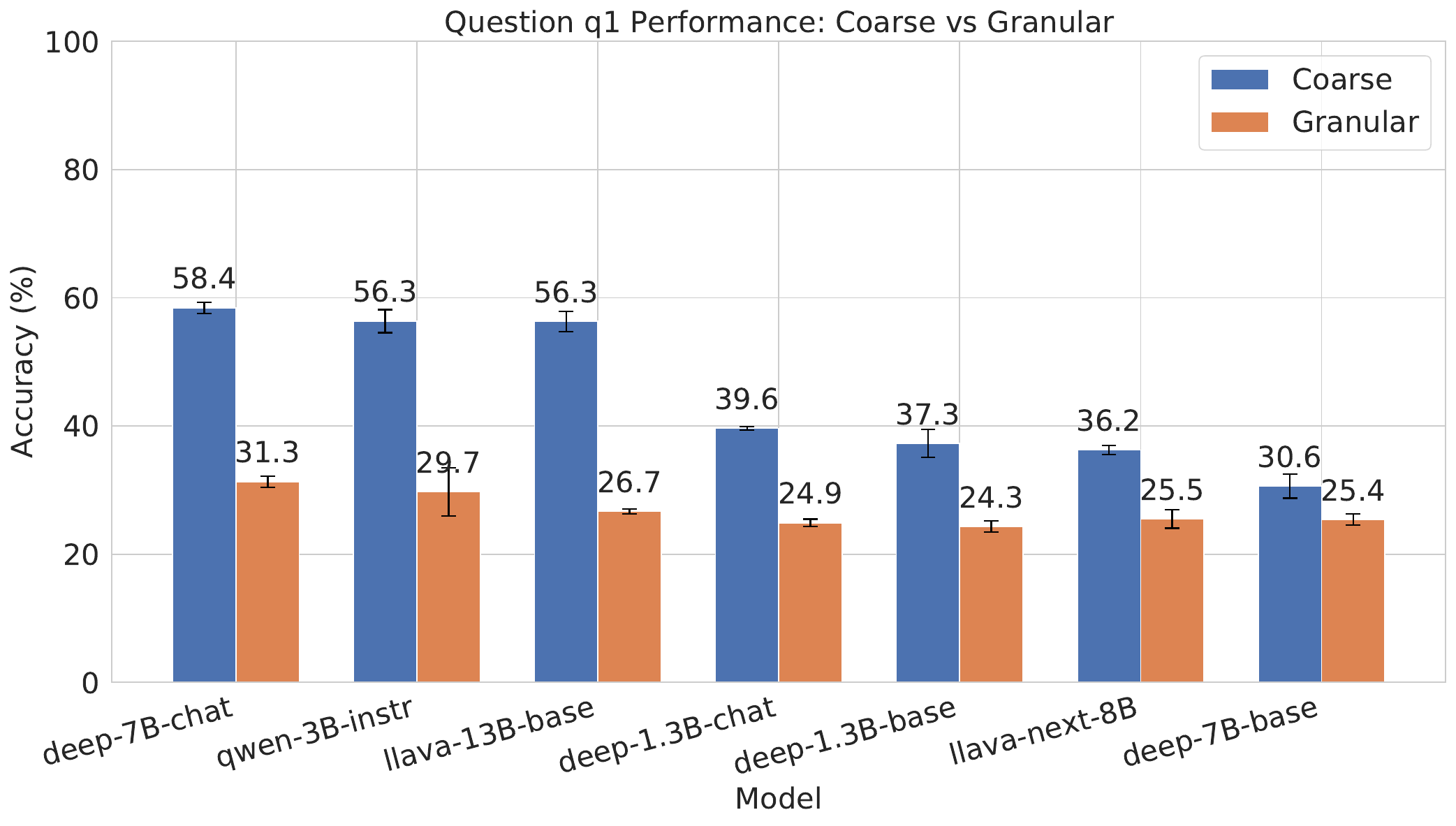}
    \caption{Mean and Standard Deviation of various models on View Parallelism task}
    \label{fig:q1_performance}
\end{figure}
\begin{figure}[ht]
    \centering
    \includegraphics[width=0.8\linewidth]{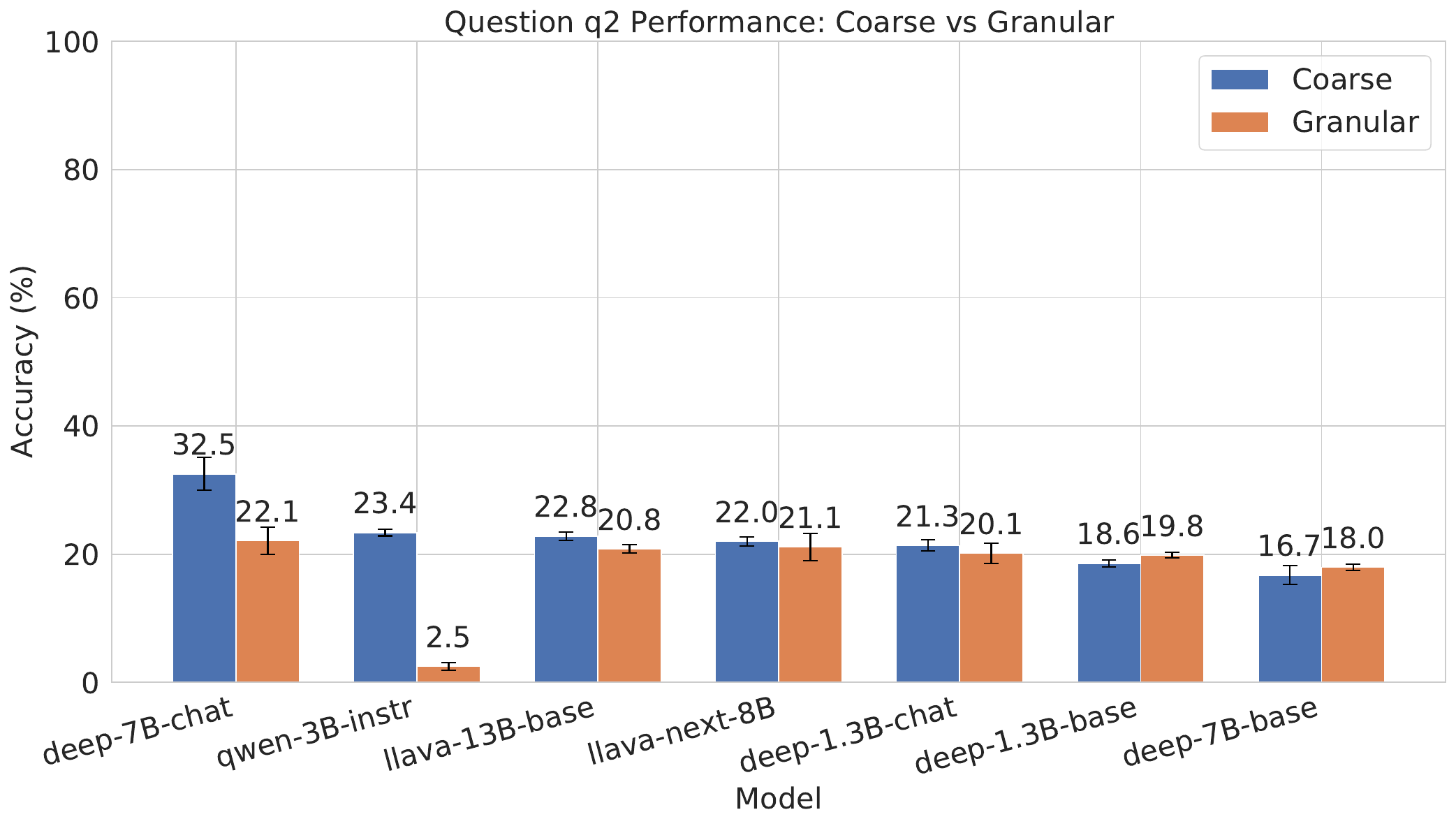}
    \caption{Mean and Standard Deviation of various models on Directional Facing task}
    \label{fig:q2_performance}
\end{figure}
\begin{figure}[ht]
    \centering
    \includegraphics[width=0.8\linewidth]{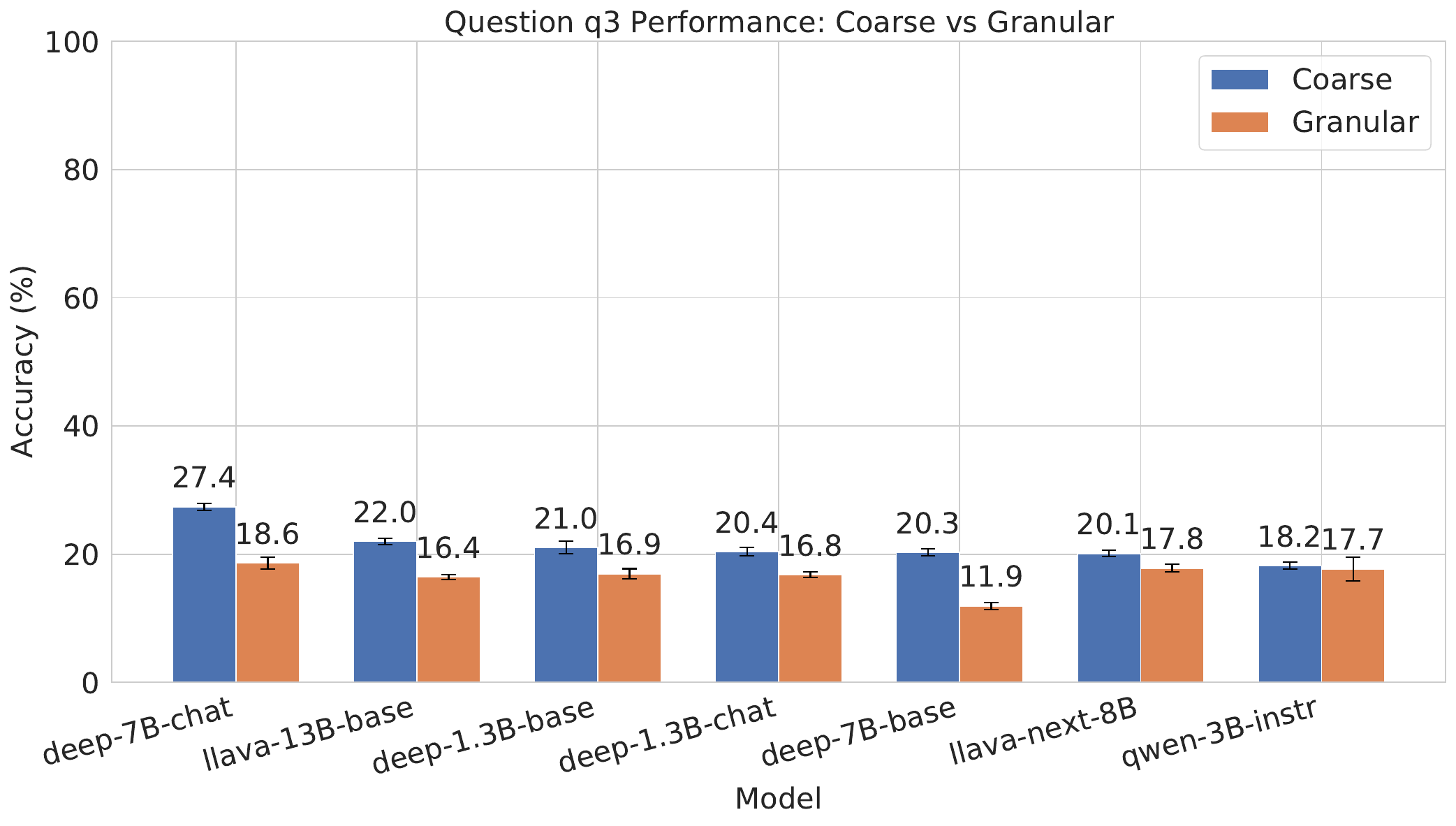}
    \caption{Mean and Standard Deviation of various models on Single-axis Rotation task}
    \label{fig:q3_performance}
\end{figure}
\begin{figure}[ht]
    \centering
    \includegraphics[width=0.8\linewidth]{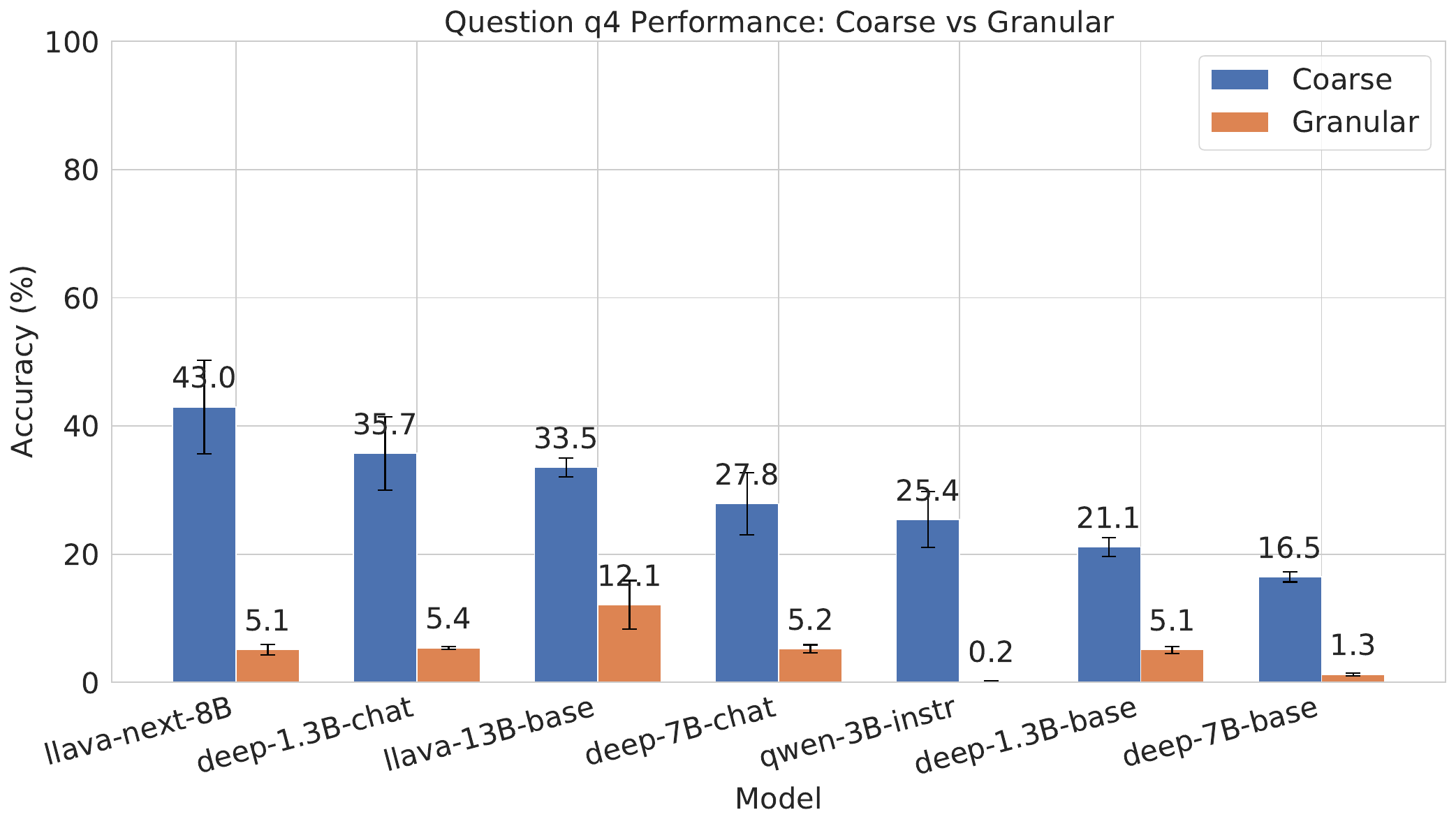}
    \caption{Mean and Standard Deviation of various models on Compound Rotation task}
    \label{fig:q4_performance}
\end{figure}
\begin{figure}[ht]
    \centering
    \includegraphics[width=0.8\linewidth]{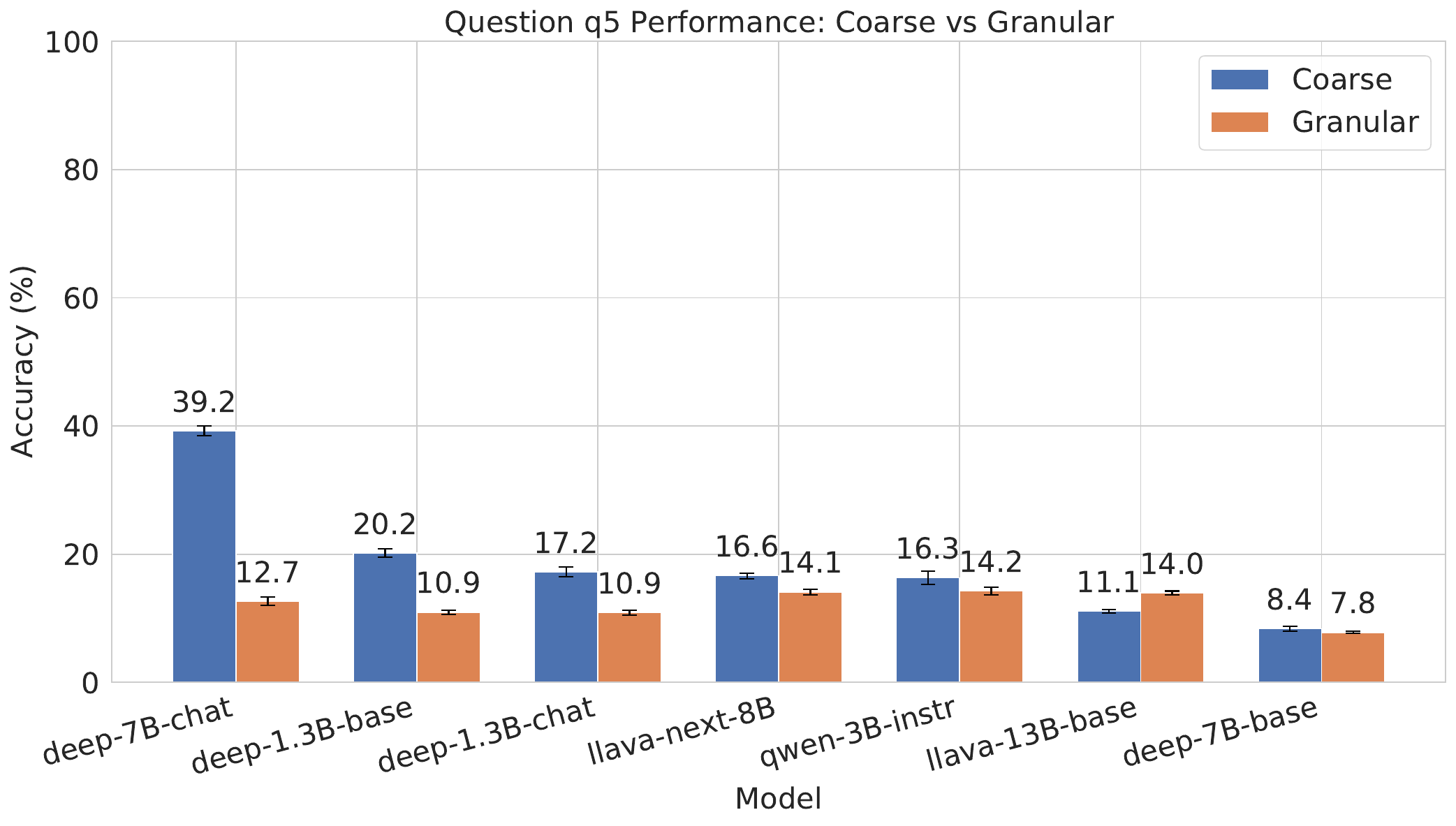}
    \caption{Mean and Standard Deviation of various models on Inter-object direction.}
    \label{fig:q5_performance}
\end{figure}

\begin{figure}[ht]
    \centering
    \includegraphics[width=0.8\linewidth]{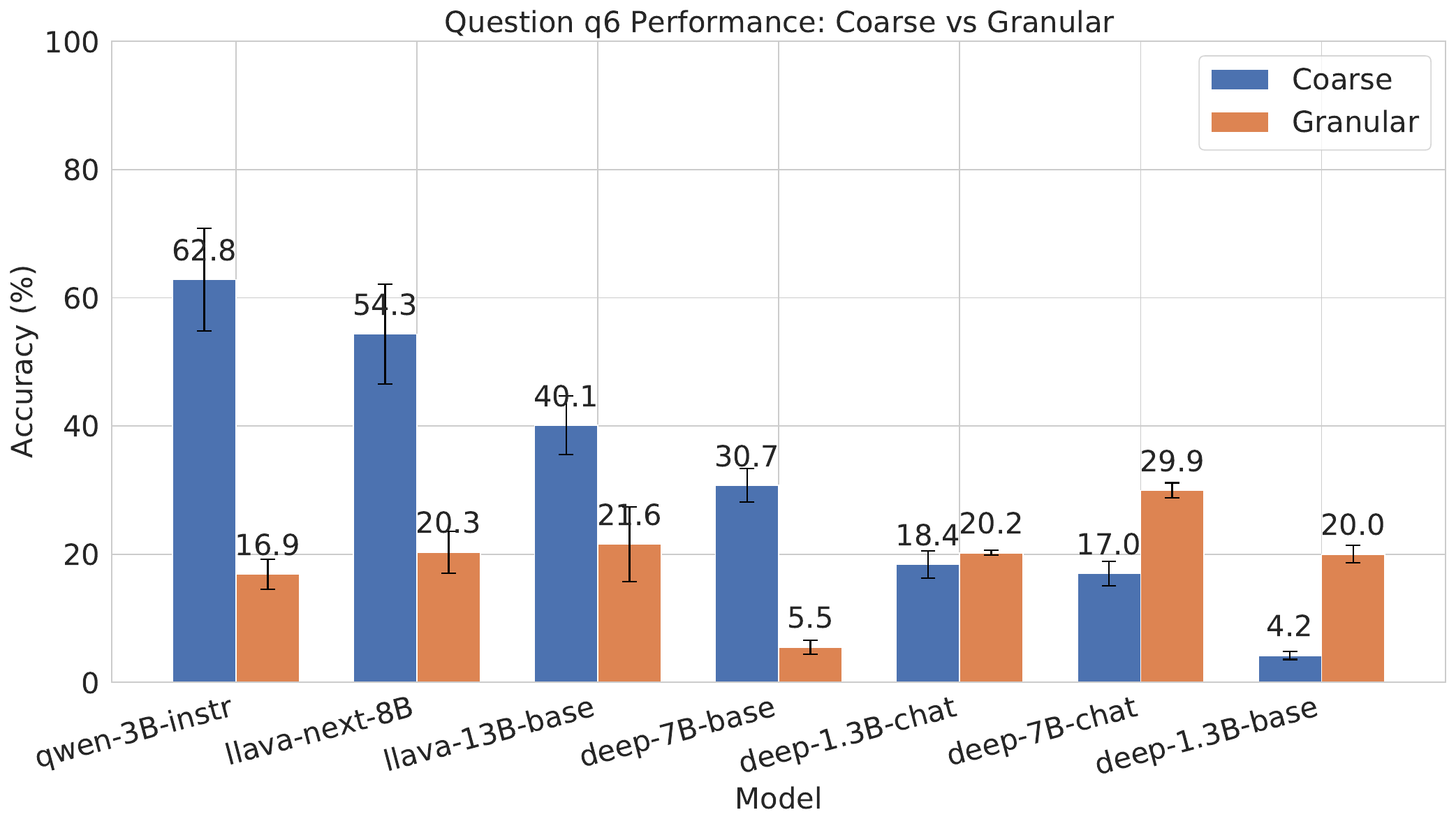}
    \caption{Mean and Standard Deviation of various models on  Viewer-Scene direction task.}
    \label{fig:q6_performance}
\end{figure}

\begin{figure}[ht]
    \centering
    \includegraphics[width=0.8\linewidth]{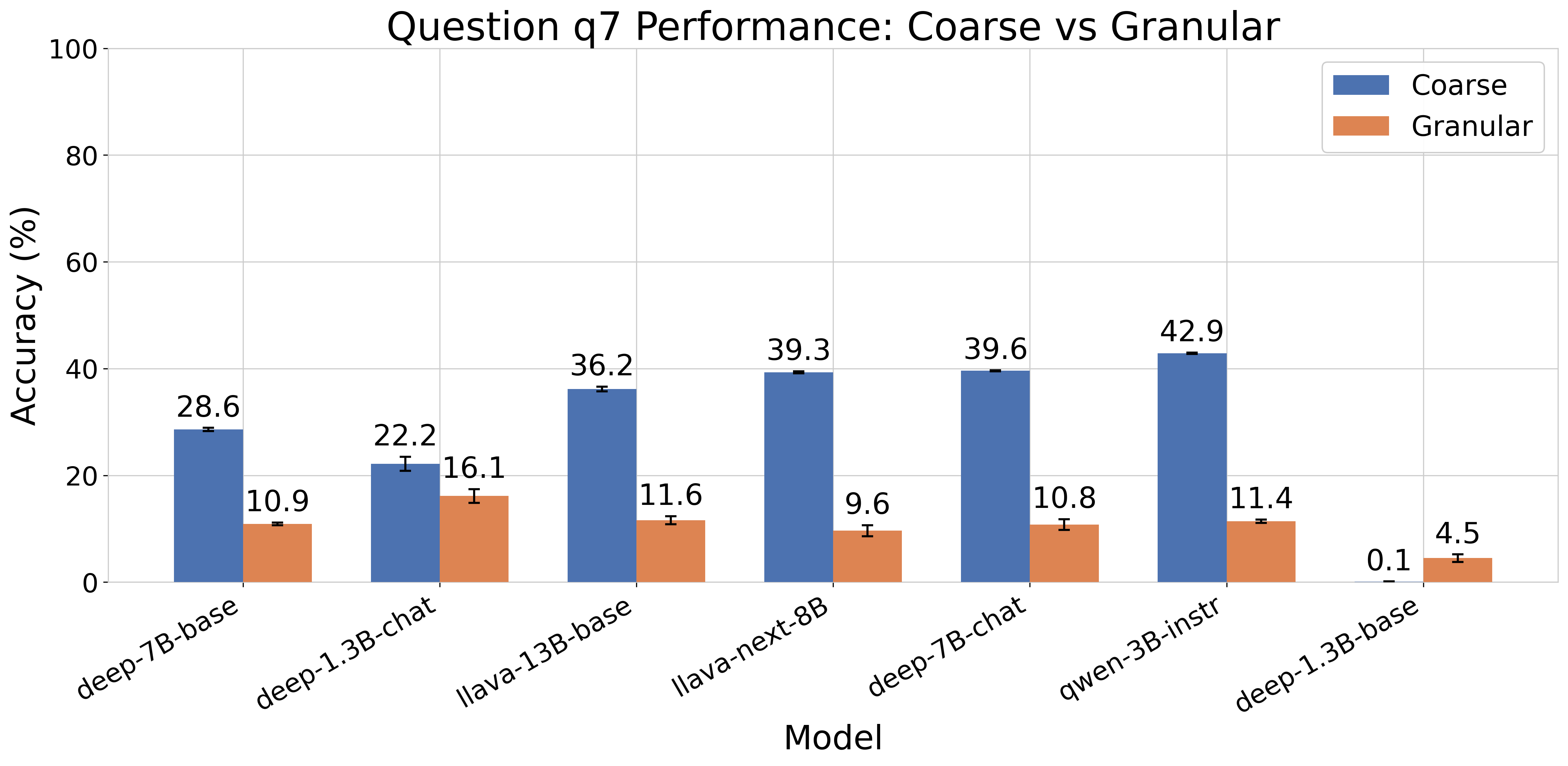}
    \caption{Mean and Standard Deviation of various models on the Canonical Orientation task.}
    \label{fig:q7_performance}
\end{figure}
\begin{table*}[t]
\centering
\caption{Out-Of-Pretraining (OOP) performance when evaluated on the COCO and Cityscapes dataset.}
\label{tab:oop_results}
\begin{tabular}{lcccc}
\toprule
COCO        & All & OOP & All & OOP  \\ \midrule
Category (Person)   & Coarse & Coarse & Granular & Granular\\
LLava-v1.6-13B\cite{liu2023llava}    & 23.0 & 25.1 & 20.8 & 21.0 \\ 
LLava-v1.6-34B\cite{liu2023llava}    & 42.6 & 41.3 & 39.6 & 39.6 \\
Yi-VL-6B\cite{ai2024yi}          & 30.3 & 28.6 & 33.5 & 33.1 \\
Deepseek-Base-7B\cite{bi2024deepseek}  & 17.7 & 17.5 & 17.7 & 17.9 \\ 
\end{tabular}
\begin{tabular}{lcccc}
\toprule
Cityscapes        & All & OOP & All & OOP  \\ \midrule
Category (Car)   & Coarse & Coarse & Granular & Granular\\
LLava-v1.6-13B\cite{liu2023llava}     & 22.6 & 25.7 & 16.7 & 18.1\\ 
LLava-v1.6-34B\cite{liu2023llava}     & 36.5 & 41.1 & 21.6 & 21.7\\
Yi-VL-6B\cite{ai2024yi}           & 25.6 & 29.1 & 21.6 & 22.1\\
Deepseek-Base-7B\cite{bi2024deepseek}  & 17.5 & 18.5 & 16.9 & 17.1\\ \bottomrule
\end{tabular}
\end{table*}
\subsection{Out-Of-Pretraining Analysis}
\label{app: OOP}
Most MLLMs have closed pretraining datasets, \ie, it is not possible to be completely sure what they were trained on. However, to provide insight into what the MLLM model might have learned (\ie, if it saw similar images in the pretraining dataset) we use the approach described in ~\cite{teterwak2024} to filter images in COCO and Cityscapes into those that were likely seen during pretraining. Specifically, we measured the cosine similarity between text features representing the object our question referred to (\eg, "a photo of a person" for a question asking about the orientation of a person) and the image using LLaVa-13B. Those with high similarity (using a 0.19 threshold for COCO and 0.18 for Cityscapes) were removed, leaving only images that were not learned well by the model during pretraining. On COCO we evaluated performance on person images, which accounted for 547 images in the Directional Facing questions, of which 253 were removing (leaving 294 images). On Cityscapes Directional Facing, we filtered based on car questions, resulting in removing 122 of 342 car images, leaving 220 images. Cars and person categories were selected as they were common objects in their respective datasets. We refer to these splits as ALL, which includes questions that have every car or person image, and Out-Of-Pretraining (OOP) for the images that remained after our filtering process. We then compare the performance on a number of models armed with these splits, as seen in \cref{tab:oop_results}.

Armed with these splits, we compared performance on LLava-13B, LLava-34B, Yi-VL-6B and Deepseek-7B, shown below. Interestingly, the OOP results are generally better than ALL, which seems counterintuitive at first glance. However, we suspect that this is an effect of being pretrained with a different objective than one focused on orientation. For example, if a model was pretrained to align images to alt-text, which give a high level description of the image that often does not include orientation information, then when the model saw similar images, it naturally assumes that the goal is to match to a high-level description and extracts features accordingly, essentially overfitting to that task. That said, there are many confounding factors that could provide alternative explanations, and is an interesting avenue for exploration in future work.
\clearpage

\section{Human Evaluation Interface}
\label{app:human_instructions}
An example of the high level instructions shown for this task can be seen in \cref{fig:human_eval_q3_example_high}, examples of questions are shown in \cref{fig:human_eval_q3_example} and \cref{fig:human_eval_q6_example}

\begin{figure}[t]
    \centering
    \includegraphics[width=\linewidth]{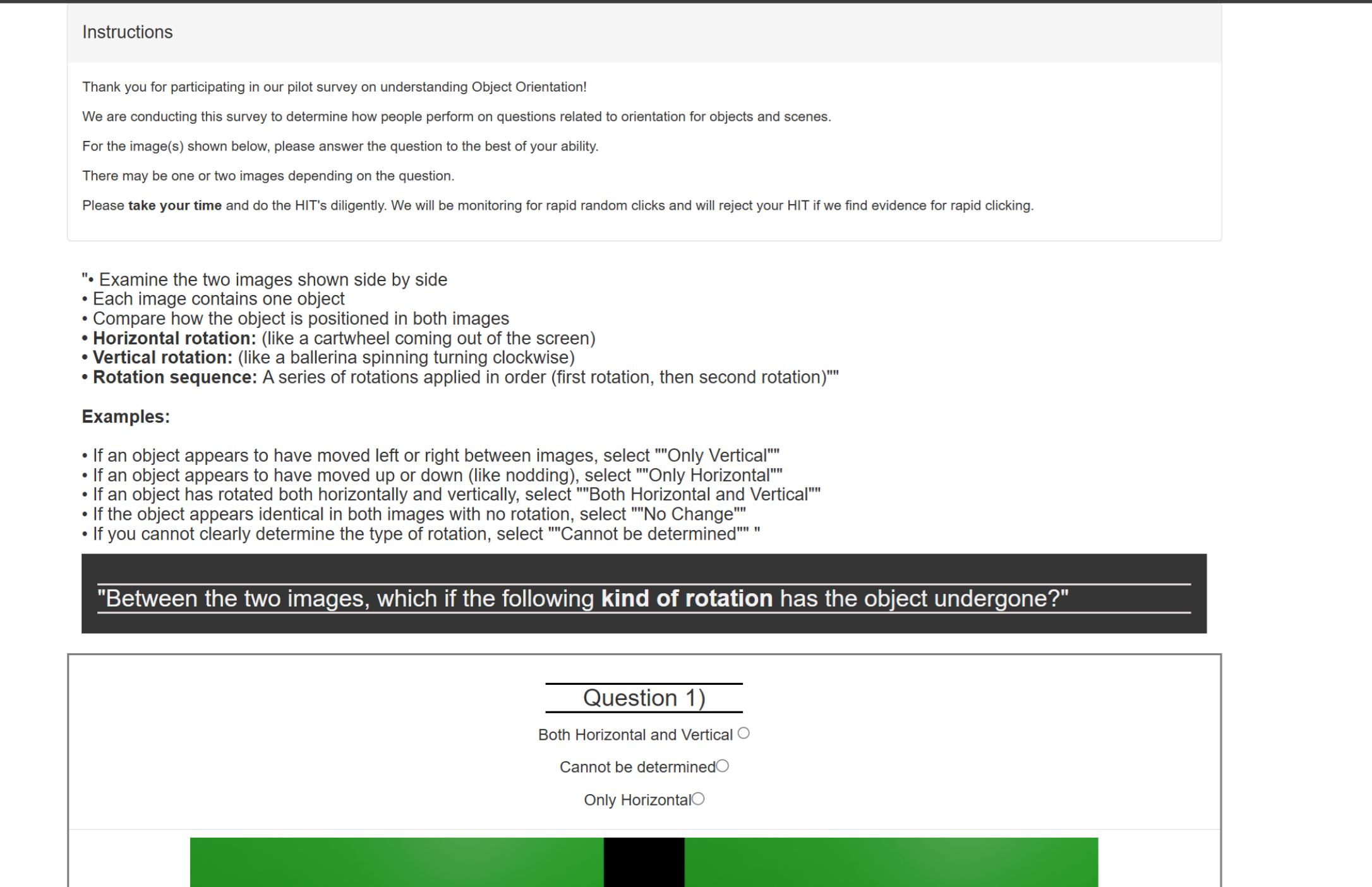}
    \caption{An example of the high level instructions shown for the Human Evaluation}
    \label{fig:human_eval_q3_example_high}
\end{figure}
\begin{figure}[t]
    \centering
    \includegraphics[width=\linewidth]{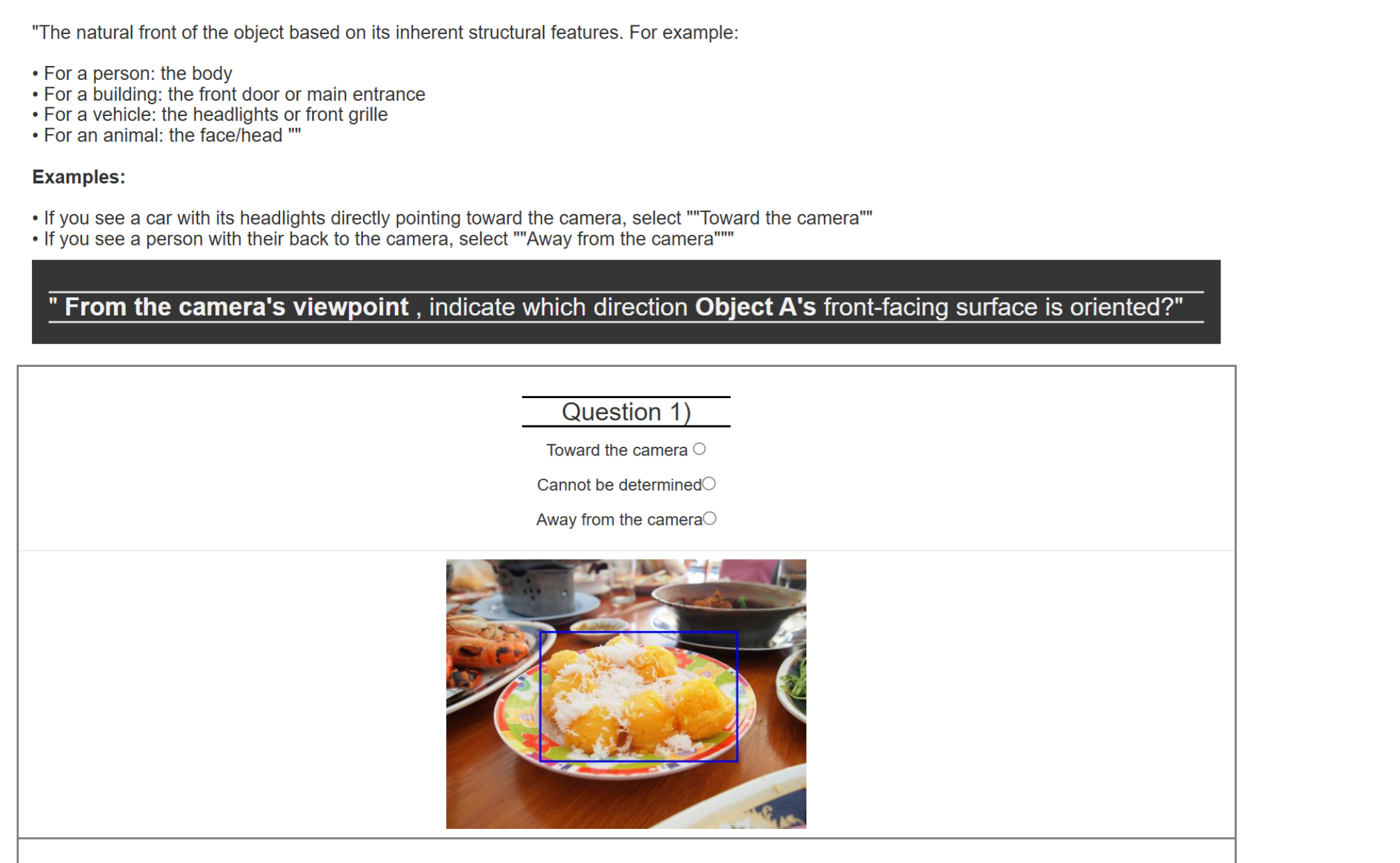}
    \caption{An example of a sample for the coarse-level VQA for the Directional Facing task in DORI shown for the Human Evaluation}
    \label{fig:human_eval_q3_example}
\end{figure}
\begin{figure}[t]
    \centering
    \includegraphics[width=\linewidth]{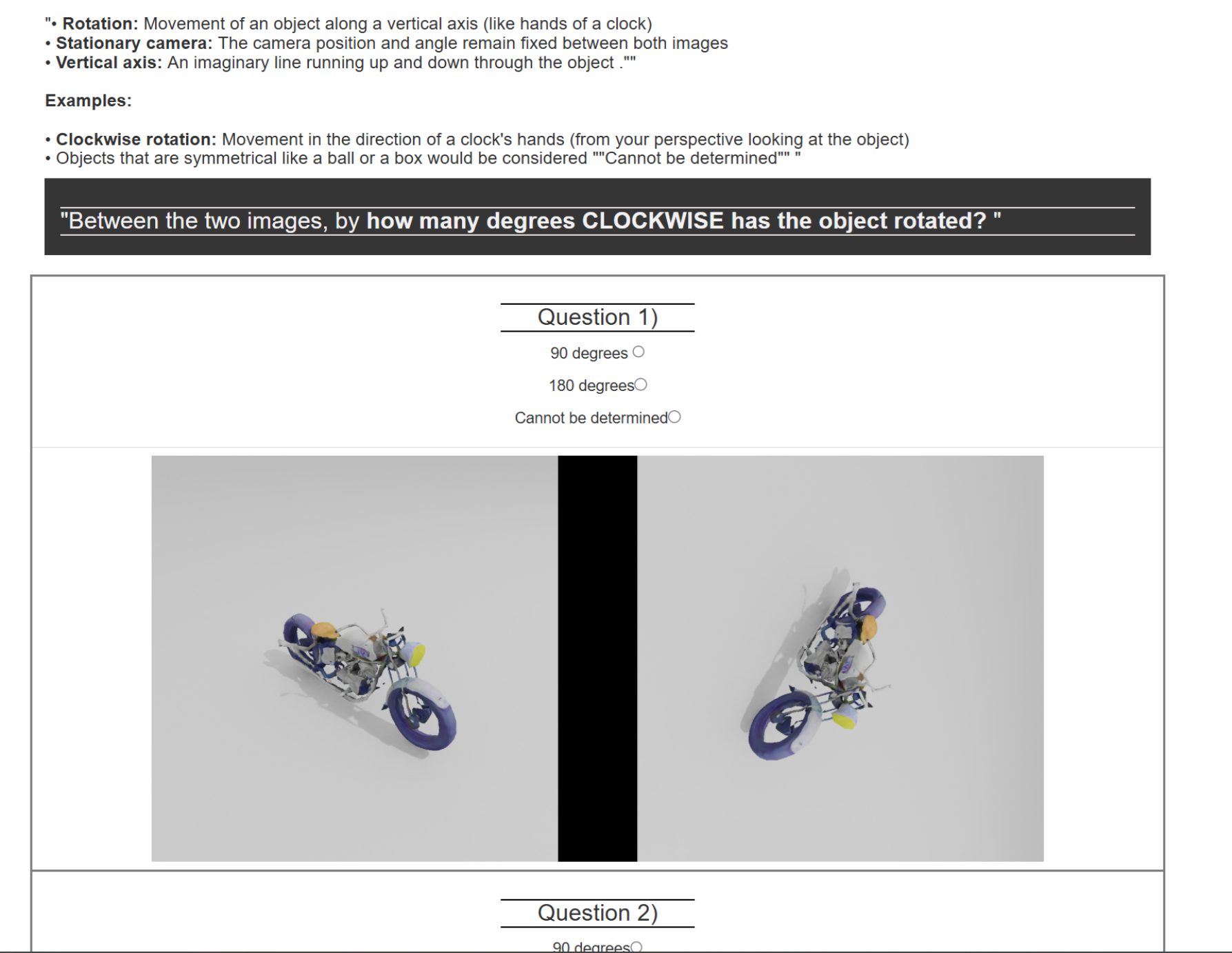}
    \caption{An example of a sample for the coarse-level VQA for the Viewer-Scenen Direction task in DORI shown for the Human Evaluation}
    \label{fig:human_eval_q6_example}
\end{figure}
\clearpage
\section{Terminology Glossary}
\label{ref: glossary}
\noindent\textbf{Front-facing surface / Frontal surface.} The front-facing surface (or frontal surface) of an object is the side that carries its primary functional or semantic features, such as a person’s face and torso, a car’s headlights and grille, or a laptop’s screen, and this surface is what is considered to be facing the camera when those features are visible and oriented toward the viewer (\eg, a person standing and looking directly at the camera is presented with their frontal surface clearly visible).
\smallskip

\noindent\textbf{Canonical orientation.}
Canonical orientation is the typical, upright, real-world pose in which an object is most commonly encountered or depicted, often aligned with gravity and human conventions (for instance, a car resting on its wheels on a horizontal road, a building with the roof at the top and the foundation at the bottom, or a mug standing upright with the handle to the side), and questions about canonical orientation ask whether the object is in this usual pose or has been rotated or flipped away from it (\eg, a building image rotated 180° so the roof is at the bottom would be judged non-canonical).
\smallskip

\noindent\textbf{Camera plane / Camera axis.}
The camera plane (or image plane) is the 2D plane onto which the 3D scene is projected by the camera, while the camera axis (optical axis) is the line perpendicular to this plane passing through the camera center, so that orientation relative to the camera is described by how an object’s surfaces are angled with respect to this plane and axis (for example, a frontal surface parallel to the image plane and centered on the optical axis corresponds to an object facing straight toward the camera).
\smallskip

\noindent\textbf{Camera Viewpoint vs. Object Viewpoint.}
The camera viewpoint is the position and orientation of the camera in the 3D scene, defining an egocentric frame where directions like “toward the camera” or “left of the camera” live, whereas the object viewpoint is the orientation and placement of the object relative to its own intrinsic front/back/left/right axes or relative to other objects, so a question might either ask how the object appears from the camera’s perspective (“Is the car facing toward the camera?”) or, alternatively, how the object is oriented if one imagines standing at the object and looking out (“From the car’s viewpoint, is it facing toward or away from the truck?”).
\smallskip

\noindent\textbf{Rotation transformation.}
A rotation transformation changes an object’s orientation by turning it around one or more axes while preserving its shape and size, such as rotating a mug 90° around the vertical axis to bring its handle from the right side to face the camera or rotating a picture frame 180° in the image plane so that it appears upside down relative to its original orientation.
\smallskip

\noindent\textbf{Sequential transformations.}
Sequential transformations refer to applying multiple geometric operations in a specific order, typically rotations and flips, where order matters because compositions can yield different final poses, for example first rotating a building image 90° clockwise in the image plane and then flipping it horizontally produces a different orientation than performing the horizontal flip first and then rotating, even though the same primitive operations are used.
\smallskip

\noindent\textbf{Vertical axis rotation.}
Vertical axis rotation is a 3D rotation around an axis that runs approximately up–down through the object and aligns with gravity (analogous to yaw), changing which horizontal direction the object’s front faces, so turning a car so that it goes from facing east to facing north is a vertical axis rotation of about 90° and a person spinning in place to look from the camera to the right-hand side is another example of such a rotation.
\smallskip

\noindent\textbf{Horizontal axis rotation.}
Horizontal axis rotation is a 3D rotation around an axis that runs roughly left–right through the object (analogous to pitch), changing how much the object is tilted toward or away from the camera, so when a person nods their head “yes” or a box tips forward so its top leans toward the camera, these are horizontal axis rotations that alter how the object’s surfaces are foreshortened in the image.
\smallskip

\noindent\textbf{Lateral Axis Rotation.}
Lateral axis rotation is a rotation around an axis that runs roughly front–back through the object (analogous to roll), causing the object to lean sideways relative to gravity and the image frame, so tilting your head so that your ear moves toward your shoulder or rolling a car so that it lies partially on its side are examples of lateral axis rotations that change which parts appear above or below in the image without changing which direction the object’s front is pointing.
\smallskip

\noindent\textbf{Clockwise rotation.}
Clockwise rotation is defined with respect to a chosen viewing direction and axis: in this context, for rotations around the vertical axis, it means that when viewed from above along the vertical axis, the object’s front turns in the same direction as the hands of a clock (\eg, a car that initially faces north and then rotates to face east has undergone a 90° clockwise rotation about the vertical axis), and in image-plane contexts, clockwise means the top of the object appears to rotate toward the right side of the image.
\smallskip

\noindent\textbf{Counter-clockwise rotation.}
Counter-clockwise rotation is the opposite directional sense from clockwise around a specified axis: for vertical axis rotations observed from above, the object’s front turns in the opposite direction to clock hands (\eg, a car rotating from facing east to facing north has rotated 90° counter-clockwise about the vertical axis), and in the image plane, counter-clockwise means the top of the object appears to rotate toward the left side of the image. We include examples of what clockwise and counter-clockwise rotations looks like as seen in \cref{fig:clockwise_and_counter_clockwise} for the Inter-object direction perception task, we utilize samples from the NOCS REAL dataset to highlight these different question samples.
\smallskip

\noindent\textbf{Facing each other (objects' perspective).}
Two objects are said to be facing each other when each object’s front-facing surface points approximately toward the other’s position in the scene in an allocentric or object-centric frame, so that if you imagine rays extending from their fronts, those rays intersect between them (for example, two chairs arranged on opposite sides of a table with their seats and backs oriented toward the table center are facing each other).
\smallskip

\noindent\textbf{Facing same/opposite directions.} Objects face the same direction when their front-facing axes are approximately parallel in 3D space, meaning they are oriented with similar yaw so they would move side-by-side if they advanced forward together, whereas they face opposite directions when their front-facing axes differ by about 180°, such as two cars parked in the same lane but one pointing north and the other pointing south along the road.
\smallskip

\noindent\textbf{Perpendicular facing.}
Two objects are perpendicular in facing when the angle between their front-facing directions is approximately 90° in yaw, such that one object’s front points roughly to the side of the other, for example, when one car is parked facing north and another is parked facing east at an intersection, so their fronts form an L-shape orientation relative to each other.
\smallskip

\noindent\textbf{Flip / Mirror.}
A flip or mirror transformation reverses an image or object along a specified axis without changing its depth order, so a horizontal flip mirrors the scene left–right (like viewing it in a mirror placed vertically), turning a car that originally faces left into one that appears to face right, while a vertical flip mirrors the scene top–bottom, making a building appear upside down with the roof at the bottom of the image.
\smallskip

\noindent\textbf{Ambiguous objects / Symmetrical objects.}
Ambiguous or symmetrical objects are those whose shape or appearance does not specify a unique front or canonical orientation because many rotations produce indistinguishable images, such as a perfect sphere, a uniform cylinder, or a highly symmetric abstract logo, so for such objects questions about “which way it is facing” or “whether it is upside down” may have no well-defined answer and are therefore treated as “Cannot be determined” cases in the dataset.
\clearpage

\bibliographystyle{splncs04}
\bibliography{main}
\end{document}